\documentclass[lettersize,journal]{IEEEtran}
\usepackage{amsmath,amsfonts}
\usepackage{algorithmic}
\usepackage{algorithm}
\usepackage{array}
\usepackage[caption=false,font=normalsize,labelfont=sf,textfont=sf]{subfig}
\usepackage{textcomp}
\usepackage{stfloats}
\usepackage{float}
\usepackage{url}
\usepackage{verbatim}
\usepackage{multirow} 
\usepackage{booktabs}
\usepackage{pifont}       
\usepackage{xcolor}

\newcolumntype{C}[1]{>{\centering\arraybackslash}p{#1}}

\usepackage{graphicx}
\usepackage{cite}  
\usepackage[colorlinks=true, linkcolor=blue, citecolor=blue, urlcolor=blue]{hyperref}
\makeatletter
\renewcommand\@cite[2]{\textcolor{blue}{[#1\if@tempswa, #2\fi]}}
\makeatother
\begin{document}

\title{MultiShadow: Multi-Object Shadow Generation for Image Compositing via Diffusion Model}

\author{Waqas Ahmed, Dean Diepeveen, Ferdous Sohel~\IEEEmembership{Senior Member,~IEEE}
        
\thanks{}

\thanks{Waqas Ahmed, Dean Diepeveen and Ferdous Sohel are with the School of Information  Technology, Murdoch University, Murdoch, WA 6150, Australia. E-mail: waqas.ahmed@murdoch.edu.au, d.diepeveen@murdoch.edu.au and F.Sohel@murdoch.edu.au}

\thanks{Corresponding author: Ferdous Sohel.}
\thanks{}
}

\markboth{}%
{Shell \MakeLowercase{\textit{et al.}}: A Sample Article Using IEEEtran.cls for IEEE Journals}

\maketitle

\begin{abstract}
Realistic shadow generation is crucial for achieving seamless image compositing, yet existing methods primarily focus on single-object insertion and often fail to generalize when multiple foreground objects are composited into a background scene. In practice, however, modern compositing pipelines and real-world applications often insert multiple objects simultaneously, necessitating shadows that are jointly consistent in terms of geometry, attachment, and location. In this paper, we address the under-explored problem of multi-object shadow generation, aiming to synthesize physically plausible shadows for multiple inserted objects. Our approach exploits the multimodal capabilities of a pre-trained text-to-image diffusion model. An image pathway injects dense, multi-scale features to provide fine-grained spatial guidance, while a text-based pathway encodes per-object shadow bounding boxes as learned positional tokens and fuses them via cross-attention. An attention-alignment loss further grounds these tokens to their corresponding shadow regions. To support this task, we augment the DESOBAv2 dataset by constructing composite scenes with multiple inserted objects and automatically derive prompts combining object category and shadow positioning information. Experimental results demonstrate that our method achieves state-of-the-art performance in both single and multi-object shadow generation settings.
\end{abstract}


\section{INTRODUCTION}
\IEEEPARstart{T}{he} advent of generative models has significantly advanced the field of image compositing~\cite{niu2021making}, which involves inserting a foreground object into a new background scene. However, generating physically plausible shadows for the newly inserted objects remains a critical challenge, as even subtle inconsistencies in shadow shape, location, intensity, or contextual alignment can immediately break the illusion of realism. This difficulty is further amplified by recent compositing workflows that increasingly support inserting multiple foreground objects into a single scene. In such multi-object composites, shadows must not only be individually correct for each object, but also globally consistent; i.e., the shadow generation process must jointly reason about all shadow locations, geometries, and alignments to ensure global consistency across the scene. In this paper, we propose a framework for realistic shadow generation in both single and multi-object settings.

Existing shadow-generation methods perform reasonably well for single-object shadow generation but exhibit clear limitations in multi-object settings. Methods such as DAMASNet \cite{tao2024shadow}, ASG-EF \cite{meng2023automatic},  and CFDiffusion \cite{yu2024cfdiffusion} do not support one-pass multiple object shadow generation; instead, they are typically applied sequentially, generating shadows one object at a time. This per-object approach leads to shadow inconsistencies due to error accumulation, where artifacts or misplacement in shadows generated for earlier objects propagate and bias subsequent objects shadow predictions. In contrast, architectures such as SGDiffusion \cite{liu2024shadow}, GPSDiffusion \cite{zhao2025shadow}, and MetaShadow \cite{wang2025metashadow}, though originally designed for single-object shadow generation, can be applied in a single pass to multi-object composites. However, when extended to multiple object shadow generation, they often introduce artifacts and exhibit inconsistencies in shadow direction, geometry, or intensity across objects. 

Most recent shadow-generation pipelines \cite{liu2024shadow}, \cite{zhao2025shadow}, \cite{wang2025metashadow} adopt diffusion models~\cite{rombach2022high} and rely solely on visual conditioning. Concretely, they train image-to-image translation networks on paired data consisting of a shadow-free composite, object and shadow masks, and the corresponding ground-truth shadowed image. While such pixel-space conditioning provides better local guidance, it faces a fundamental difficulty in multiple object composites, which is maintaining a consistent association between each foreground object and its corresponding shadow while enforcing global consistency across all objects. To improve shadow placement and geometry, existing pipelines often incorporate auxiliary image-based predictors (e.g., shadow location or shape estimators) and use their outputs as additional image-based conditioning inputs to the diffusion model. However, extending these designs to multi-object scenes requires predicting and coordinating multiple per-object shadow properties in image space (e.g., location, extent, and shape), which increases system complexity and makes the overall pipeline less scalable and more brittle as the number of inserted objects grows.

This raises a key question: can we exploit the multimodal capabilities of diffusion backbones to provide an explicit and scalable conditioning mechanism for multi-object shadow generation? A line of work~\cite{yang2023reco}, \cite{li2023gligen}, \cite{zheng2023layoutdiffusion} has shown that diffusion models exhibit significantly enhanced controllability when conditioned on grounding information, such as bounding-box coordinates and labels, which provide region-level specificity and a consistent bounding box frame for shape reasoning. At the same time, pre-trained text-to-image diffusion models \cite{rombach2022high, yu2024uncovering} are trained on large-scale image–caption datasets and thus encode rich semantic knowledge, implicitly understanding object categories (e.g., cat, girl), interactions (e.g., driving, sitting), and relational phrases such as “casting a shadow”. Motivated by these observations, we introduce a text-grounded layout mechanism that augments dense image conditioning with explicit per-object shadow grounding. Specifically, we encode each object’s predicted shadow layout as discrete positional tokens derived from shadow bounding boxes, and explicitly supervise the corresponding cross-attention maps to focus on the correct shadow regions. This design provides an object-aware indexing mechanism that complements pixel-aligned geometric guidance from the image pathway: the image branch supplies fine-grained local geometry, while grounded tokens enforce consistent object–shadow association and reduce cross-instance interference in multi-object scenes.

Building on this idea, we introduce a diffusion-based framework that adopts a pre-trained text-to-image diffusion model with dual conditioning: an image pathway and a shadow-grounded prompt pathway. First, we extend the DESOBAv2 \cite{liu2023desobav2} dataset by scenes with multiple inserted foreground objects, each requiring a realistic shadow. On the image side, the shadow-free composite and the foreground object mask are fed to an image conditioning module that injects multi-scale, pixel-aligned features into the UNet via geometry-aware affine modulation, providing strong fine-grained guidance. On the shadow-grounded prompt side, we construct a generic prompt for each image by combining the object categories with their corresponding shadow bounding box information, in the form of phrases such as “a cat casting shadow [shadow-box tokens]”. For object category extraction we applied ViP-LLaVA \cite{cai2024vip} model. When interactions like “riding or sitting” are detected between two objects, the action is appended to produce more expressive prompts. The shadow box positions are predicted by a separately trained shadow-box predictor network and transformed into shadow positional tokens. The resulting prompt is encoded by the text encoder and conditions the UNet through cross-attention. Meanwhile, an attention alignment loss explicitly encourages the attention maps of the shadow positional tokens to focus on their corresponding shadow regions. Together, these components turn per-object shadow information into an explicit signal that complements image-based conditioning, enabling the model to generate realistic, spatially coherent shadows for all composited objects in the scene. In summary, our contributions are:
\begin{itemize}
    \item We present the first shadow generation framework that explicitly addresses the problem of multi-object shadow generation, overcoming the limitations of prior methods.
    \item We introduce a novel text-grounded shadow generation mechanism that augments dense image conditioning with a text-grounded layout pathway, where each object’s shadow layout is represented by learnable positional tokens derived from predicted shadow boxes and injected via cross-attention, explicitly targeting multi-object failure modes.

    \item  We propose an attention alignment loss that supervises cross-attention to focus on corresponding shadow regions, improving object–shadow correspondence and effectively grounding these tokens in the image plane.

    \item  Extensive experiments demonstrate that our method significantly outperforms purely image-based methods, producing realistic and spatially consistent shadows across all composited objects.
\end{itemize}

\begin{figure*}[h]
    \centering
    \includegraphics[width=1.0\linewidth]{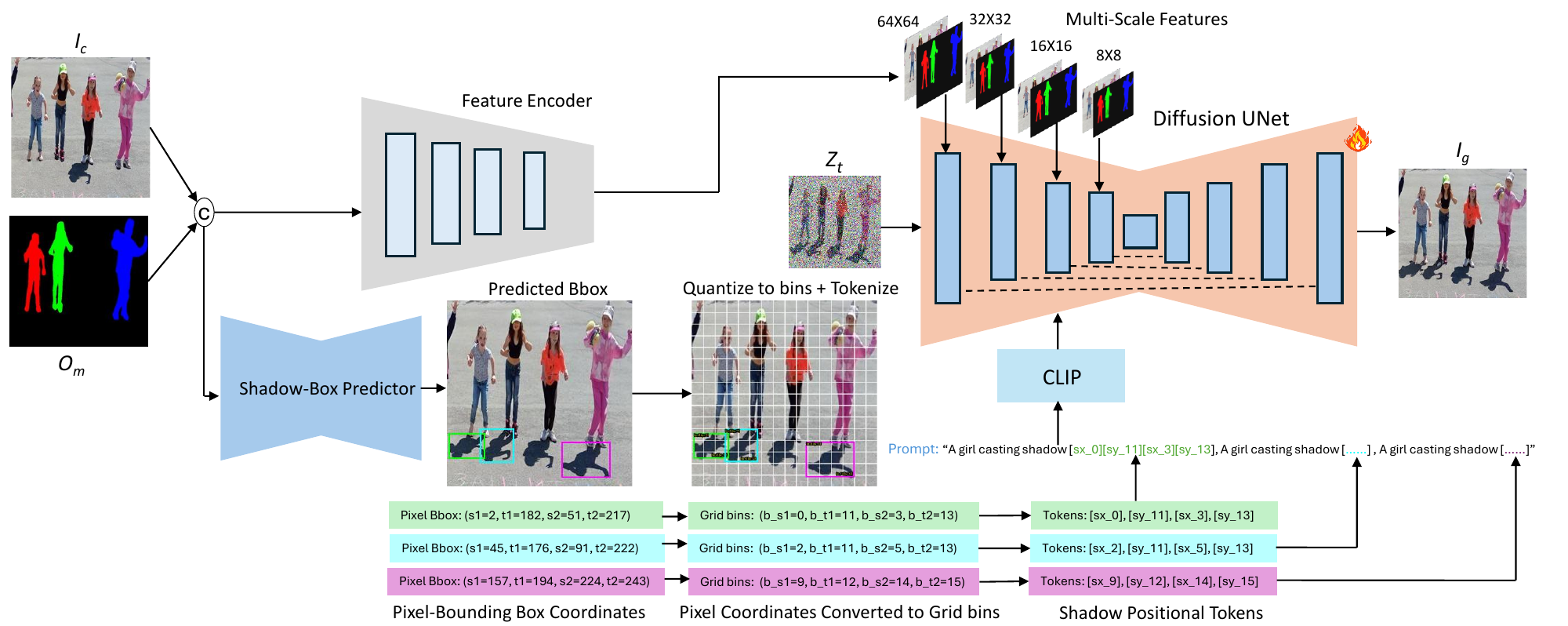}
    \caption{\textbf{Architectural Overview.} From a composite image $I_c$ and corresponding object masks $O_m$, a Shadow-Box Predictor estimates per-object shadow bounding boxes, which are quantized into grid bins and converted into shadow positional tokens. These positional tokens are inserted into the prompt and processed through a CLIP encoder to provide text-conditioned spatial grounding. On the image pathway, a Feature Encoder extracts multi-scale visual features and injects them into the diffusion UNet. Both conditioning streams jointly guide the diffusion model to generate the final shadowed image $I_g$.}
    \label{fig:architecture1}
\end{figure*}
 \section{RELATED WORK}
\subsection{Image Composition}
Image compositing~\cite{niu2021making} aims to seamlessly integrate foreground objects into a background scene. With the growing sophistication of modern image synthesis techniques, the field has progressed beyond traditional copy–paste and basic blending methods to diffusion-based approaches~\cite{rombach2022high}. Models such as ObjectStitch~\cite{song2022objectstitch}, Paint-by-Example~\cite{yang2023paint}, and AnyScene~\cite{chen2024anyscene} have significantly improved texture blending, foreground–background alignment, and contextual coherence. More recent works, such as AnyDoor~\cite{chen2025anydoor}, Multitwine~\cite{tarres2025multitwine}, and BOOTPLACE~\cite{zhou2025bootplace}, further push the boundary by enabling simultaneous multi-object insertion into a single scene. However, despite these advances in object placement and appearance blending, existing compositing methods either neglect shadows entirely or produce only coarse and unrealistic shadow representations, severely undermining the visual plausibility of the composite. This limitation highlights the need for advanced shadow generation methods to support realistic image compositing. 

\subsection{Shadow Generation}
Existing shadow generation methods can be broadly categorized into two groups: rendering-based~\cite{sheng2021ssn, sheng2023pixht} and non-rendering-based~\cite{zhou2024foreground, yu2024cfdiffusion, zhao2025shadow}. Rendering-based shadow generation methods simulate the physical behavior of light by modeling scene geometry, lighting, and materials. While they offer better results, these methods often rely on detailed scene information or manual input, which limits their scalability and use in real-world settings.

Non-rendering-based methods bypass this requirement by learning a direct mapping from a shadow-free composite to a shadowed image. Early learning-based approaches commonly use GAN-style objectives to improve realism. For example, ShadowGAN~\cite{zhang2019shadowgan} employs both global and local conditional discriminators to enhance the realism of generated shadows, while ARShadowGAN~\cite{liu2020arshadowgan} emphasizes background shadows and leverages them to guide foreground shadow generation~\cite{liu2024shadow}. Beyond adversarial training, SGRNet~\cite{hong2022shadow} pioneered this direction by introducing DESOBA, a real-world dataset composed of object and shadow pairs, and proposed a mask-then-fill strategy, where one branch predicts a foreground shadow mask, then a second stage synthesis and blends the shadow onto the background. ASG-EF \cite{meng2023automatic} propose an automatic shadow generation approach based on exposure fusion, where multiple under-exposed versions of the composite are synthesized and fused to form the final shadow region. DMASNet~\cite{tao2024shadow} extended this using a rendered dataset (RdSOBA) and produced a two-stage method. First, shadow location and shape prediction, followed by appearance-aware filling using features from real scene shadows.  

Recent advances leverage the generative power of diffusion models~\cite{rombach2022high} for this task. CFDiffusion~\cite{yu2024cfdiffusion} first predicts a shadow mask via a dedicated encoder and then feeds it into the diffusion model for shadow synthesis. SGDiffusion~\cite{liu2024shadow} adapts ControlNet~\cite{zhang2023adding} and adds an intensity encoder to align the darkness of generated shadows to background shadows.  GPSDiffusion~\cite{zhao2025shadow} incorporates geometry priors in the form of image-based rotated bounding-box and discrete shape information, which are fed into the ControlNet to preserve the shadow geometry and placement. MetaShadow \cite{wang2025metashadow} combines an object-centered GAN component with a diffusion-based shadow synthesizer to leverage shadow references.

Despite this progress, current shadow-generation pipelines remain largely single-object–centric, and even in single-object settings they are not consistently reliable under diverse scenes. Extending these methods to multi-object composites is often inefficient (e.g., sequential per-object inference) or unstable (e.g., inconsistent geometry and intensity across instances), leaving a gap for robust and scalable multi-object shadow generation, an increasingly common requirement in modern image compositing pipelines.

\subsection{Grounding Conditions in Diffusion Model}
Recent advances in text-to-image diffusion models have demonstrated that explicit spatial grounding significantly enhances controllability over object placement and scene composition. GLIGEN \cite{li2023gligen} introduces an object grounding mechanism that binds semantic concepts to specific bounding boxes, enabling precise object localization. LayoutDiffusion \cite{zheng2023layoutdiffusion} treats user layouts and image patches as objects in a shared box space, enabling joint reasoning over positions, sizes, and inter-object relations. ReCo \cite{yang2023reco} further explores region-controlled generation by augmenting text prompts with quantized bounding-box coordinates for each region. These works collectively establish that region-specific grounding offers a powerful and flexible framework for spatially controlled image generation.

\section{METHODOLOGY}
Given a composite image \( I_c \) containing one or multiple objects inserted into a background, the goal is to synthesize a realistic shadowed image \( I_g \), where all inserted objects cast physically plausible shadows. 

To this end, we adapt a pre-trained diffusion model using a dual-conditioning design that combines an image-based pathway with a text-grounded shadow conditioning pathway, as illustrated in Figure~\ref{fig:architecture1}. The remainder of this section details our methodology, Section~\ref{sec:image_Cond} describes the image-based conditioning mechanism and its role in modulating UNet features, while Section~\ref{sec:text_layout} presents our novel shadow positional tokens and their injection strategy via cross-attention to ensure spatially coherent shadow generation across multiple objects.

\subsection{Image-Based Conditioning}
\label{sec:image_Cond}
To give the denoiser fine-grained, pixel-aligned control over shadow generation, we employ an image-based conditioning module that modulates the UNet at multiple depths. The module takes as input the concatenation of the shadow-free composite and the objects mask, and passes them through a feature extractor to produce a multi-scale feature pyramid aligned with the spatial resolutions of the UNet encoder. These features act as explicit guidance maps, helping the model infer accurate shadow attachment and geometry. We inject these features into the UNet via a Geometry-Aware Affine Modulation (GAAM) mechanism. Concretely, for a latent feature map \( x \) and its scaled local feature \( L \), we compute:
\begin{equation}
\text{GAAM}(x; L) = \text{GroupNorm}(x) \odot (1 + \Delta(L)) + B(L)
\end{equation}

where \( \Delta() \) and \( B() \) are spatially adaptive convolutional mappings that predict per-pixel scale and bias. This acts as an object-aware, location-sensitive gate that enhances activations near contact regions and mask boundaries, leading to more realistic attachment and geometry.

In practice, each multi-scale image feature is first added to its corresponding encoder skip connection and later concatenated in the decoder. This preserves the expressivity of the base UNet while steering it toward shadow-specific synthesis. Finally, a mask predictor head aggregates these features to predict the object shadow mask, supervised by a weighted binary cross-entropy loss that refines shadow geometry.

\subsection{Text-Grounded Shadow Prompt Conditioning}
\label{sec:text_layout}
While image-based conditioning offers fine-grained spatial guidance by operating in pixel space, it lacks an explicit and scalable object–shadow grouping mechanism. To bridge this gap, we introduce a parallel prompt pathway. Specifically, we iteratively predict shadow bounding boxes, quantize them into discrete layout tokens, and leverage these tokens to guide the cross-attention mechanism towards the appropriate regions. The following subsections provide a detailed explanation of each component.
\subsubsection{Shadow-Box Predictor}
\label{subsec:box_predictor}
We first train a shadow-box predictor to estimate per-object shadow bounding boxes directly from the shadow-free composite and instance masks. The network takes as input the concatenation of the composite and foreground objects mask, and predicts \(O\) shadow boxes \(\hat{B}^{\text{shd}}_o\) (one per object), each in top–left / bottom–right form \((s_1,t_1,s_2,t_2)\).
It is trained independently of the diffusion model using ground-truth boxes \(B^{\text{shd}}_o\), extracted from per-object shadow masks, with a loss that combines coordinate regression and overlap quality:
\begin{equation}
\mathcal{L}_{\text{box}}
=
\sum_{o=1}^{O}
\left\|
\hat{B}^{\text{shd}}_o - B^{\text{shd}}_o
\right\|_{1}
+
\lambda_{\text{IoU}}\!\left(1 - \operatorname{IoU}\!\left(\hat{B}^{\text{shd}}_o, B^{\text{shd}}_o\right)\right)
\end{equation}
After training, we freeze this predictor and use its outputs, \(\hat{B}^{\text{shd}}_o\), to drive text-grounded conditioning for the diffusion model during both training and inference.

\subsubsection{Grounding Shadow Positional Encoding}
\label{sec:text_layout_encoding}
Given the predicted boxes \(\hat{B}^{\text{shd}}_o=(s_1,t_1,s_2,t_2)\), we normalize coordinates to \([0,1]\) by image width \(W\) and height \(H\), then discretize onto a coarse grid of size \(H_g \times W_g\) (e.g., \(H_g=W_g=16\)). Each coordinate is quantized to the nearest bin, yielding indices in \(\{0,\ldots,N_{\text{bins}}-1\}\) with \(N_{\text{bins}}=H_g=W_g\). For each object \(o\), this produces
\begin{equation}
B_{\text{shd}}^{(o)} = \{b^s_1, b^t_1, b^s_2, b^t_2\}
\end{equation}
a compact, grid-aligned descriptor of its shadow layout.

We inject this layout into the text-conditioning stream via learnable shadow positional tokens. For an object with label ``cat'', we form:
\begin{center}
\emph{``cat casting shadow [sx\_bs1][sy\_bt1][sx\_bs2][sy\_bt2]''}.
\end{center}
Here, each token, such as [sx\_bs1] or [sy\_bt2], is chosen from a vocabulary of learnable embeddings specific to shadow coordinates along the horizontal or vertical axis (e.g., [sx\_3] for a shadow x-coordinate in bin 3, [sy\_12] for a shadow y-coordinate in bin 12. These layout tokens are randomly initialized and jointly trained with the diffusion model, allowing the model to learn over time to interpret different bin tokens as distinct positions in the image plane through supervision from the shadow generation task.
 We generate one such phrase per object and concatenate them to form the full prompt, which is then fed to a pre-trained CLIP text encoder. We keep all pre-trained CLIP parameters (transformer layers and original word embeddings) frozen, and learn only the embeddings of the newly introduced positional tokens. Using predicted (rather than ground-truth) boxes ensures training and inference are aligned.

\subsubsection{Attention Alignment for Shadow Tokens}
\label{sec:attn_align}
To ensure that the model makes meaningful and spatially coherent use of the shadow positional tokens, we introduce an attention alignment loss. Intuitively, we want the cross-attention associated with an object shadow box positional tokens to focus on that object shadow region, rather than diffuse over unrelated areas of the image.

Let \(\mathcal{L}_{\text{ca}}\) denote a set of cross-attention layers in the UNet; in practice, we use middle and late-stage cross-attention blocks. For a given layer \(\ell \in \mathcal{L}_{\text{ca}}\), the cross-attention weights can be written as
\begin{equation}
A^{(\ell)} \in \mathbb{R}^{N_\ell \times L}
\end{equation}
where \(N_\ell\) is the number of spatial positions in that layer and \(L\) is the length of the text sequence. Each column \(A^{(\ell)}_{:, j}\) gives an attention map over spatial positions for text token \(j\).

For each object \(o \in \{1, \dots, O\}\), we collect the indices of its shadow box tokens \(\mathcal{J}^{\text{shd}}_o\). We then aggregate their attention maps at layer \(\ell\) by averaging across the corresponding columns:
\begin{equation}
A^{(\ell),\text{shd}}_o = \frac{1}{|\mathcal{J}^{\text{shd}}_o|} \sum_{j \in \mathcal{J}^{\text{shd}}_o} A^{(\ell)}_{:, j}
\end{equation}
We reshape these vectors back to spatial maps of size \(H_\ell \times W_\ell\) and normalize them to sum to one.

As supervision targets, we downsample the ground-truth shadow mask for object \(o\) to the same spatial resolution, obtaining \(T^{(\ell),\text{shd}}_o\), which we also normalize to form a probability map. The attention alignment loss is then defined as
\begin{equation}
\mathcal{L}_{\text{align}} 
= \sum_{\ell \in \mathcal{L}_{\text{ca}}} \sum_{o=1}^{O} 
\operatorname{KL}\big( A^{(\ell),\text{shd}}_o \,\|\, T^{(\ell),\text{shd}}_o \big)
\end{equation}
where \(\operatorname{KL}(\cdot \,\|\, \cdot)\) denotes the Kullback--Leibler divergence applied over spatial positions.

This loss encourages the attention associated with the discrete shadow box tokens to align spatially with the true shadow regions, reinforcing a grounded interpretation of layout. As a result, the diffusion model learns to associate each shadow positional token not only with a semantic concept (e.g., ``cat'', ``shadow'') but also with a consistent, object-specific spatial footprint in the image plane.

\subsection{Training Objective}
\label{subsec:training_objective}

The diffusion model is trained with the standard noise-prediction loss on \(I_g\), augmented with the auxiliary losses already used in our image pathway. The total objective is
\begin{equation}
    \mathcal{L}
    =
    \mathcal{L}_{\text{diff}}
    +
    \lambda_{\text{mask}} \mathcal{L}_{\text{mask}}
    +
    \lambda_{\text{bg}} \mathcal{L}_{\text{bg}}
    +
    \lambda_{\text{align}} \mathcal{L}_{\text{align}}
\end{equation}
where \(\mathcal{L}_{\text{diff}}\) is the denoising loss, \(\mathcal{L}_{\text{mask}}\) supervises the predicted shadows mask, \(\mathcal{L}_{\text{bg}}\) preserves background regions outside shadows, and \(\mathcal{L}_{\text{align}}\) is the attention alignment loss defined above.

\section{Experiments}
\subsection{Datasets and Evaluation Metrics}
\renewcommand{\arraystretch}{1.1}

To support training in multi-object scenarios, we extend DESOBAv2~\cite{liu2023desobav2} with scenes containing multiple inserted objects. Starting from shadowed images, we run the shadow detection~\cite{wang2022instance} and inpainting~\cite{SD-XL_Inpainting_0.1} pipeline to obtain a clean, shadow-free reference image per scene. We then synthesize multi-object composites by selectively removing shadows only for the target foreground objects, while keeping shadows of all non-target objects. This approach avoids shadow-edge artifacts and preserves realistic context. In total, the training set comprises 38,755 images, including 11,037 additional multi-object composites. The multi-object test set contains 320 images with 1,876 object–shadow tuples. For single-object comparison, we retain the original Desobav2 test splits. To ensure a fair evaluation, we partition the data at the image level to prevent any leakage of background or foreground information across splits. For text conditioning, we derive simple template prompts that will later host the shadow positional tokens. Using the foreground instance masks, we draw a tight, distinct-color bounding box around each object on the original image, as shown in Fig.~\ref{fig:ViP-LLaVA}, and then feed this annotated image to ViP-LLaVA~\cite{cai2024vip}. From the generated descriptions, we extract a representative category word for each object and insert it into our prompt template: ``$<\text{object category}> \text{ casting shadow } [\text{shadow-box tokens}]$''. When ViP-LLaVA identifies interaction phrases (e.g., girl riding a motorbike), we retain the associated action verb to enrich the object description.

To quantitatively evaluate the quality of the generated shadows, we used established metrics from prior works~\cite{liu2024shadow,zhao2025shadow}, which assess both image quality and the accuracy of the predicted shadow masks.  We report all metrics separately for BOS (background object-shadow) and BOS-free subsets to highlight performance with and without background shadow context. Image quality is measured using Root Mean Square Error (RMSE) and Structural Similarity (SSIM), computed both globally (GR, GS) over the entire image and locally (LR, LS) within the ground-truth shadow regions. Shadow mask quality is measured using the Balanced Error Rate (BER), which is evaluated both globally (GB) and locally (LB). Following previous work~\cite{liu2024shadow,zhao2025shadow}, we generate 5 samples per test image with different seeds and report the one closest to the ground truth.

\subsection{Implementation Details}
Our model is implemented in PyTorch and built upon the Stable Diffusion v1.5 backbone. All experiments are conducted at a spatial resolution of 512$\times$512. We use the AdamW optimizer with a fixed learning rate of $1 \times 10^{-5}$ and a batch size of 4, training for 80 epochs on a single NVIDIA RTX 4090 GPU. In addition to the core framework, we train a shadow-box predictor network responsible for estimating object-specific shadow locations. This module is trained independently using the same image resolution and the AdamW optimizer, but with a learning rate of $1 \times 10^{-4}$ and a batch size of 8. The training continues for up to 100 epochs, using only the training split of our dataset.

\newcommand{\ansone}[1]{\textcolor{red}{#1}}
\newcommand{\anstwo}[1]{\textcolor{green!60!black}{#1}}
\newcommand{\ansthree}[1]{\textcolor{orange!80!black}{#1}} % optional 

\begin{figure}[t]
\centering
\newcommand{\imgscale}{0.32}

\includegraphics[width=\imgscale\linewidth]{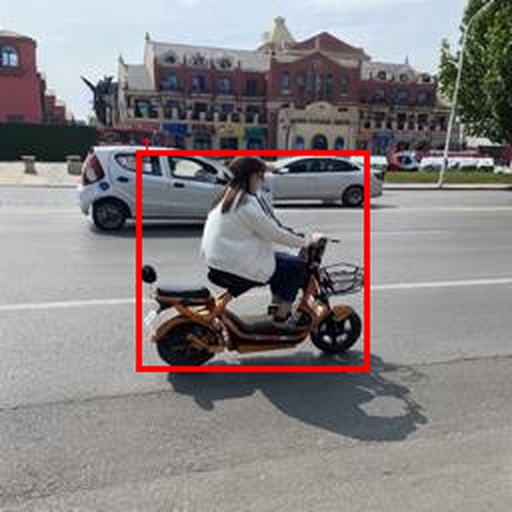}\hfill
\includegraphics[width=\imgscale\linewidth]{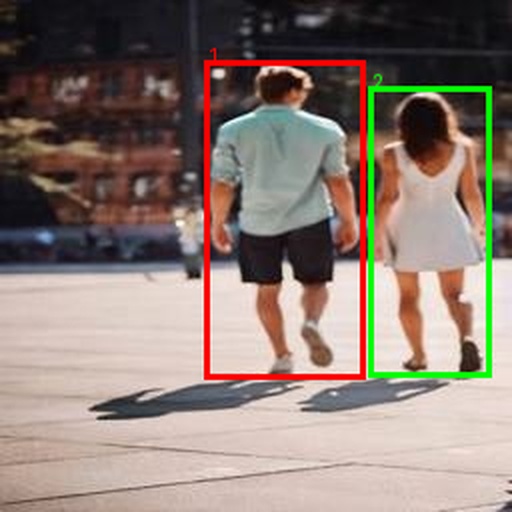}\hfill
\includegraphics[width=\imgscale\linewidth]{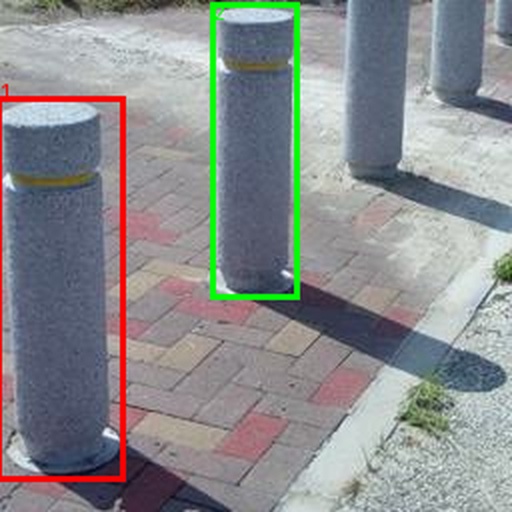}

\vspace{1mm}
{\tiny
\textbf{Left:} \ansone{Girl riding a motorbike.} \hfill
\textbf{Center:} \ansone{Man in blue shirt.} / \anstwo{Woman in white dress} \hfill
\textbf{Right:} \ansone{The pole.} \anstwo{Second pole from left.} \\
}

\caption{\noindent{ViP-LLaVA object naming from bounding-box prompts.}
Given the prompt ``Name the objects in the bounding boxes.'',
the model generates: (Left: ``Girl riding a motorbike''),
(Center: ``Man in blue shirt.'' and ``Woman in white dress.''),
(Right: ``The pole.'' and ``Second pole from the left.'').}
\label{fig:ViP-LLaVA}
\vspace{-2mm}
\end{figure}

\begin{figure*}[t]
\centering

\setlength{\tabcolsep}{1.0pt}        
\renewcommand{\arraystretch}{0}       

\newcommand{\colw}{0.105\textwidth}

\newcommand{\hdr}[1]{%
  \parbox[c][3.6mm][c]{\colw}{\centering\fontsize{9}{9}\selectfont #1}%
}
\newcommand{\im}[1]{\includegraphics[width=\colw]{#1}}

\newcommand{\hgap}{0.6mm}
\newcommand{\rgap}{0.2mm}
% -----------------------------------------------

\begin{tabular}{@{}ccccccccc@{}}
\hdr{Composite} &
\hdr{Objects Mask} &
\hdr{SGRNet} &
\hdr{DAMASNet} &
\hdr{SGDiffusion} &
\hdr{GPSDiffusion} &
\hdr{MetaShadow} &
\hdr{Ours} &
\hdr{Ground Truth} \\[\hgap]

\im{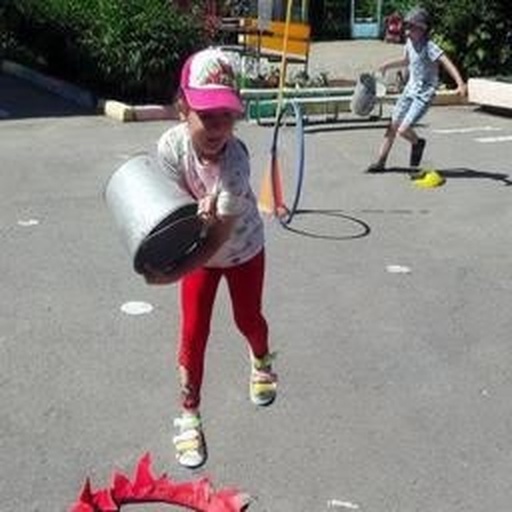} &
\im{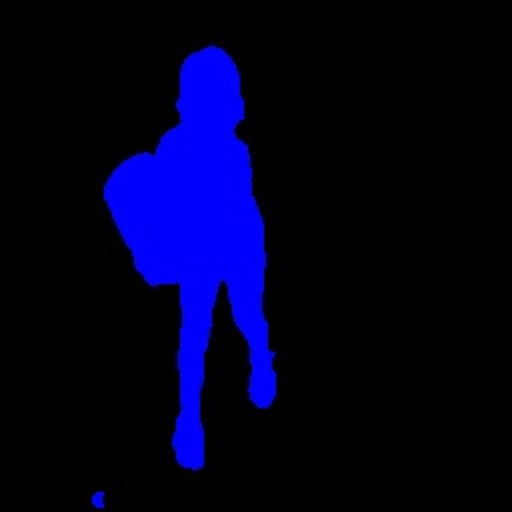} &
\im{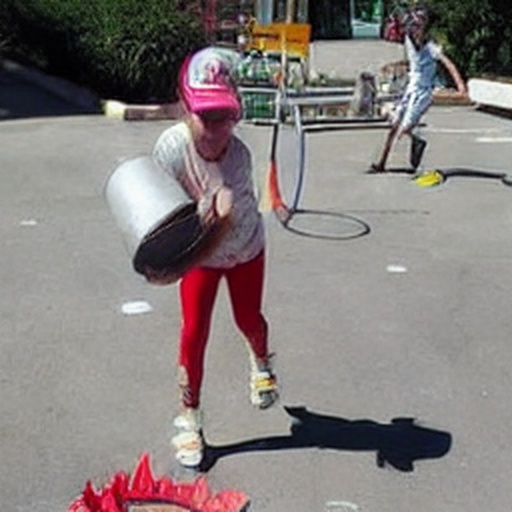} &
\im{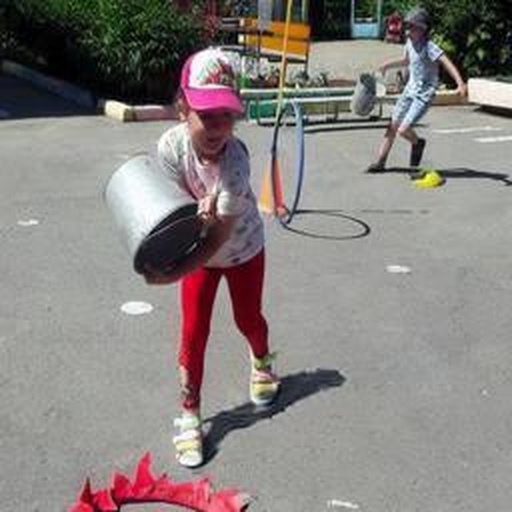} &
\im{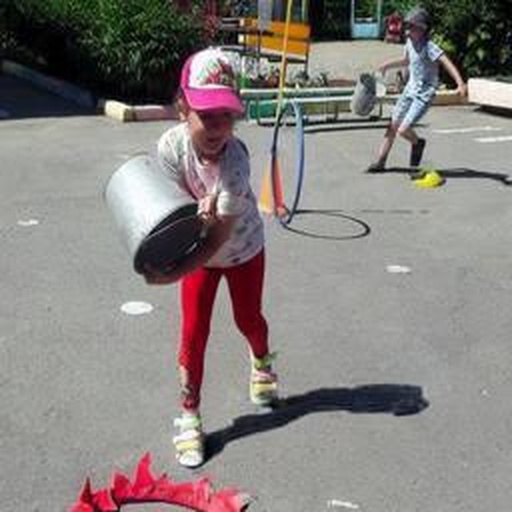} &
\im{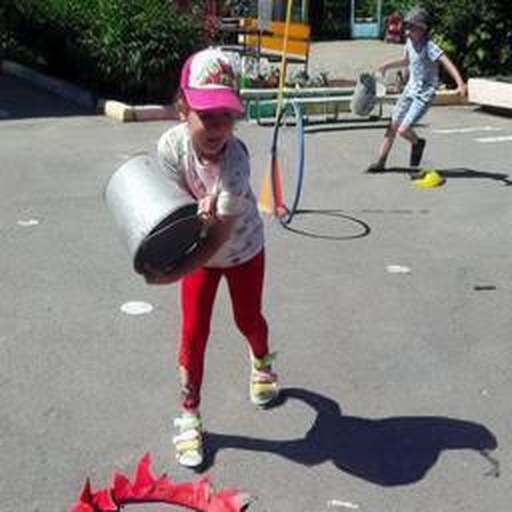} &
\im{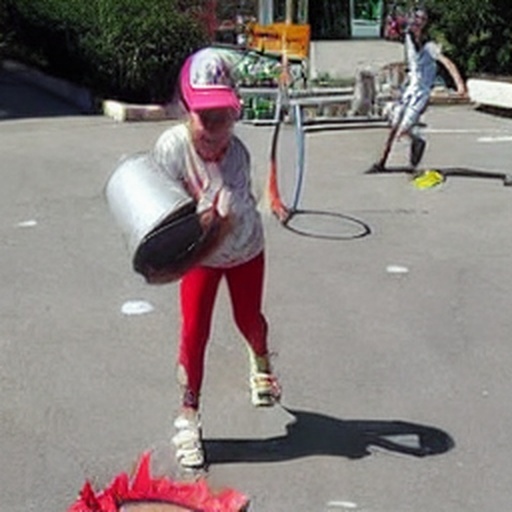} &      
\im{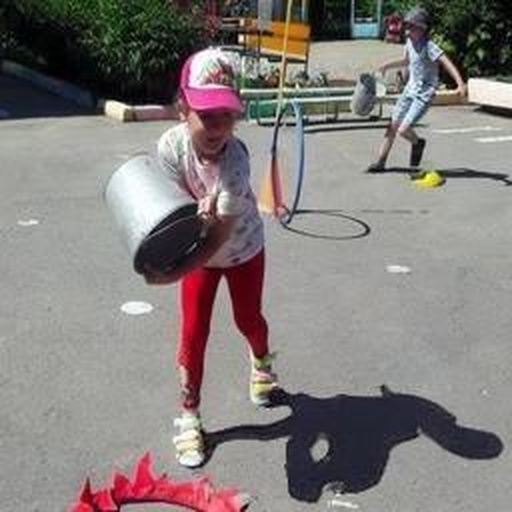} &
\im{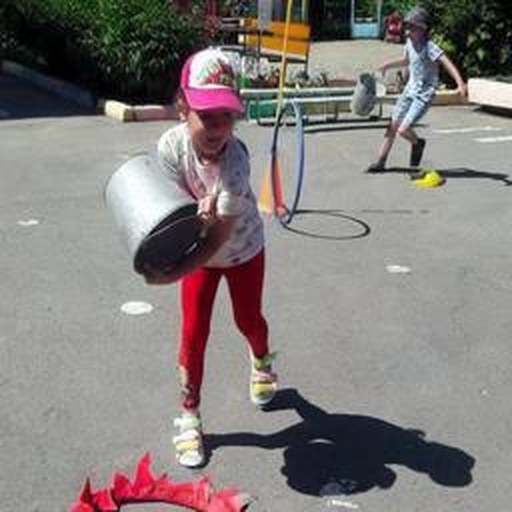} \\[\rgap]

\im{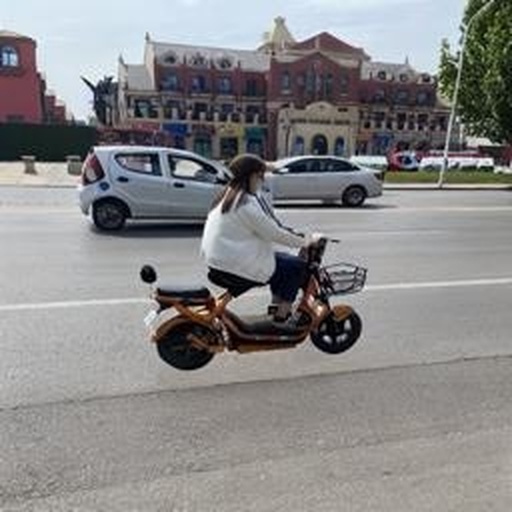} &
\im{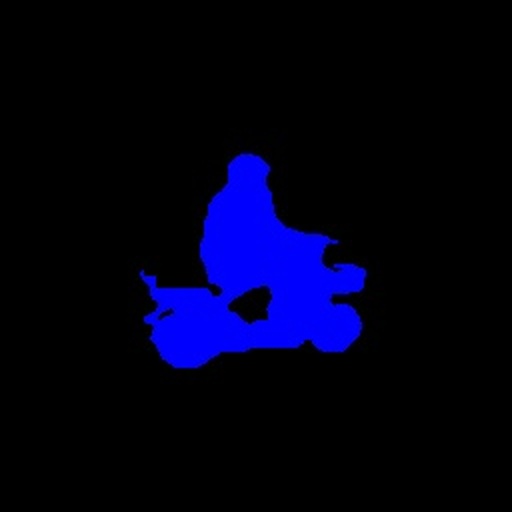} &
\im{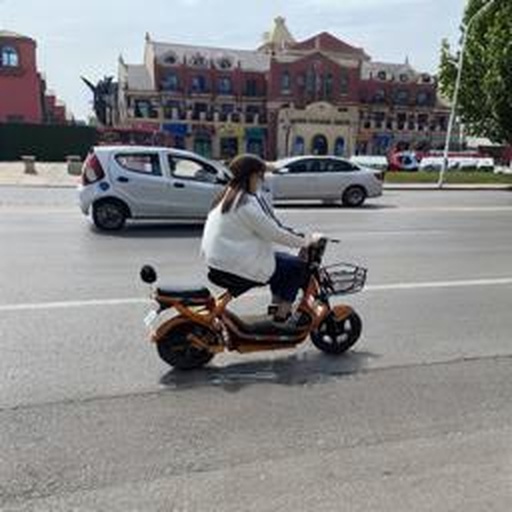} &
\im{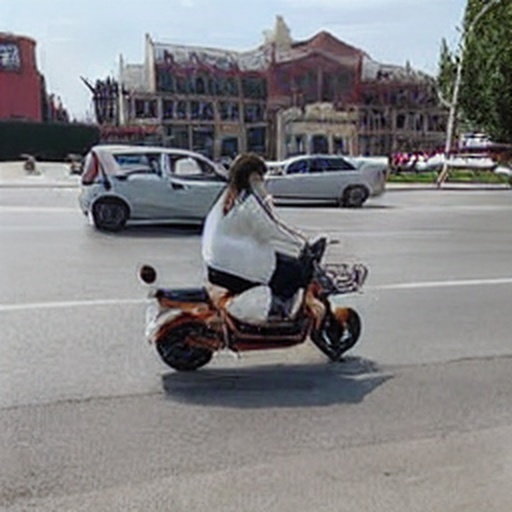} &
\im{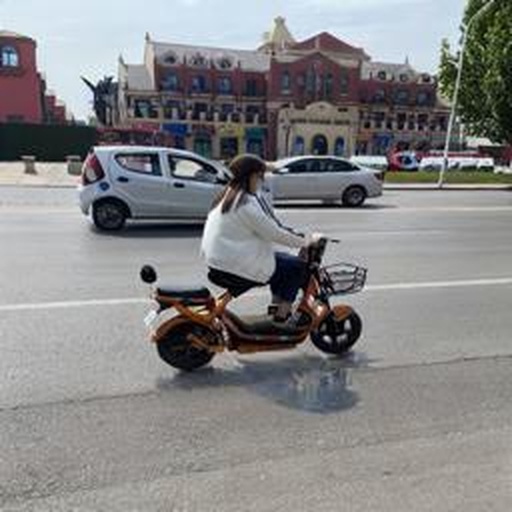} &
\im{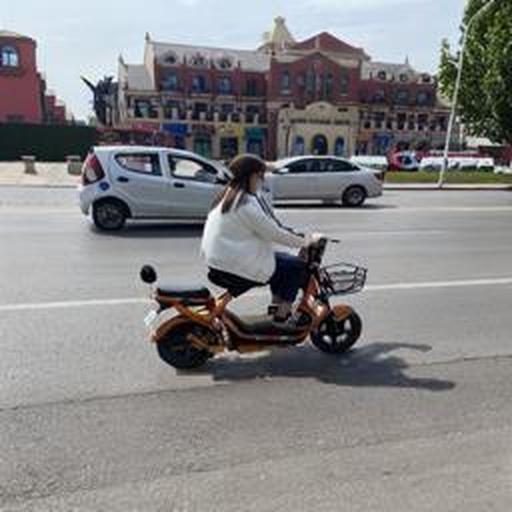} &
\im{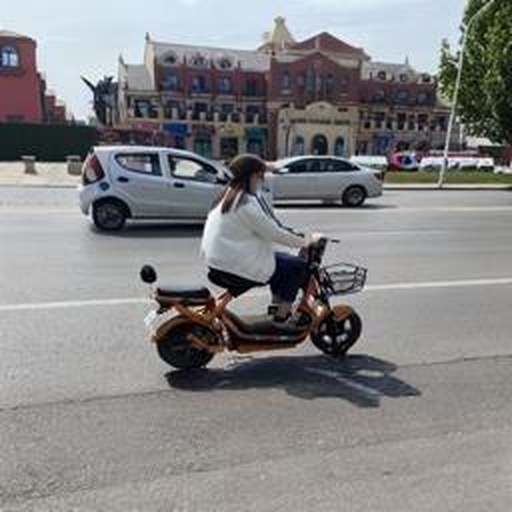} &      
\im{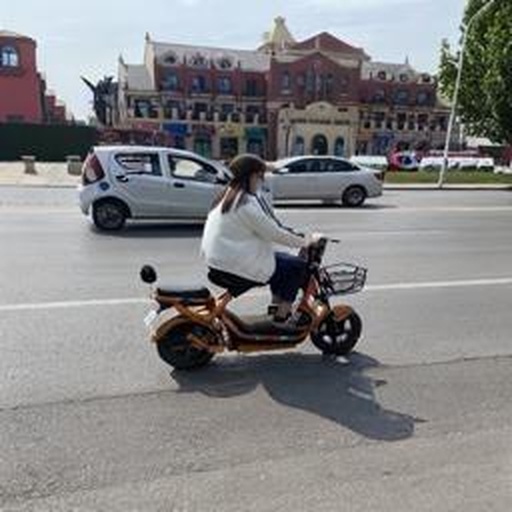} &
\im{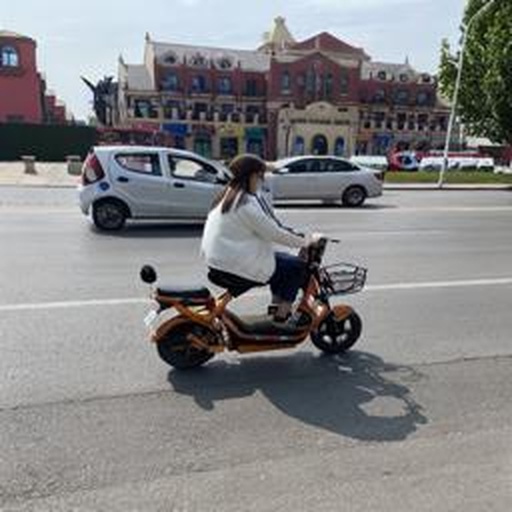} \\[\rgap]

\im{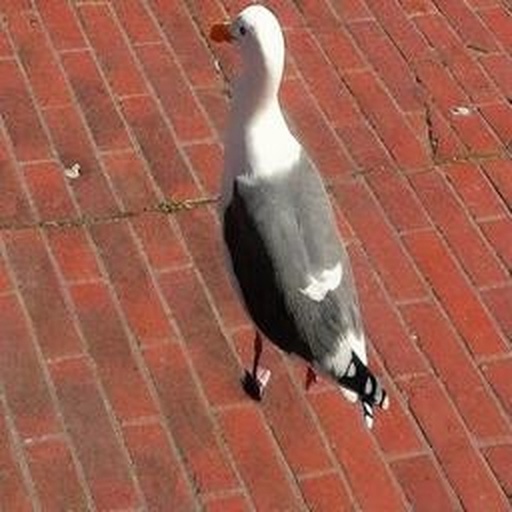} &
\im{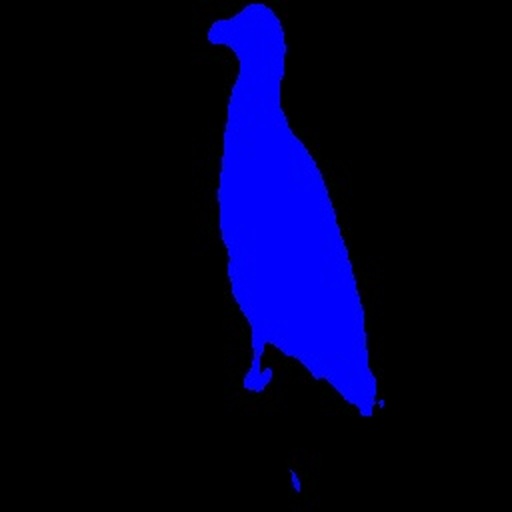} &
\im{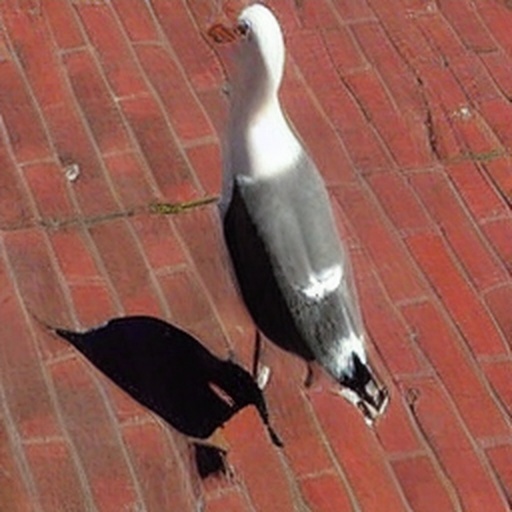} &
\im{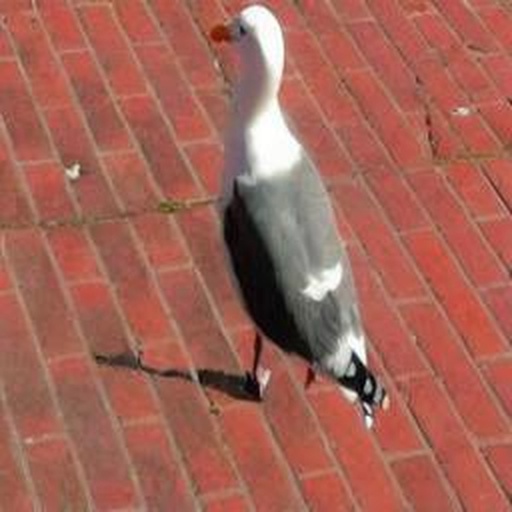} &
\im{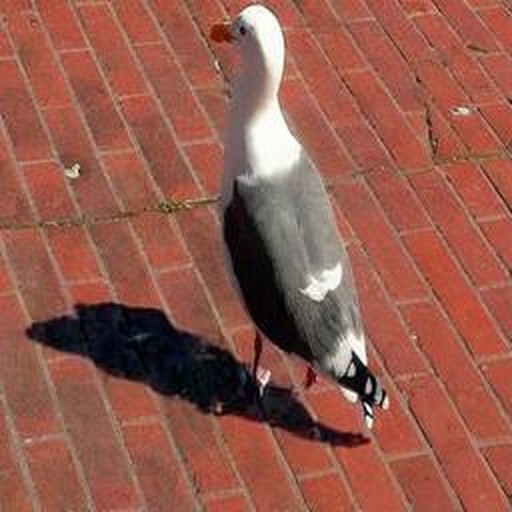} &
\im{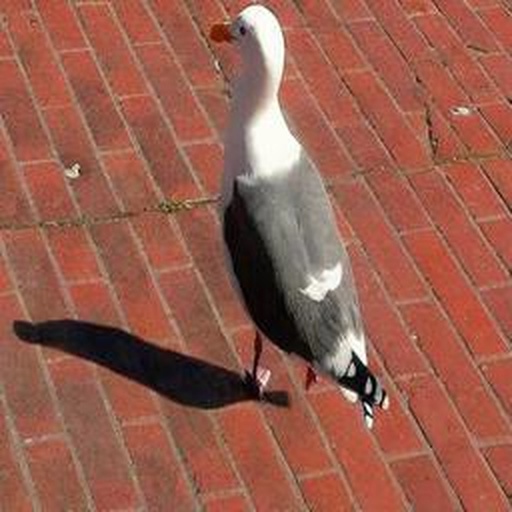} &
\im{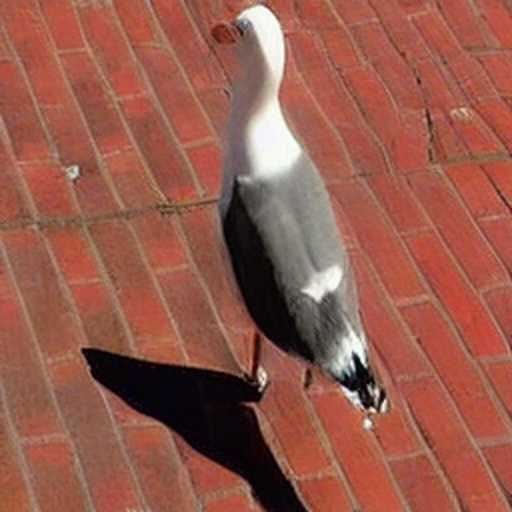} &      
\im{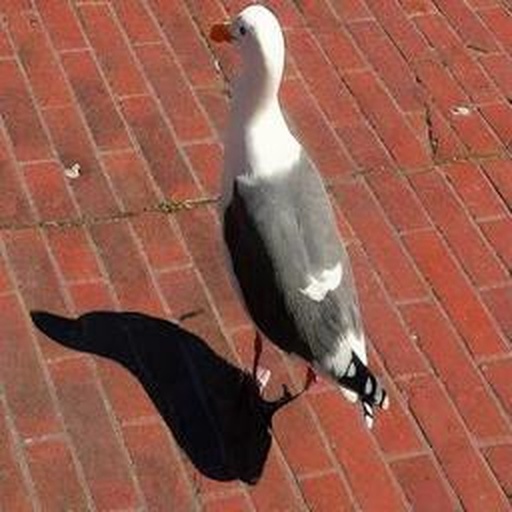} &
\im{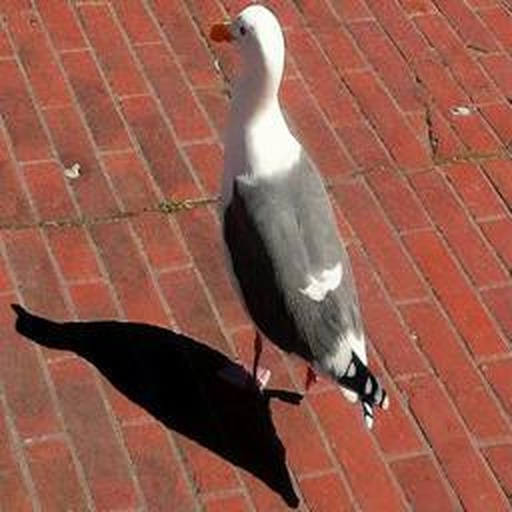} \\[\rgap]

\im{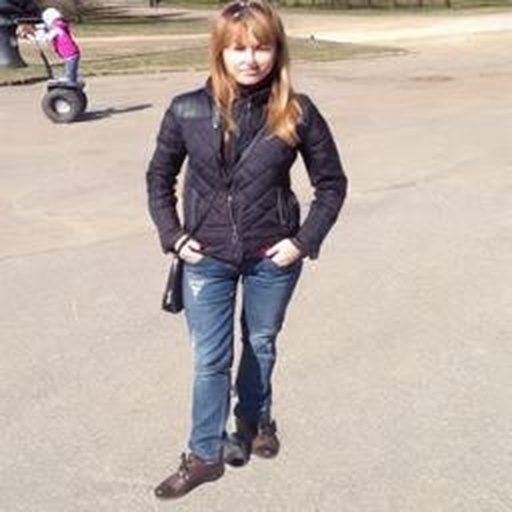} &
\im{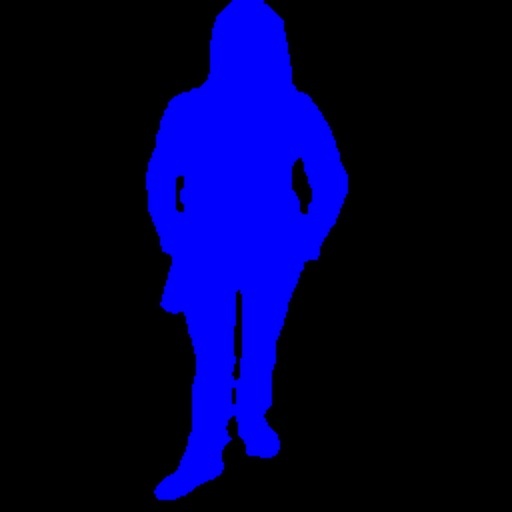} &
\im{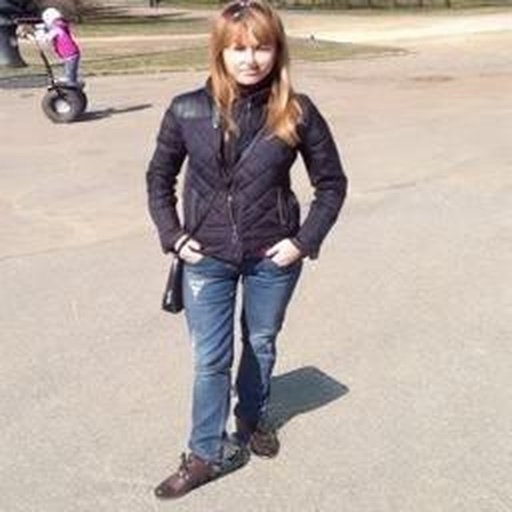} &
\im{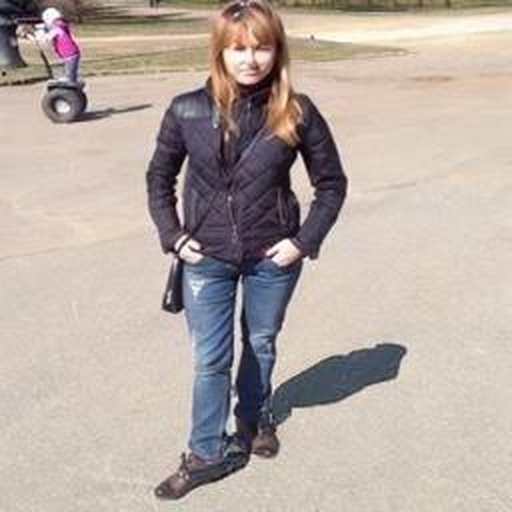} &
\im{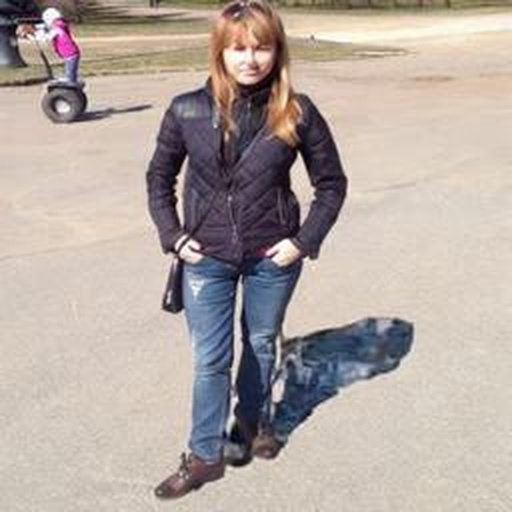} &
\im{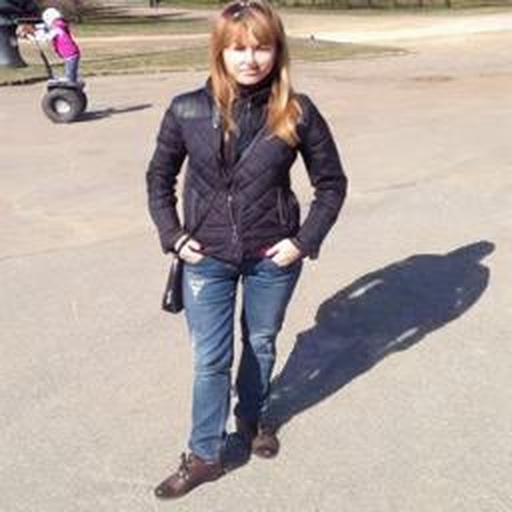} &
\im{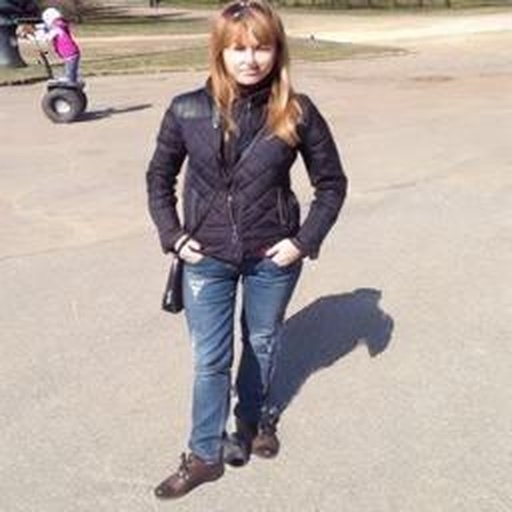} &      
\im{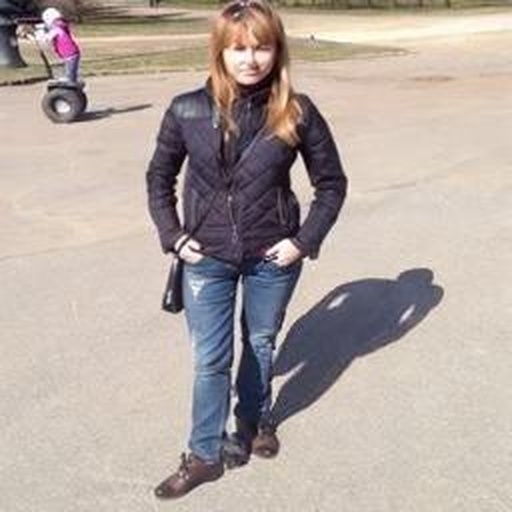} &
\im{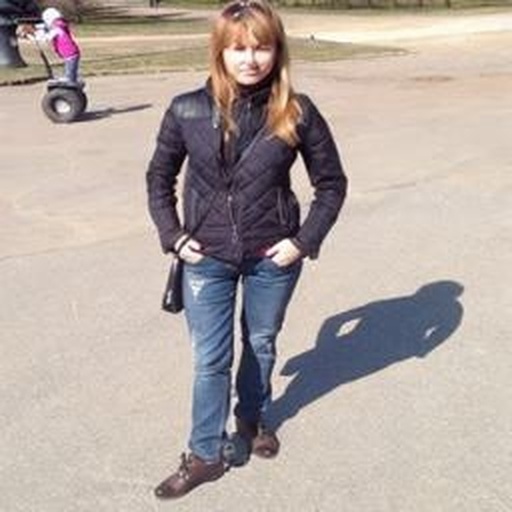} \\[\rgap]

\im{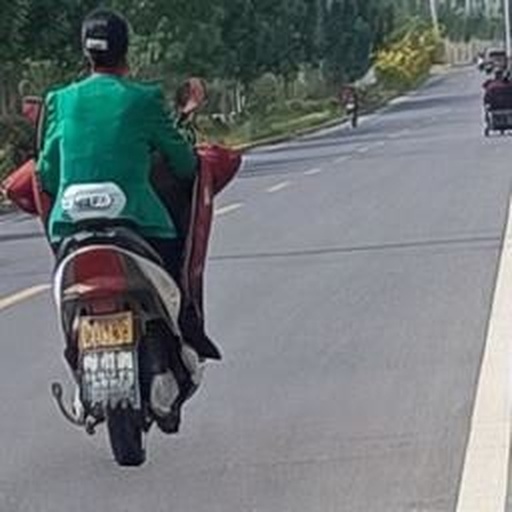} &
\im{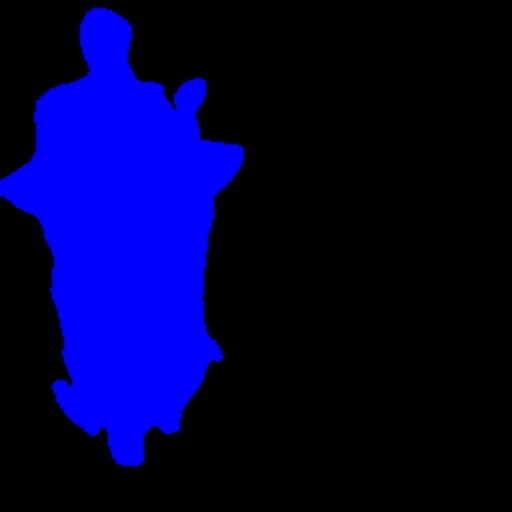} &
\im{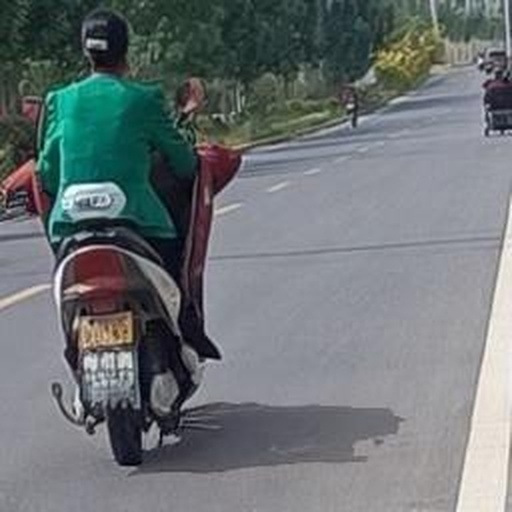} &
\im{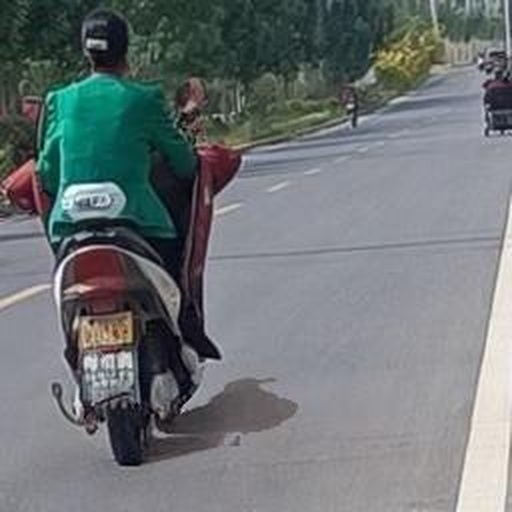} &
\im{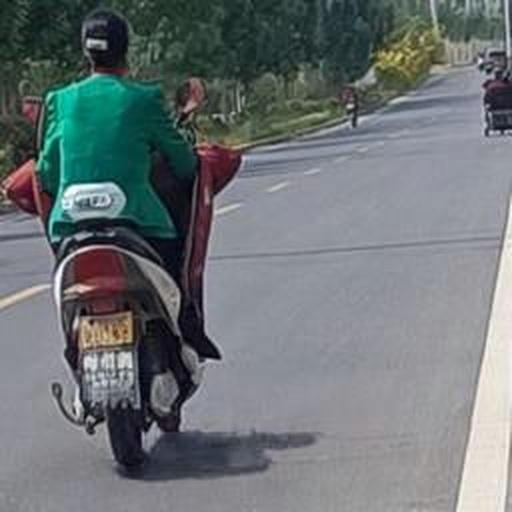} &
\im{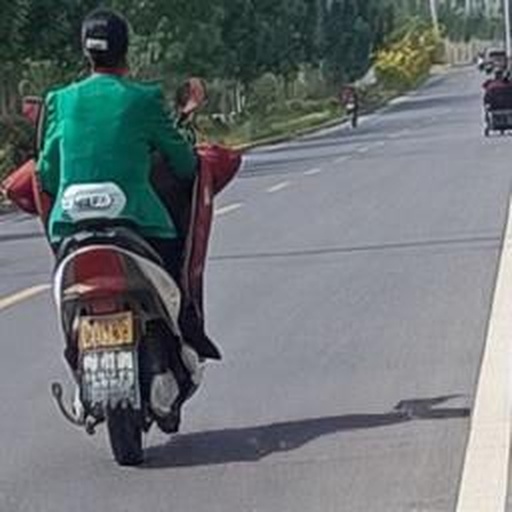} &
\im{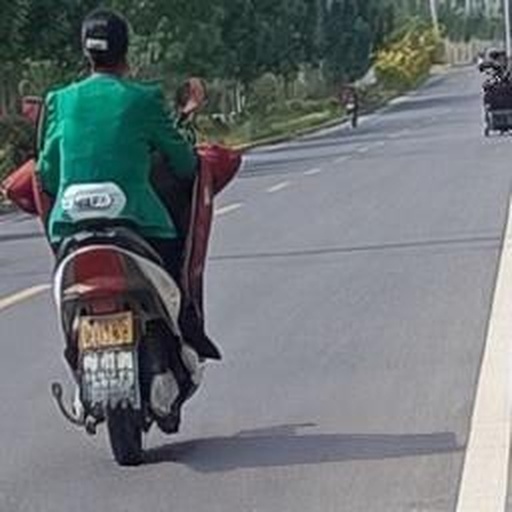} &      
\im{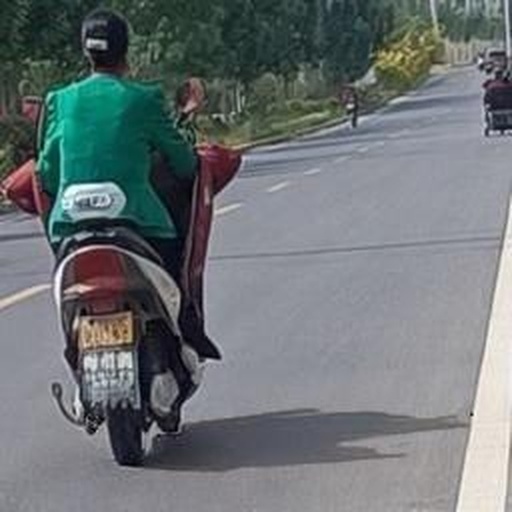} &
\im{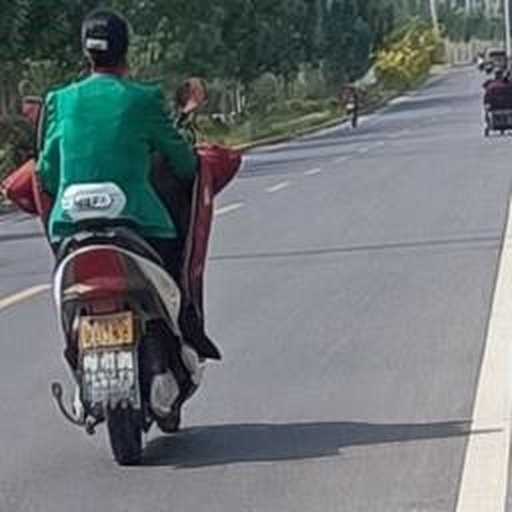} \\[\rgap]

\im{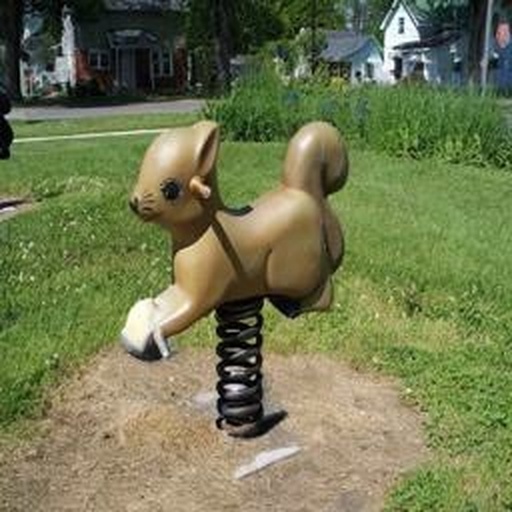} &
\im{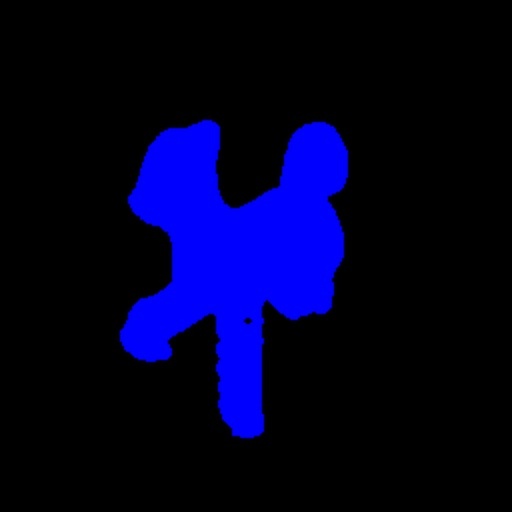} &
\im{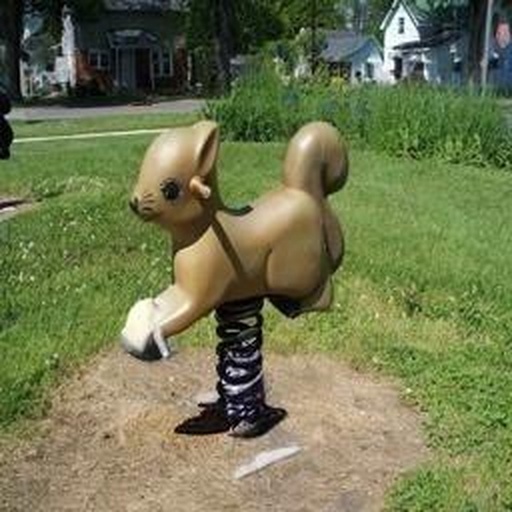} &
\im{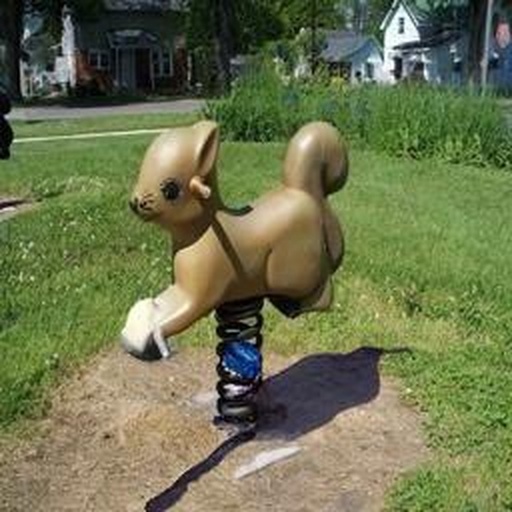} &
\im{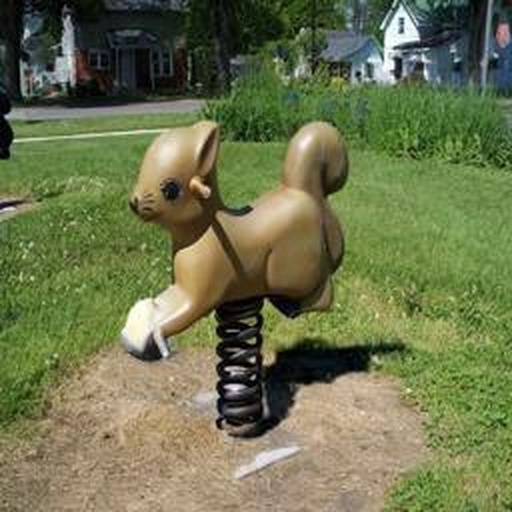} &
\im{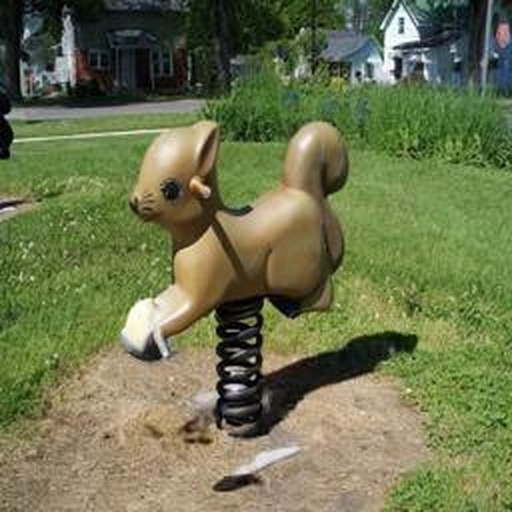} &
\im{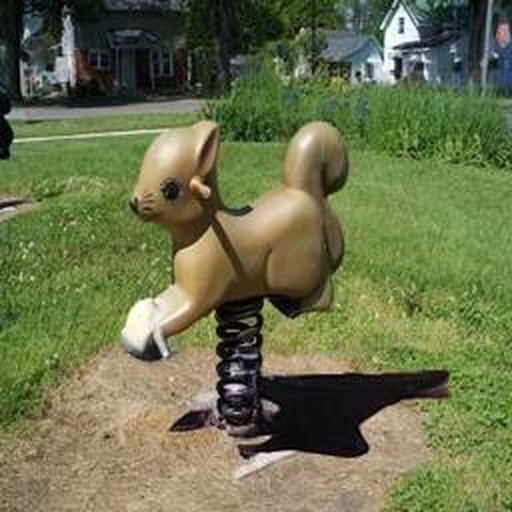} &      
\im{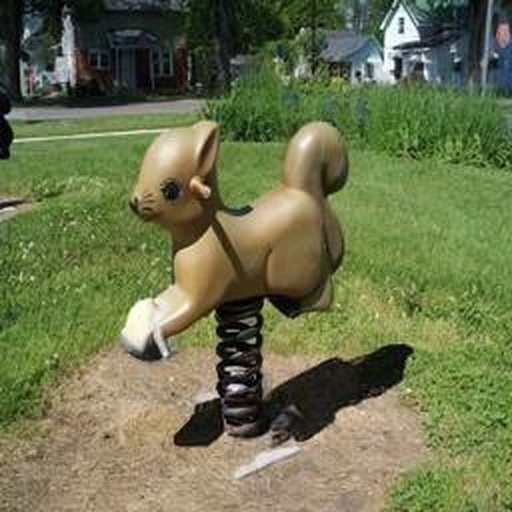} &
\im{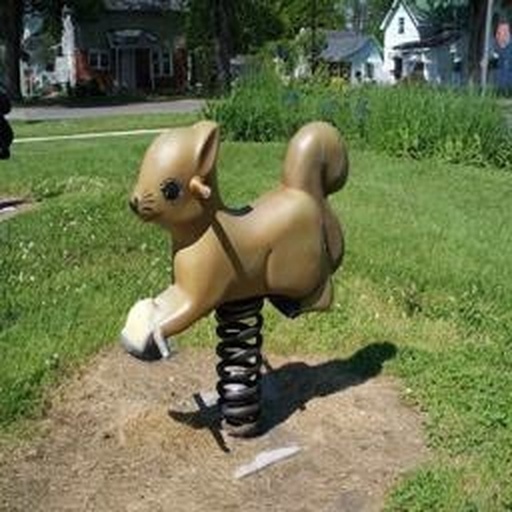} \\[\rgap]
\end{tabular}

\vspace{-1mm}
\caption{Visual comparison with state-of-the-art baseline methods for single object shadow generation. Our method (MultiShadow), given the same image inputs plus a compact text prompt of category terms and shadow positional tokens, produces better shadows for all objects (e.g., Row~1: ``a woman casting shadow [sx\_8][sy\_9][sx\_3][sy\_8]''). Positional tokens are inserted automatically; see Sec.~\ref{sec:text_layout}.}
\label{fig:single1}
\end{figure*}

\begin{figure*}[t]
\centering

\setlength{\tabcolsep}{0.4pt}        
\renewcommand{\arraystretch}{0}       

\newcommand{\colw}{0.105\textwidth}

\newcommand{\hdr}[1]{%
  \parbox[c][3.6mm][c]{\colw}{\centering\fontsize{9}{9}\selectfont #1}%
}
\newcommand{\im}[1]{\includegraphics[width=\colw]{#1}}

\newcommand{\hgap}{0.6mm}
\newcommand{\rgap}{0.2mm}

\begin{tabular}{@{}ccccccccc@{}}
\hdr{Composite} &
\hdr{Object Mask} &
\hdr{SGRNet} &
\hdr{DAMASNet} &
\hdr{SGDiffusion} &
\hdr{GPSDiffusion} &
\hdr{MetaShadow} &
\hdr{Ours} &
\hdr{GT} \\[\hgap]

\im{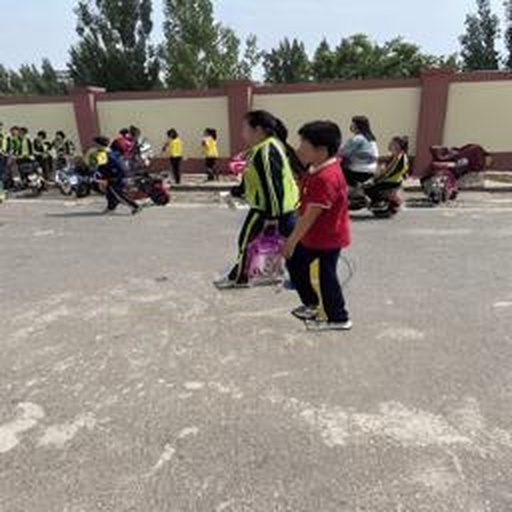} &
\im{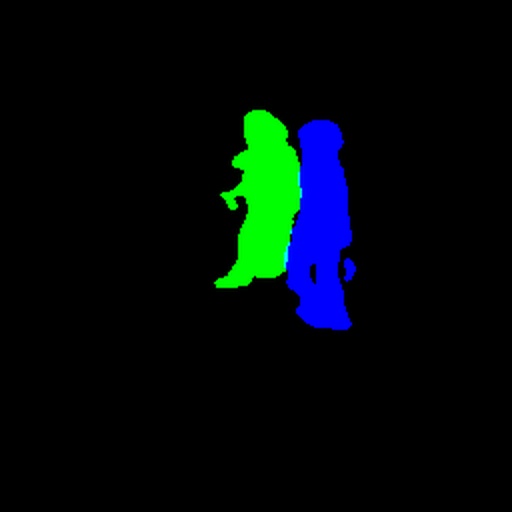} &
\im{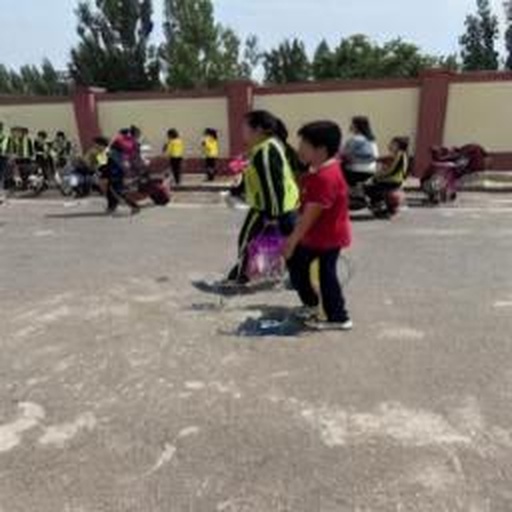} &
\im{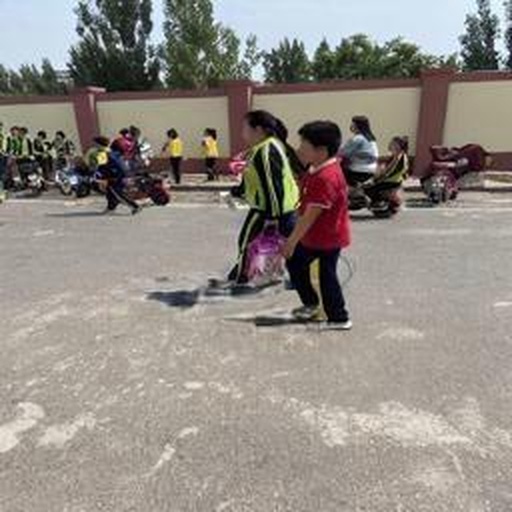} &
\im{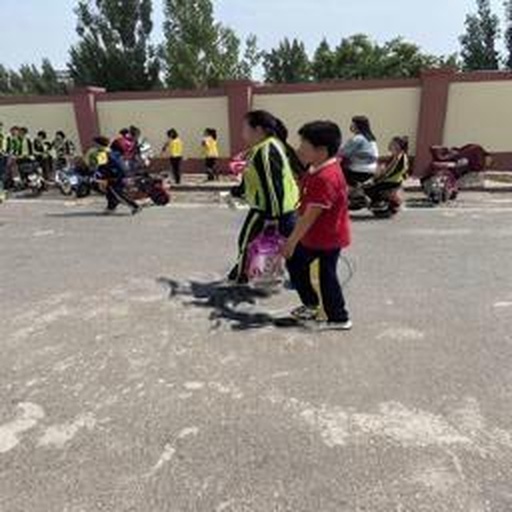} &
\im{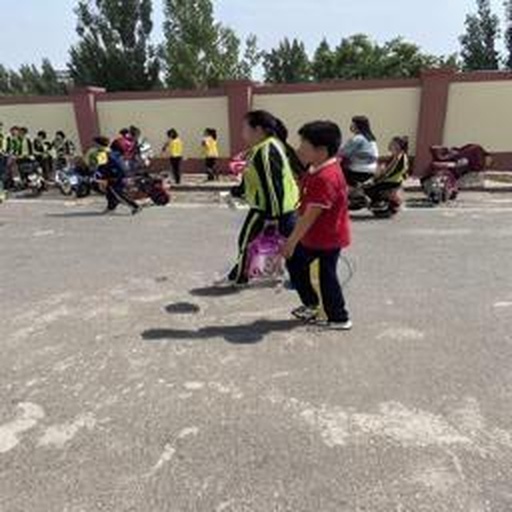} &
\im{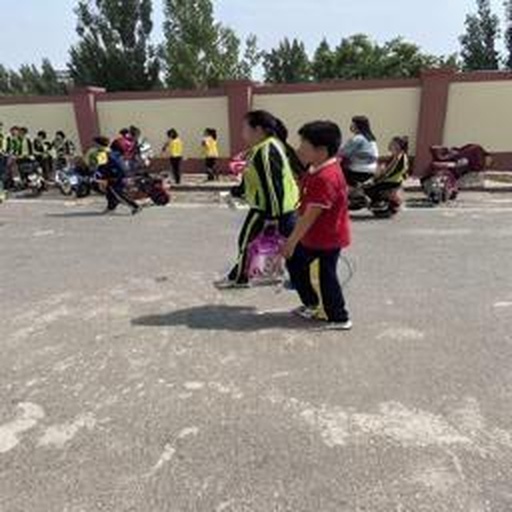} &   
\im{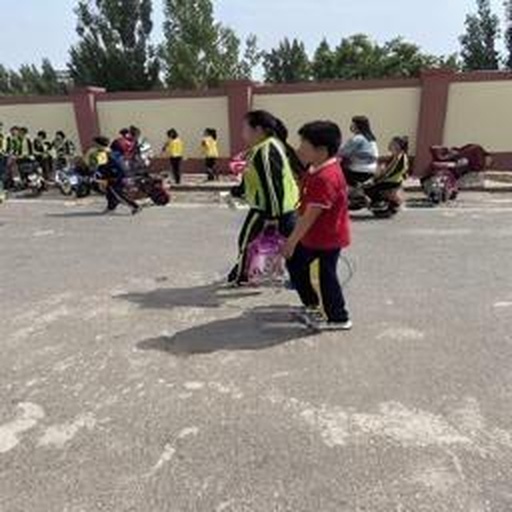} &
\im{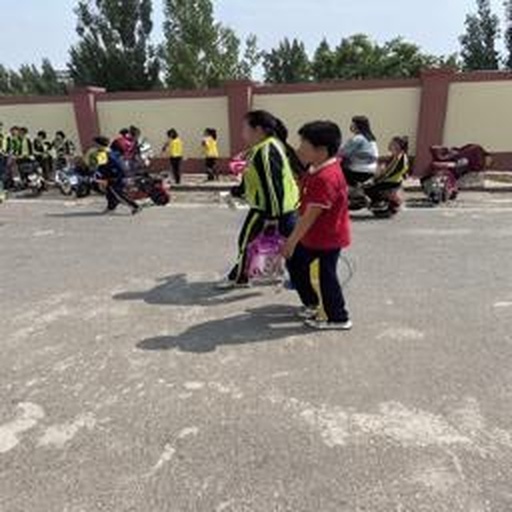} \\[\rgap]

\im{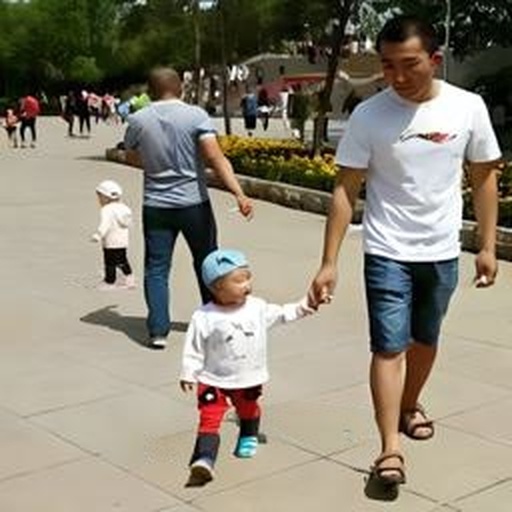} &
\im{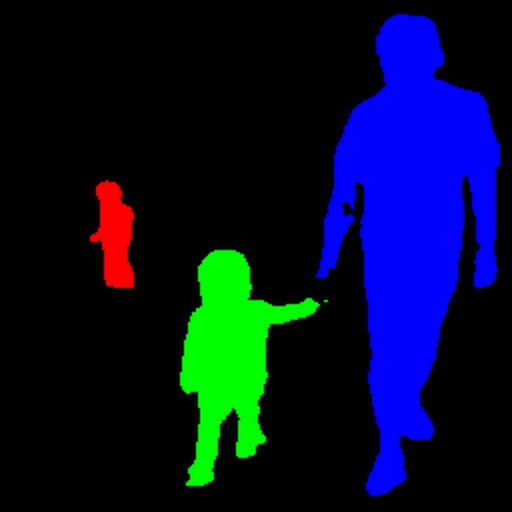} &
\im{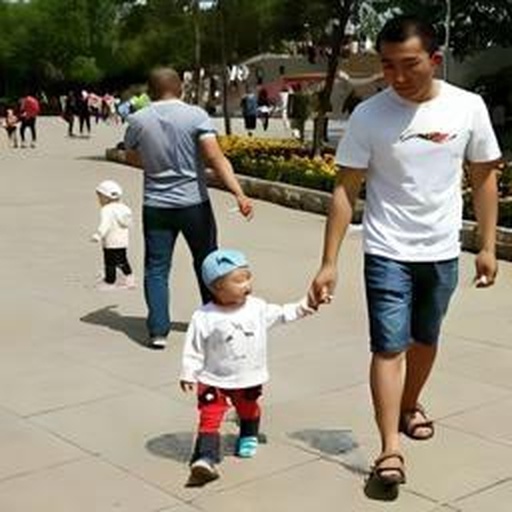} &
\im{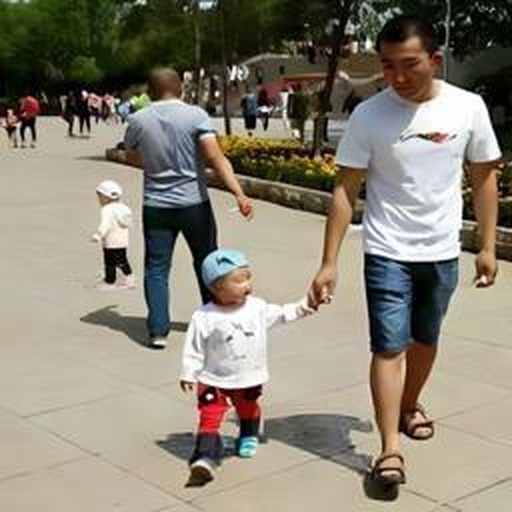} &
\im{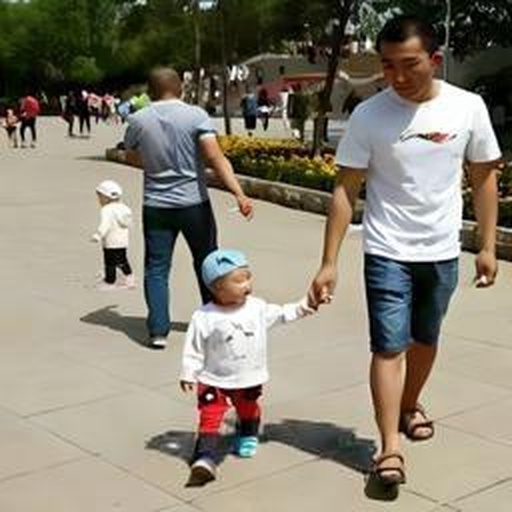} &
\im{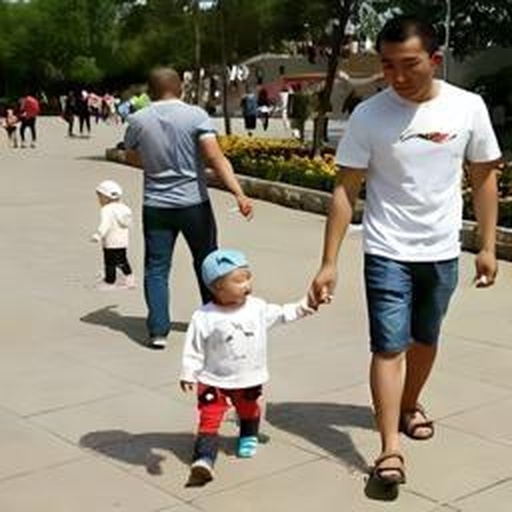} &
\im{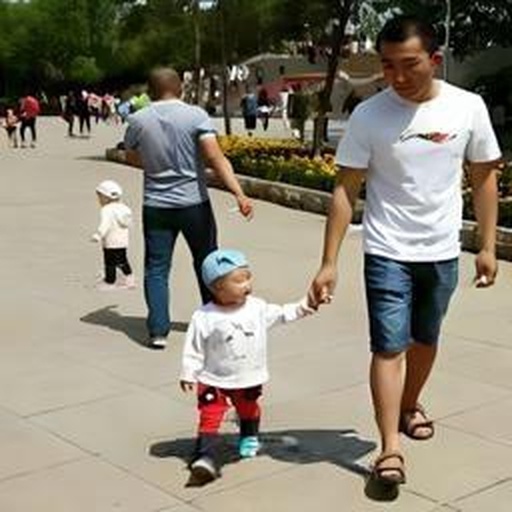} &   
\im{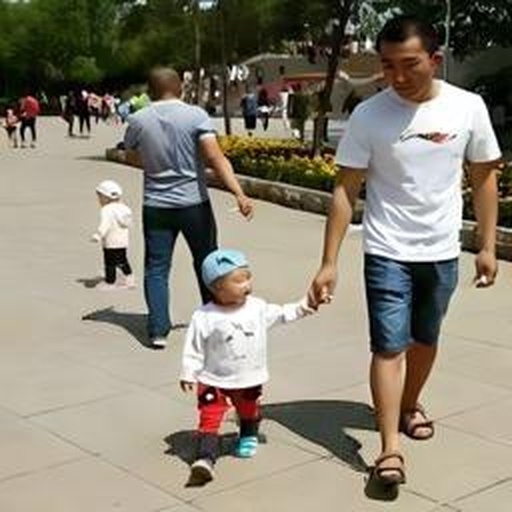} &
\im{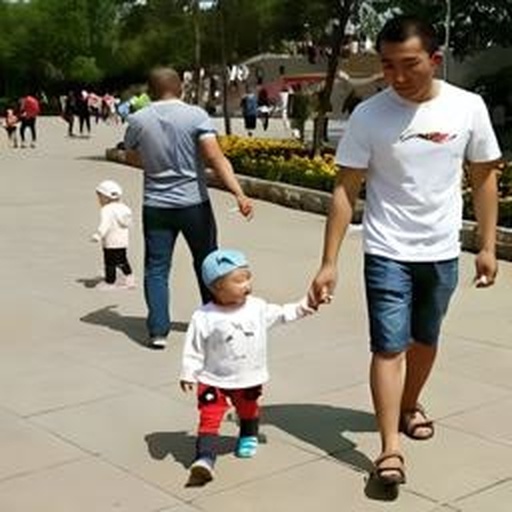} \\[\rgap]

\im{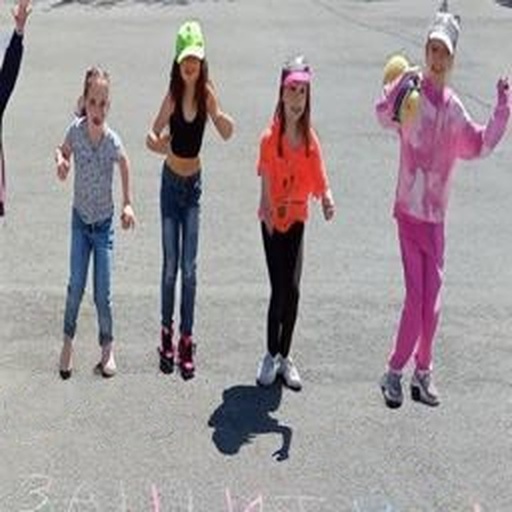} &
\im{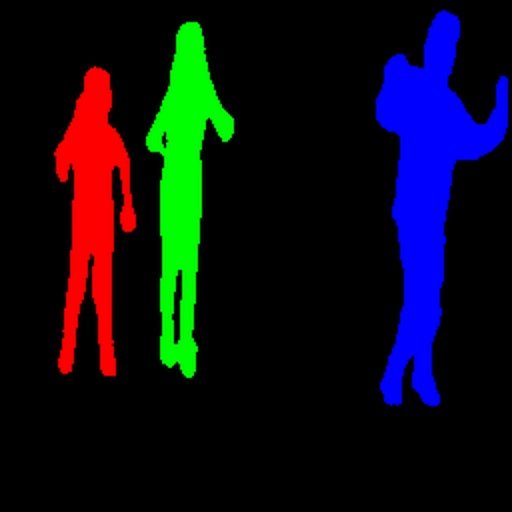} &
\im{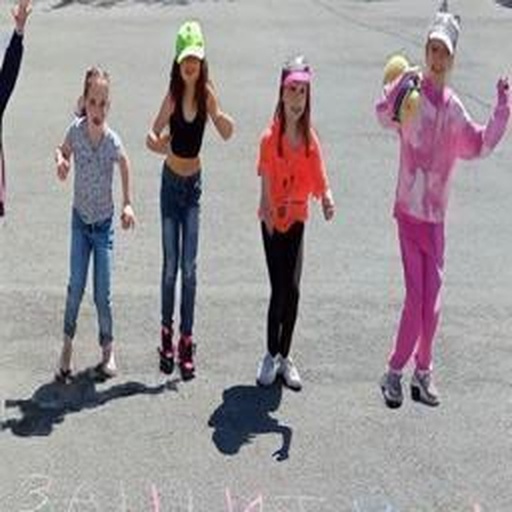} &
\im{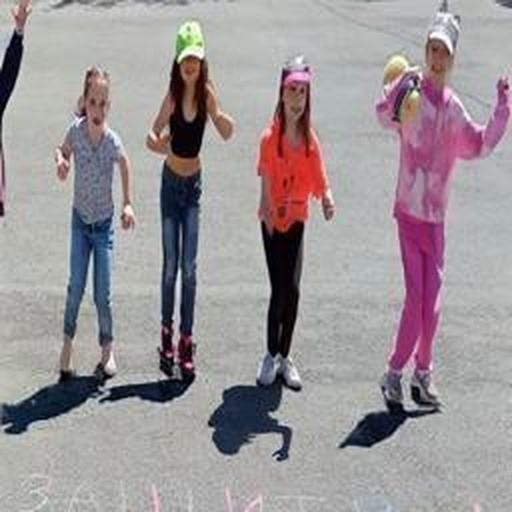} &
\im{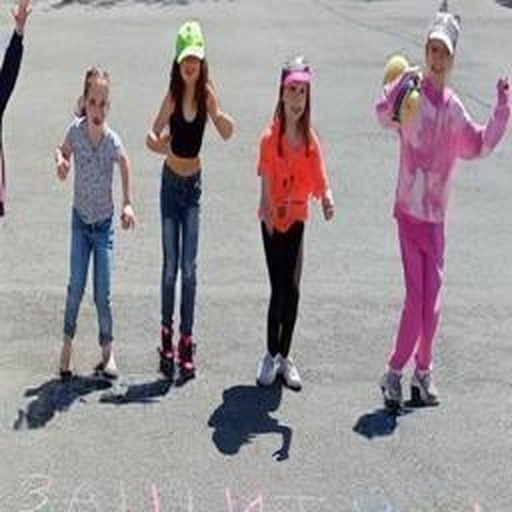} &
\im{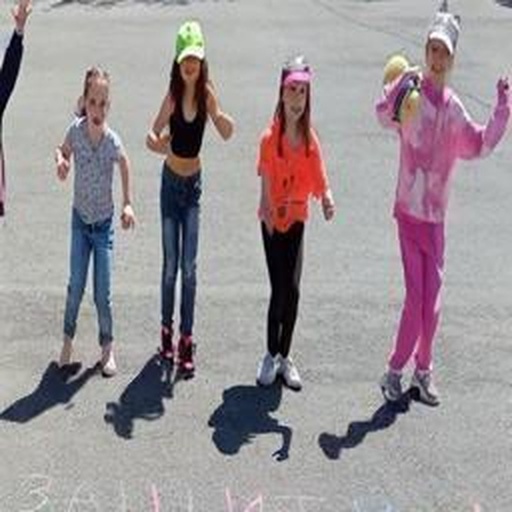} &
\im{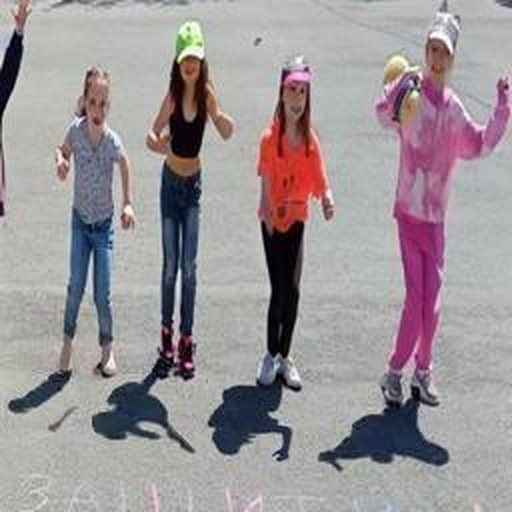} &   
\im{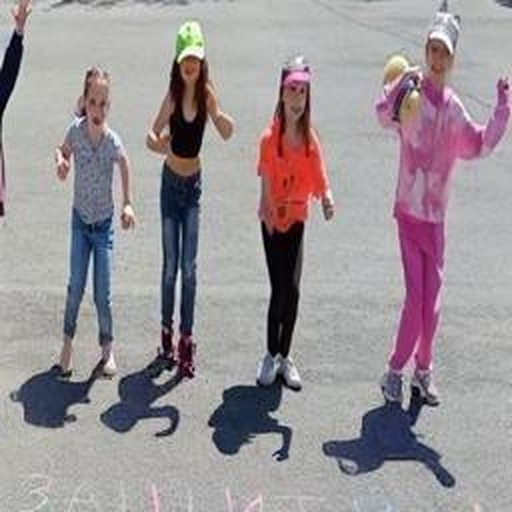} &
\im{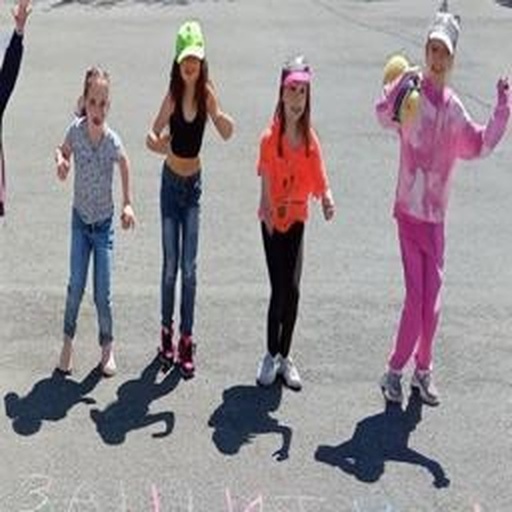} \\[\rgap]

\im{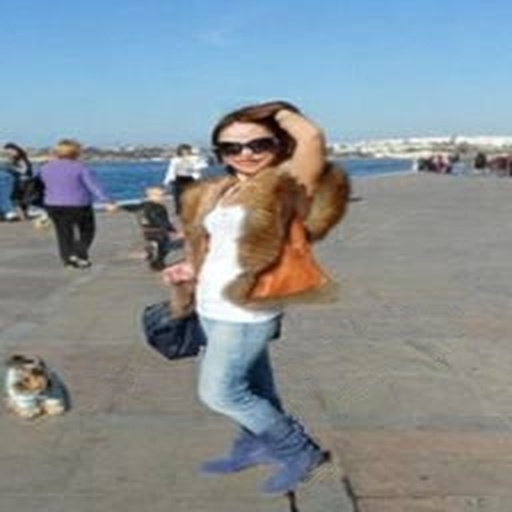} &
\im{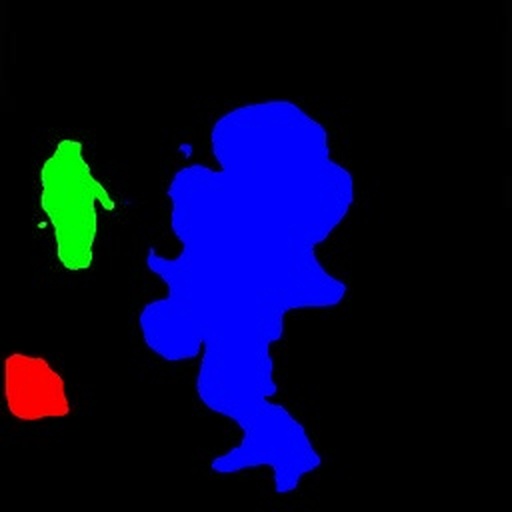} &
\im{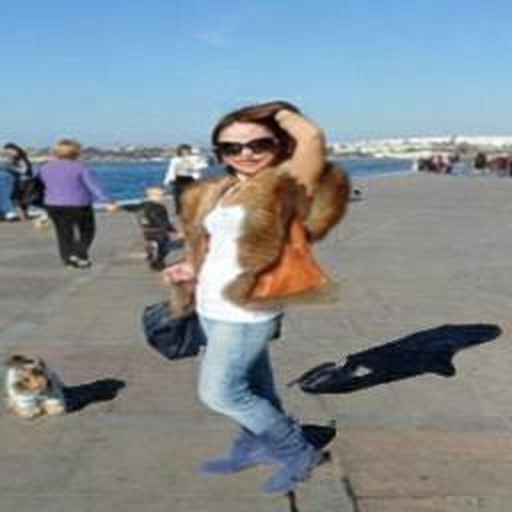} &
\im{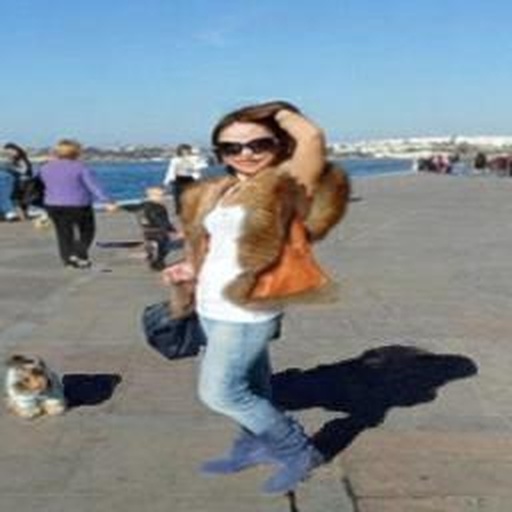} &
\im{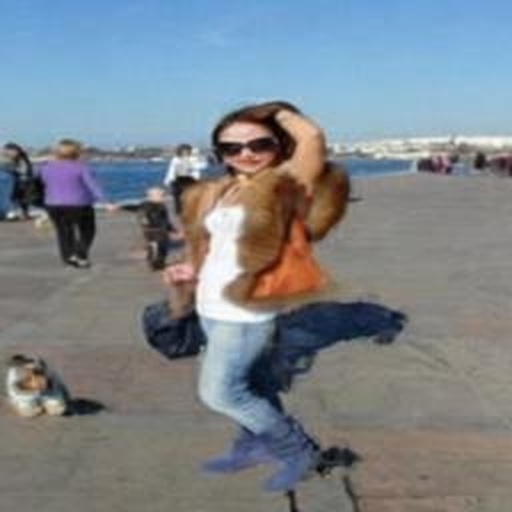} &
\im{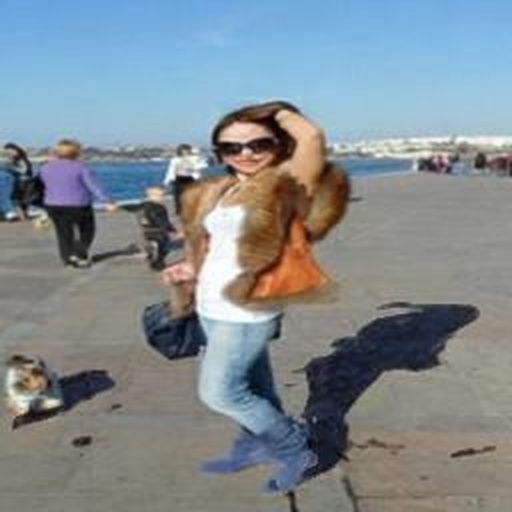} &
\im{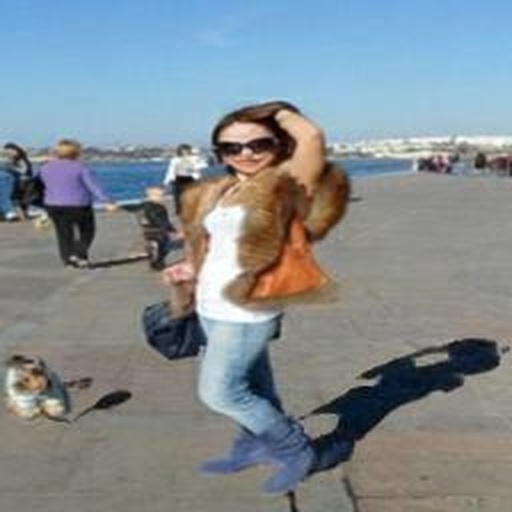} &   
\im{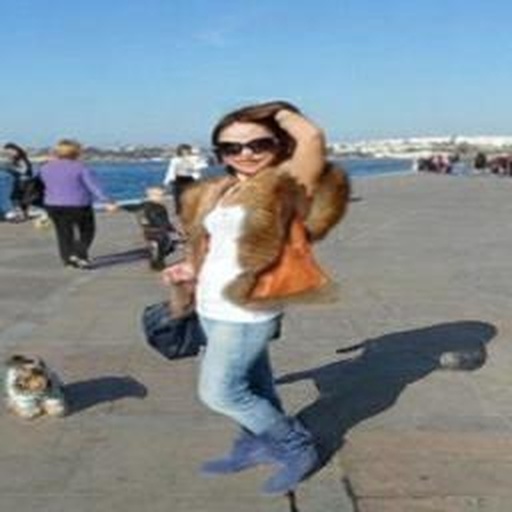} &
\im{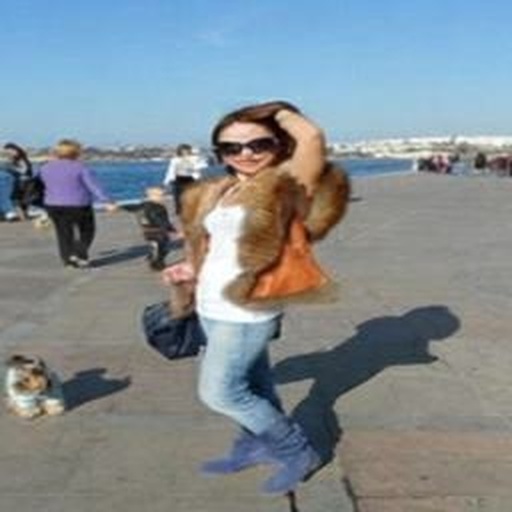} \\[\rgap]

\im{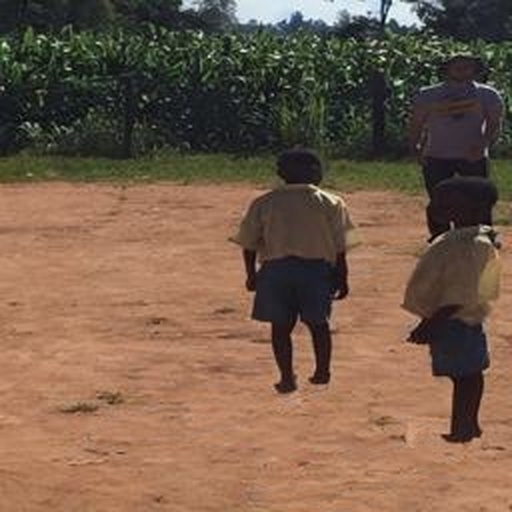} &
\im{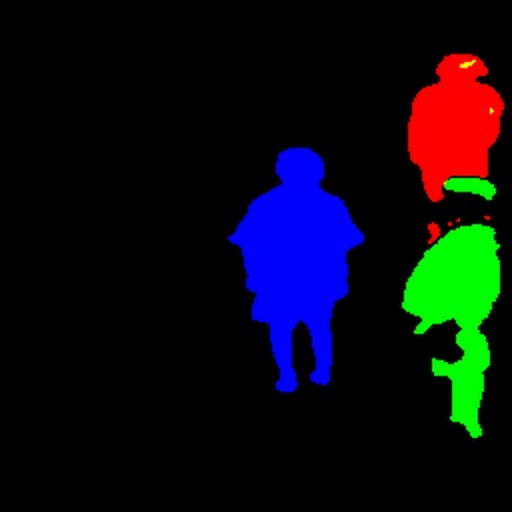} &
\im{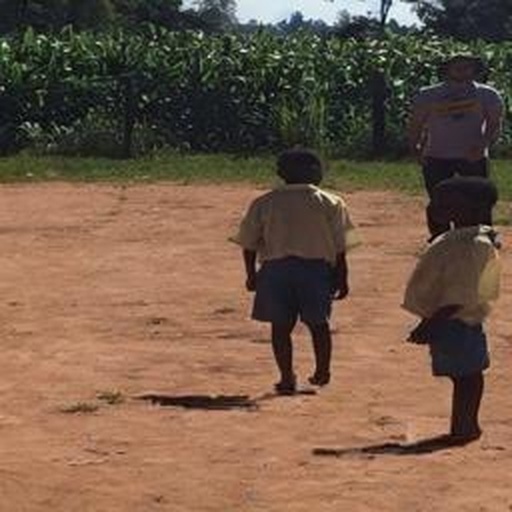} &
\im{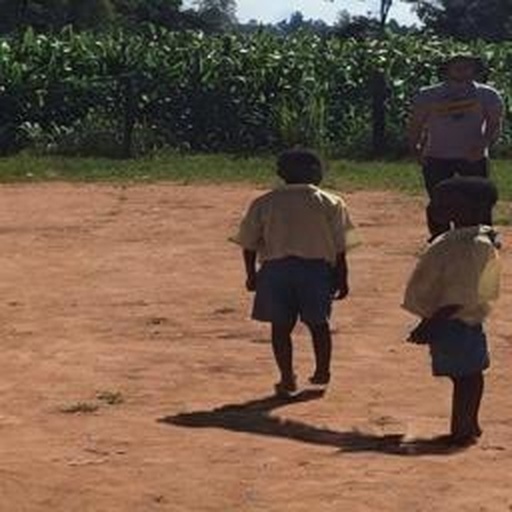} &
\im{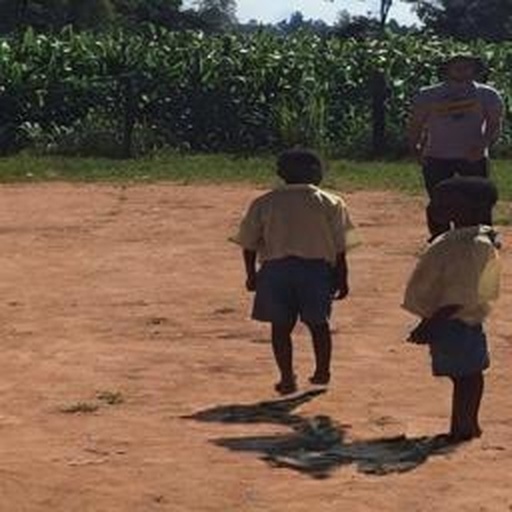} &
\im{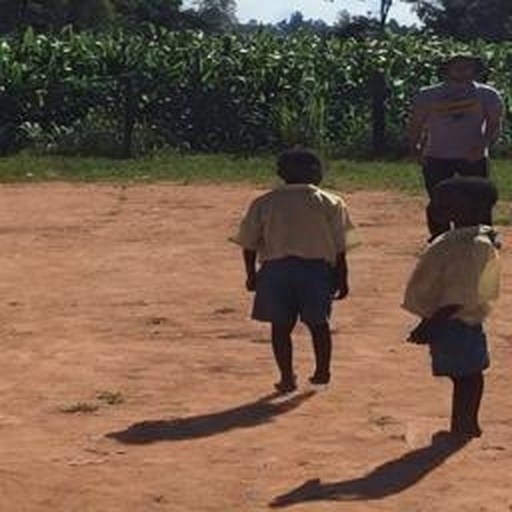} &
\im{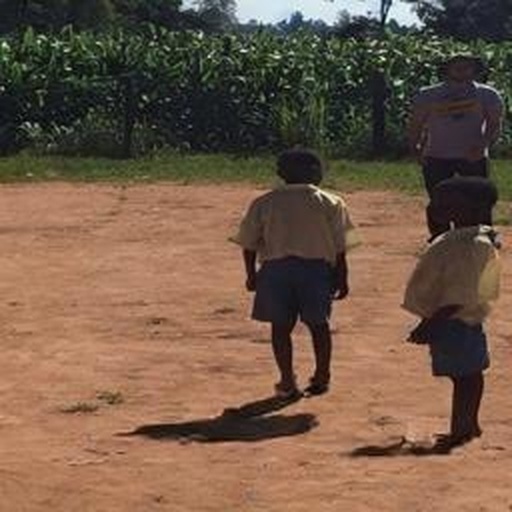} &   
\im{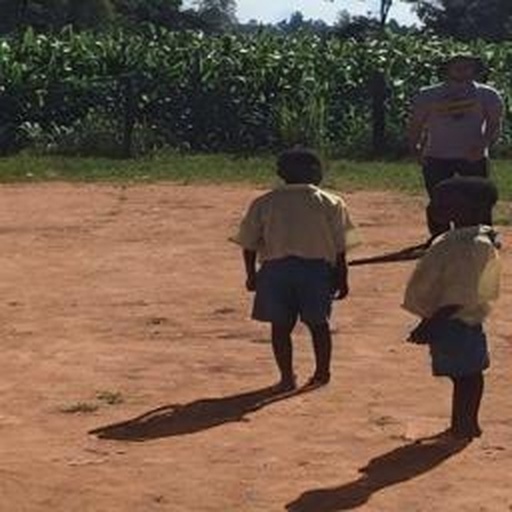} &
\im{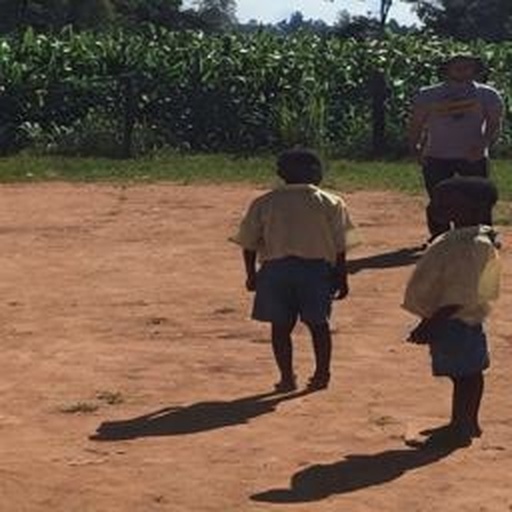} \\[\rgap]

\im{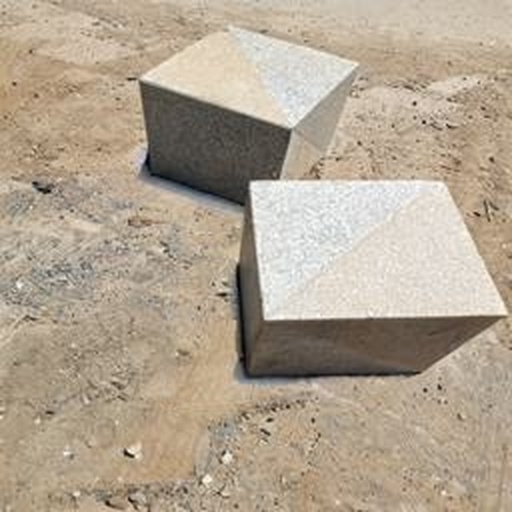} &
\im{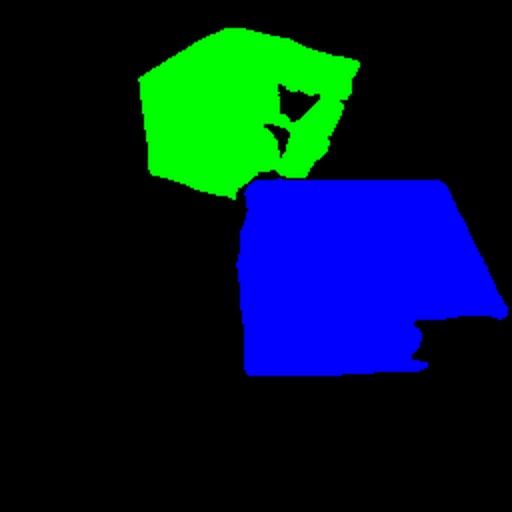} &
\im{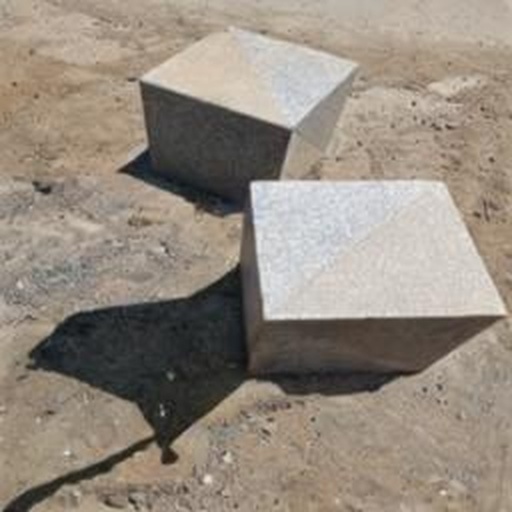} &
\im{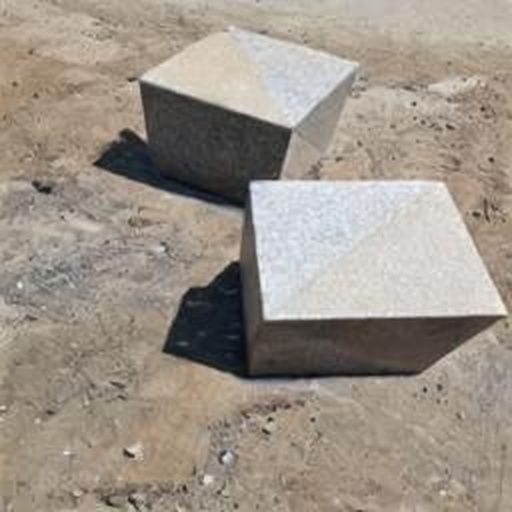} &
\im{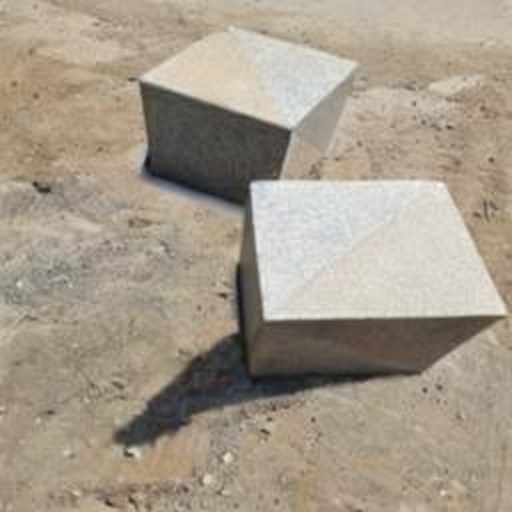} &
\im{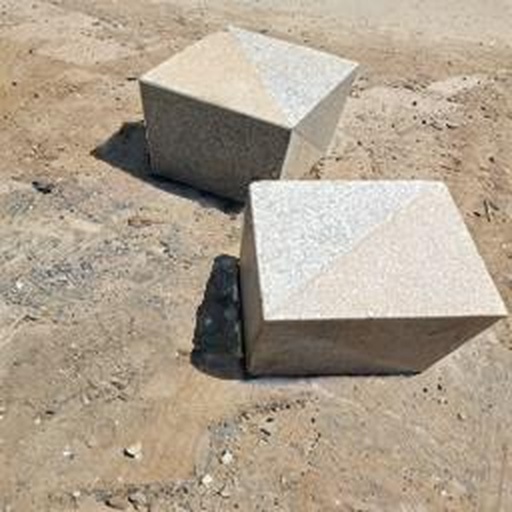} &
\im{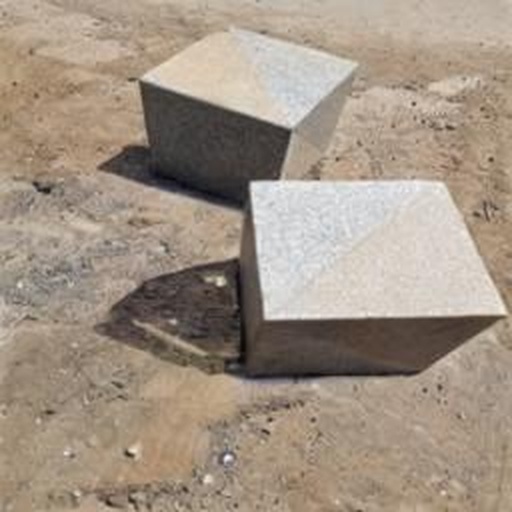} &   
\im{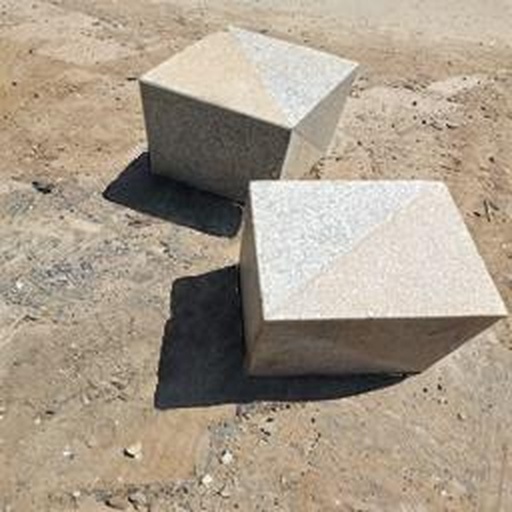} &
\im{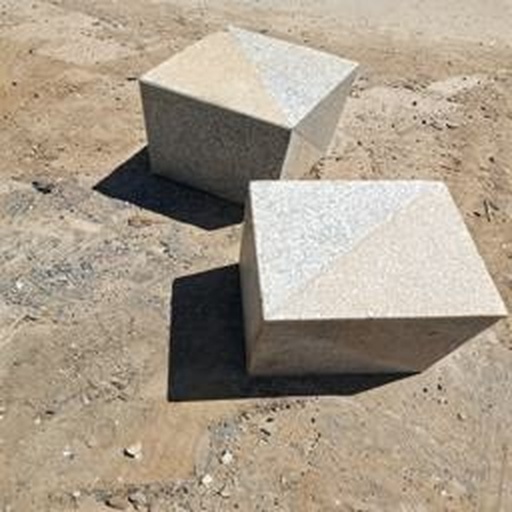} \\[\rgap]

\end{tabular}

\vspace{-1mm}
\caption{Visual comparison with state-of-the-art baseline methods for multiple object shadow generation. Our method (MultiShadow), given the same image inputs plus a compact text prompt of category terms and shadow positional tokens, produces better shadows for all objects (e.g., Row~1: ``a girl casting shadow [sx\_3][sy\_8][sx\_8][sy\_9]; a boy casting shadow [sx\_4][sy\_9][sx\_10][sy\_11]''). Positional tokens are inserted automatically; see Sec.~\ref{sec:text_layout}.}
\label{fig:mutli_comp}
\end{figure*}

\begin{figure*}[t]
\centering

\setlength{\tabcolsep}{0.6pt}        
\renewcommand{\arraystretch}{0}       

\newcommand{\colw}{0.121\textwidth}   

\newcommand{\hdr}[1]{%
  \parbox[c][3.6mm][c]{\colw}{\centering\fontsize{9}{9}\selectfont #1}%
}
\newcommand{\im}[1]{\includegraphics[width=\colw]{#1}}

\newcommand{\hgap}{0.6mm}
\newcommand{\rgap}{0.2mm}
% -----------------------------------------------

\begin{tabular}{@{}cccccccc@{}}
\hdr{Composite} &
\hdr{Mask} &
\hdr{SGRNet} &
\hdr{DAMASNet} &
\hdr{SGDiffusion} &
\hdr{GPSDiffusion} &
\hdr{MetaShadow} &
\hdr{Ours} \\[\hgap]

\im{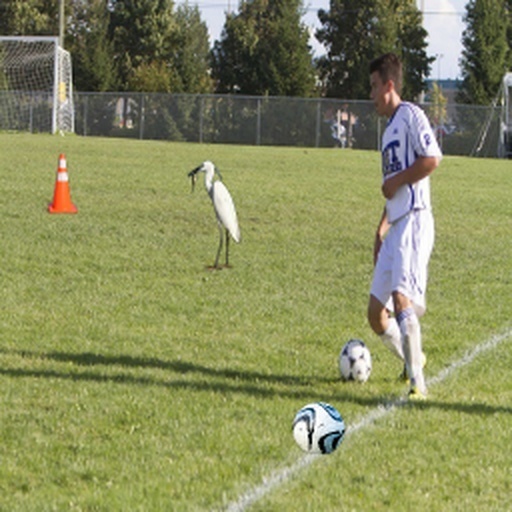} &
\im{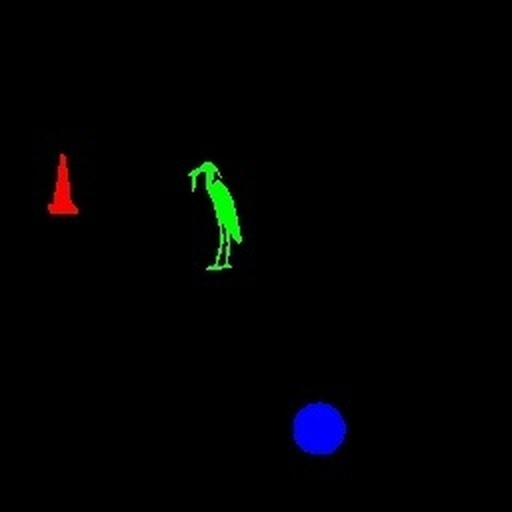} &
\im{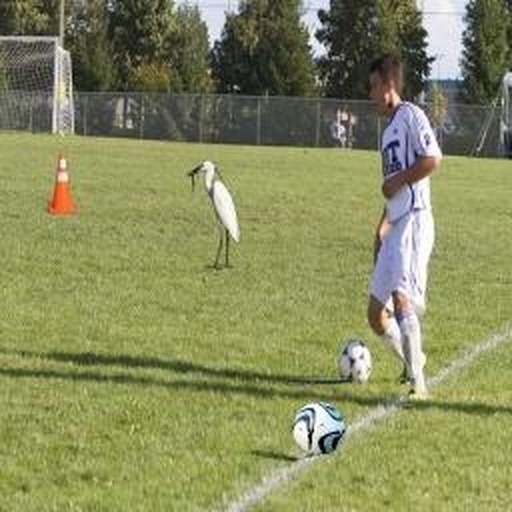} &
\im{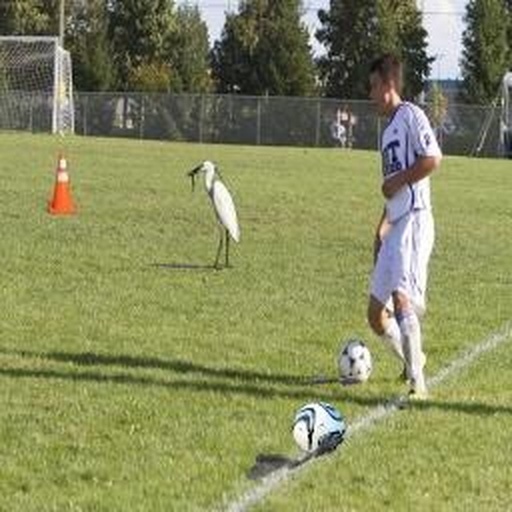} &
\im{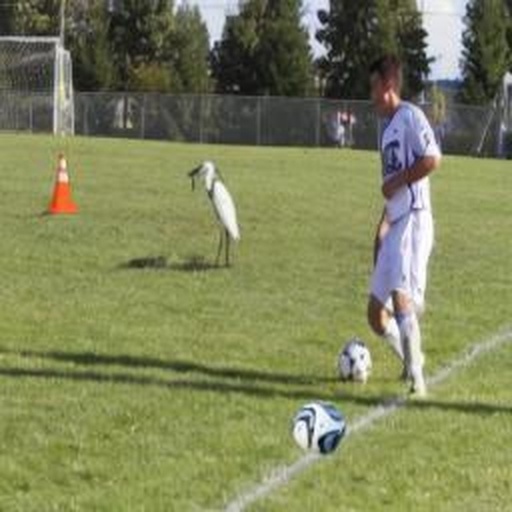} &
\im{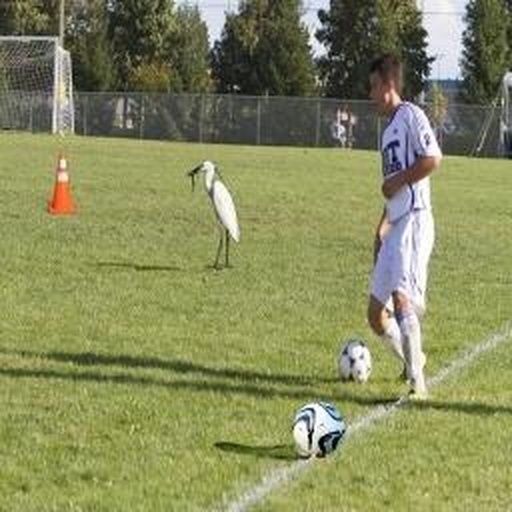} &
\im{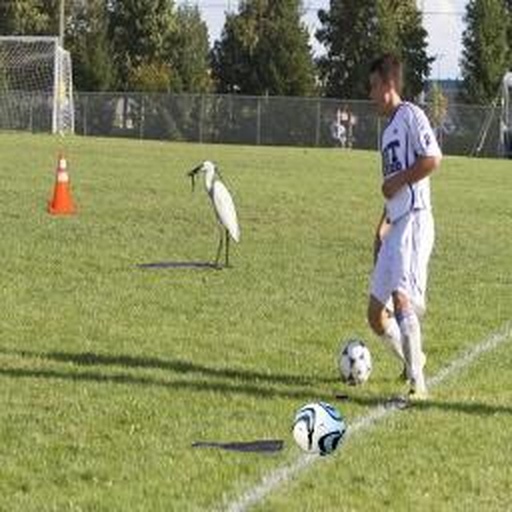} &
\im{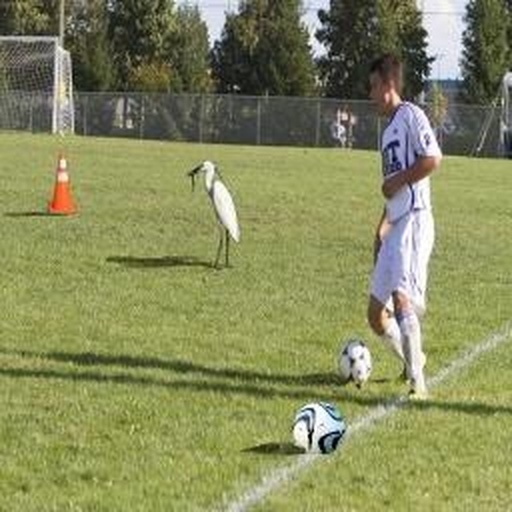} \\[\rgap]

\im{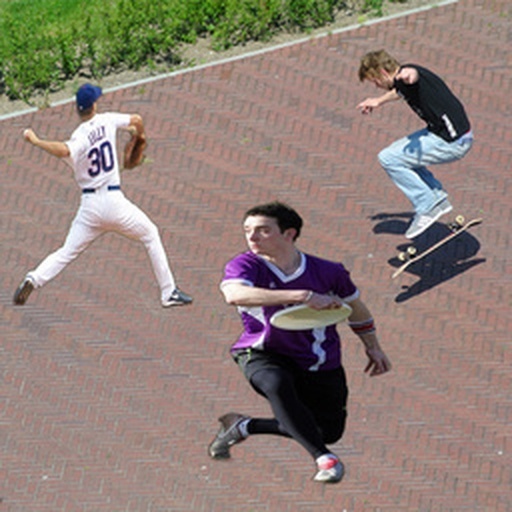} &
\im{Row2RMASK.jpg} &
\im{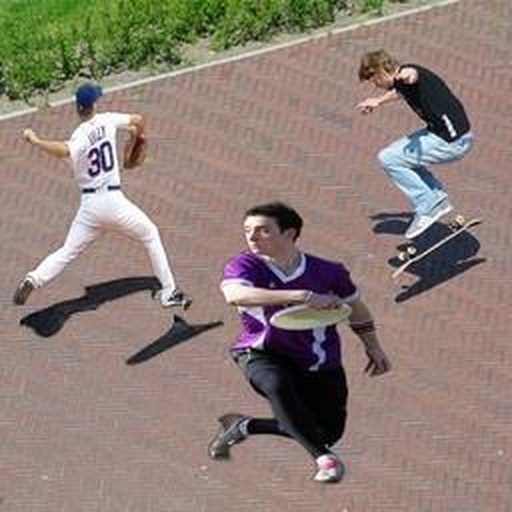} &
\im{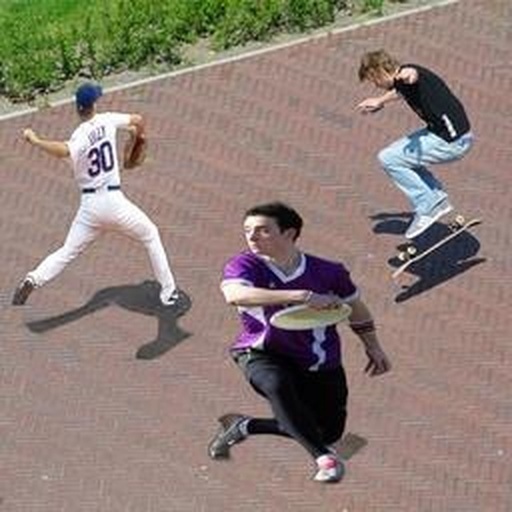} &
\im{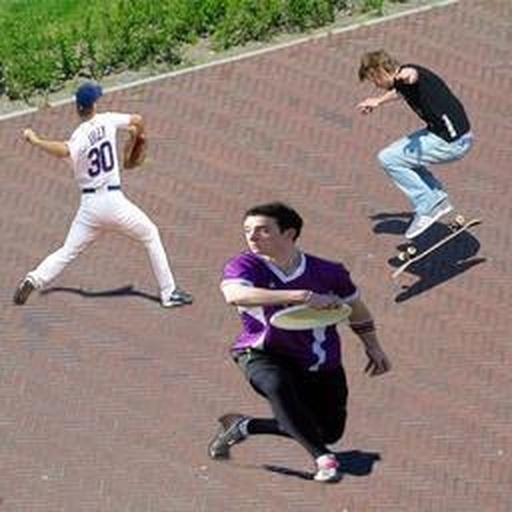} &
\im{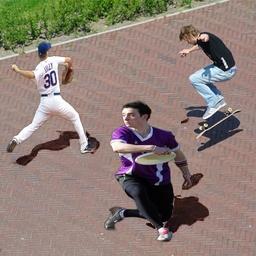} &
\im{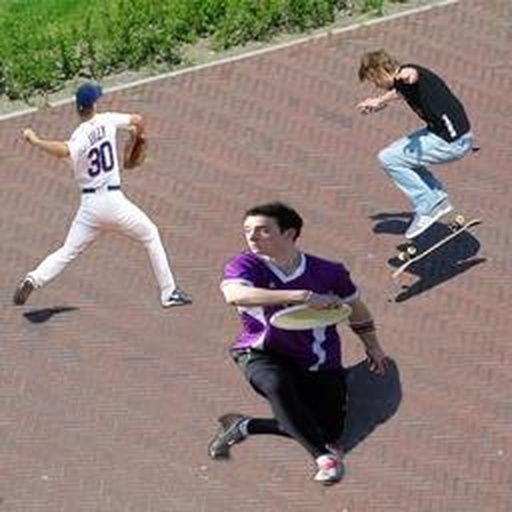} &
\im{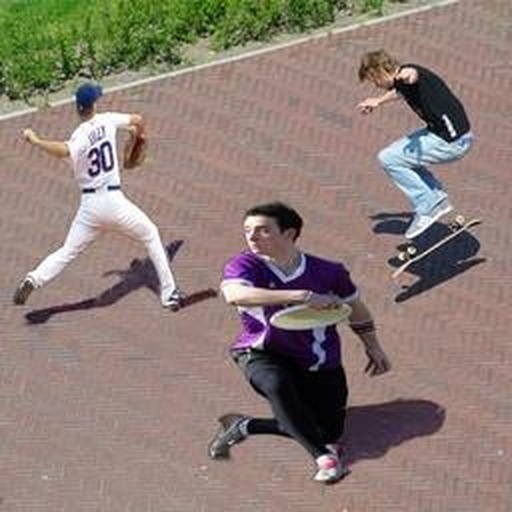} \\[\rgap]

\im{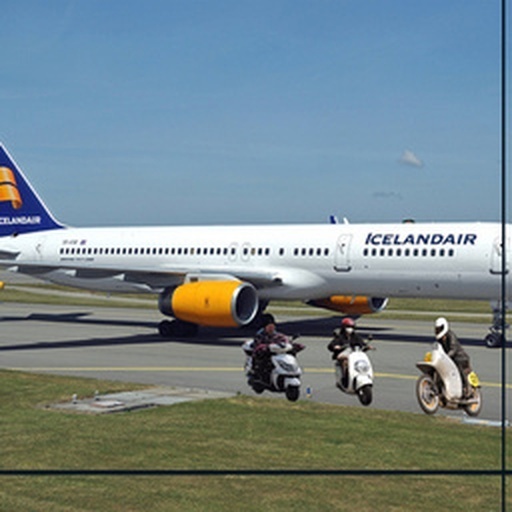} &
\im{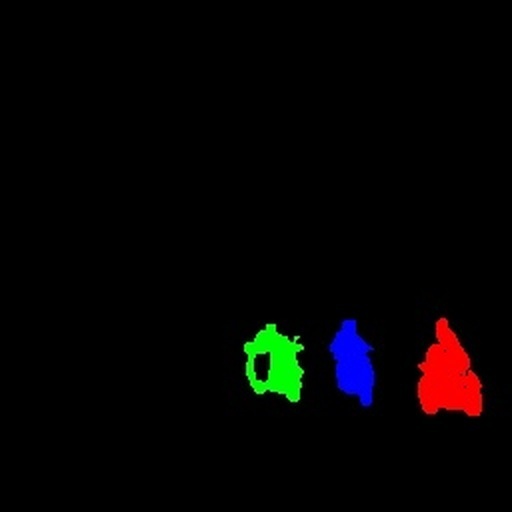} &
\im{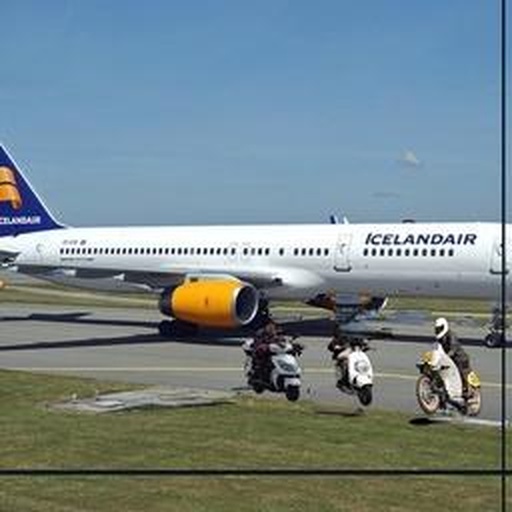} &
\im{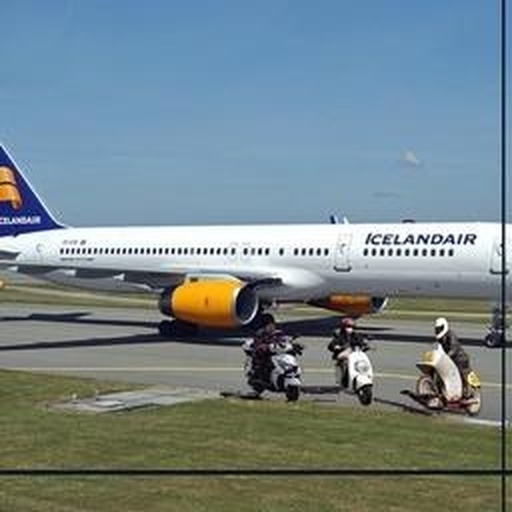} &
\im{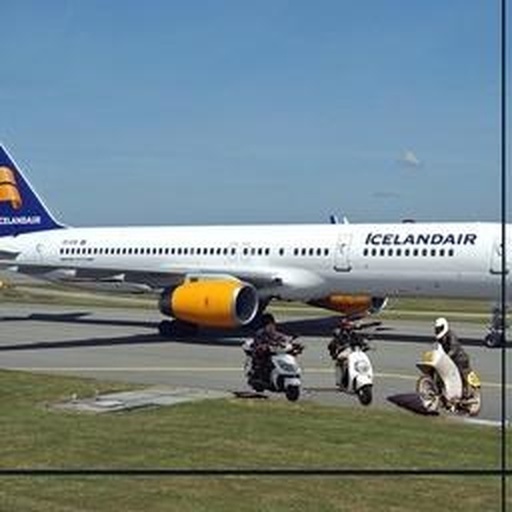} &
\im{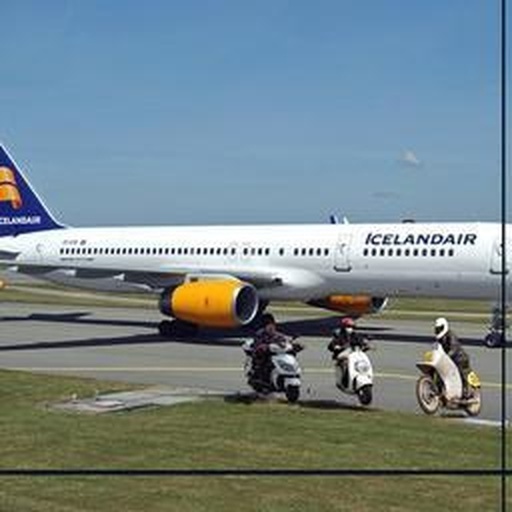} &
\im{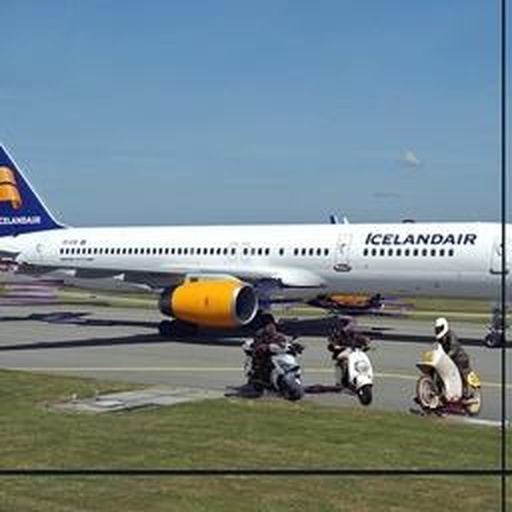} &
\im{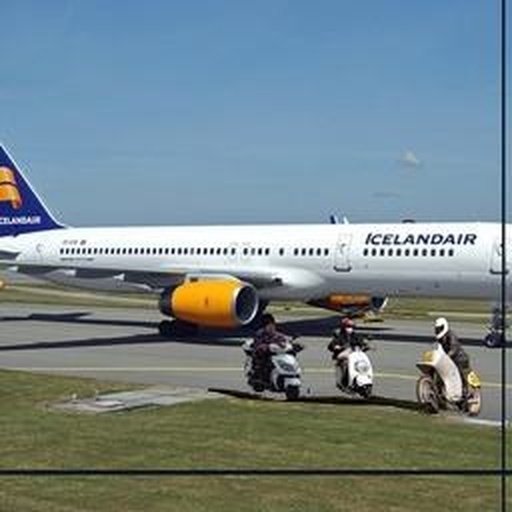} \\[\rgap]

\im{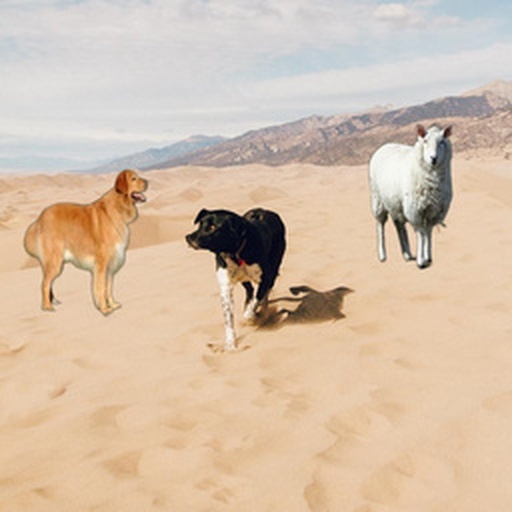} &
\im{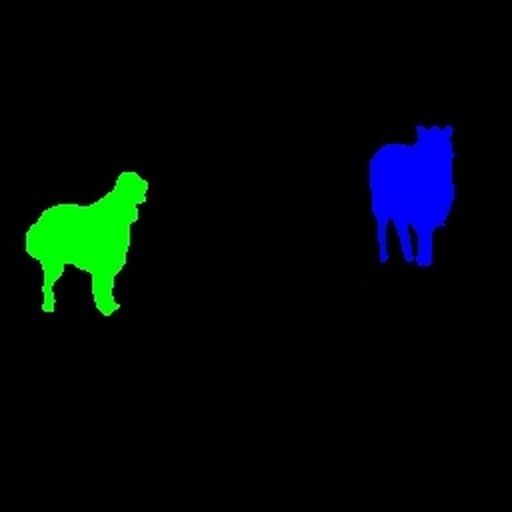} &
\im{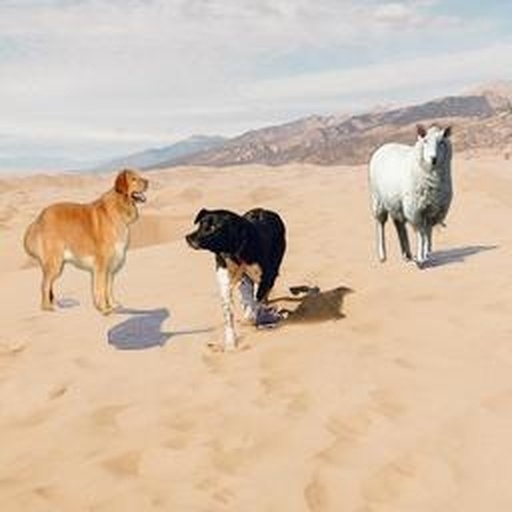} &
\im{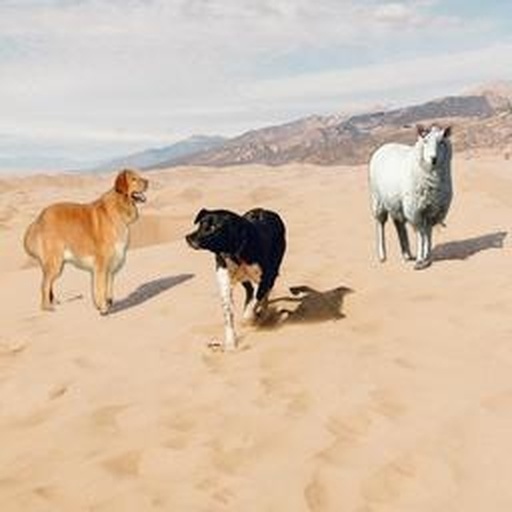} &
\im{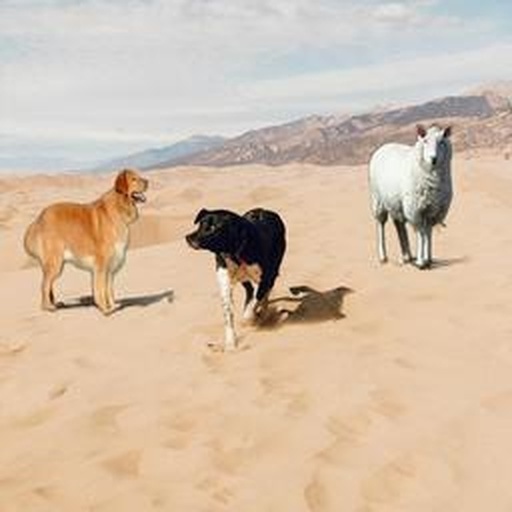} &
\im{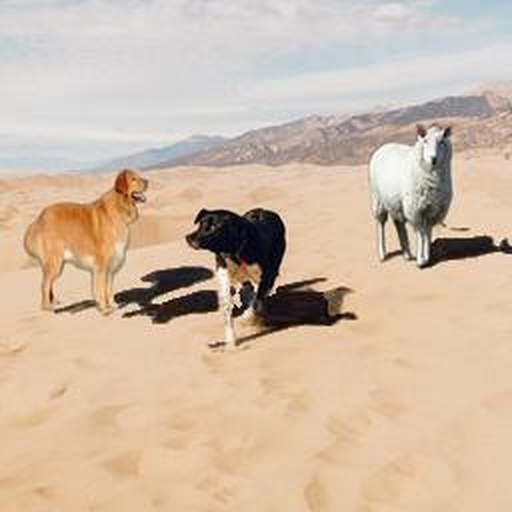} &
\im{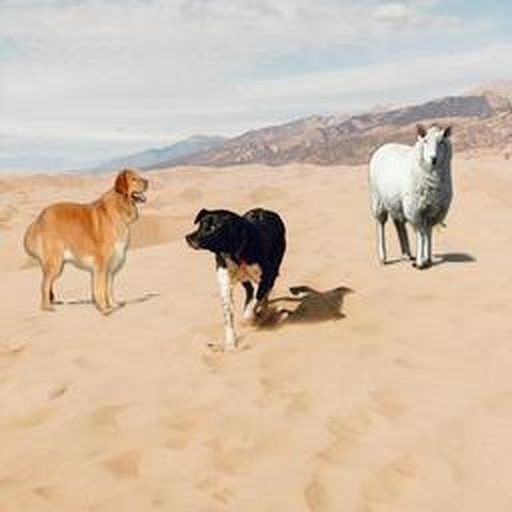} &
\im{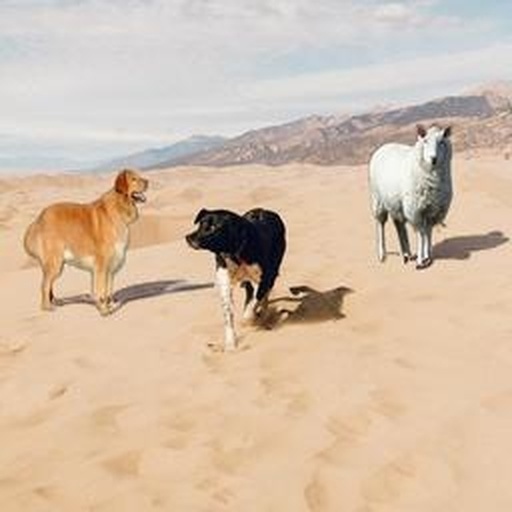} \\[\rgap]

\im{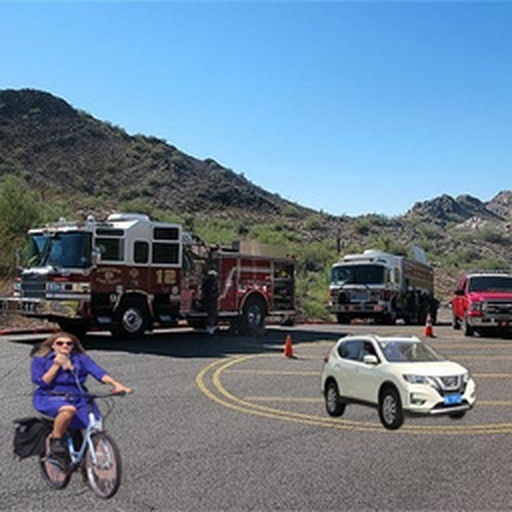} &
\im{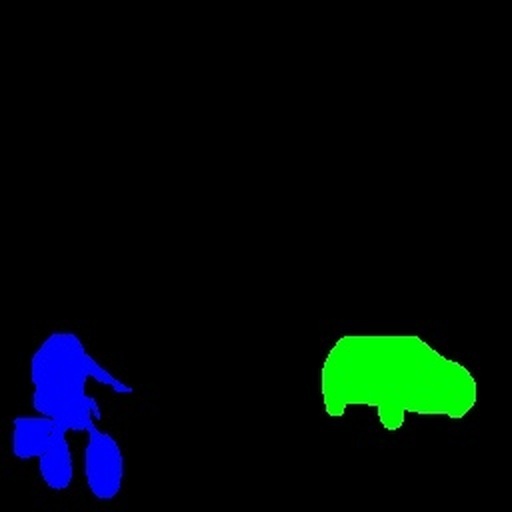} &
\im{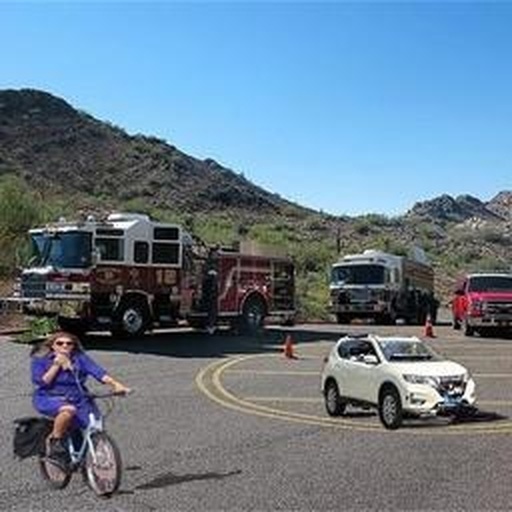} &
\im{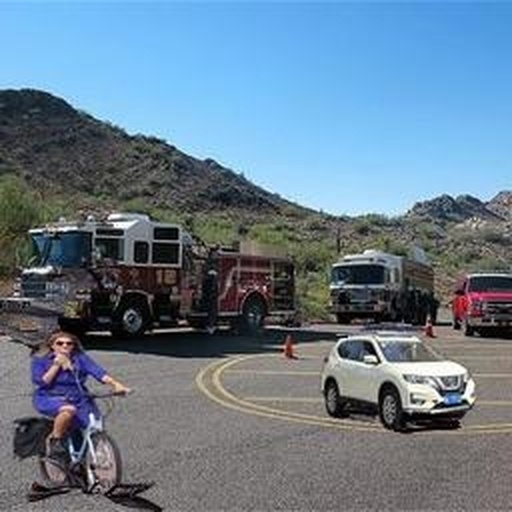} &
\im{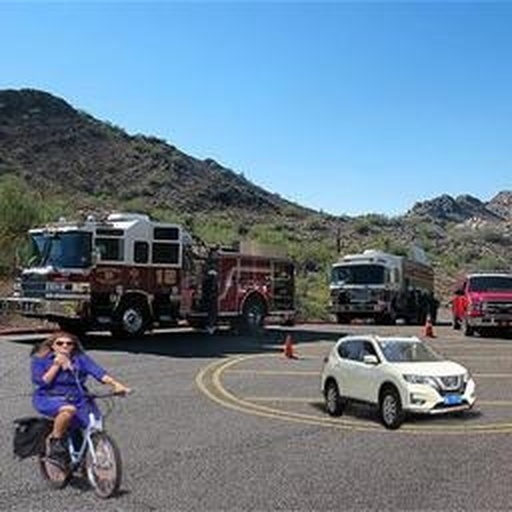} &
\im{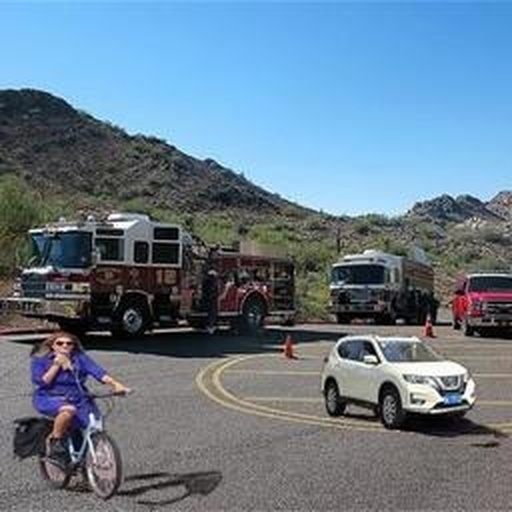} &
\im{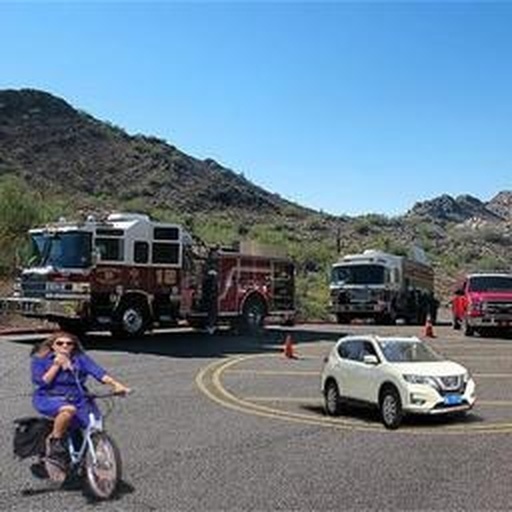} &
\im{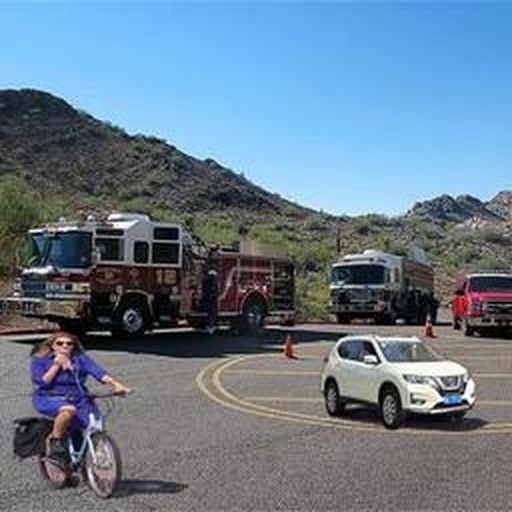} \\[\rgap]

\im{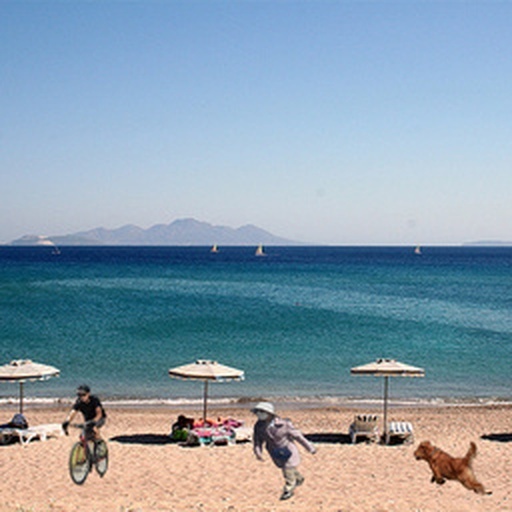} &
\im{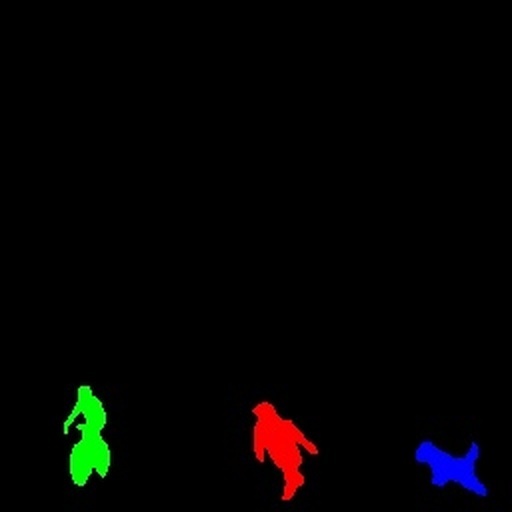} &
\im{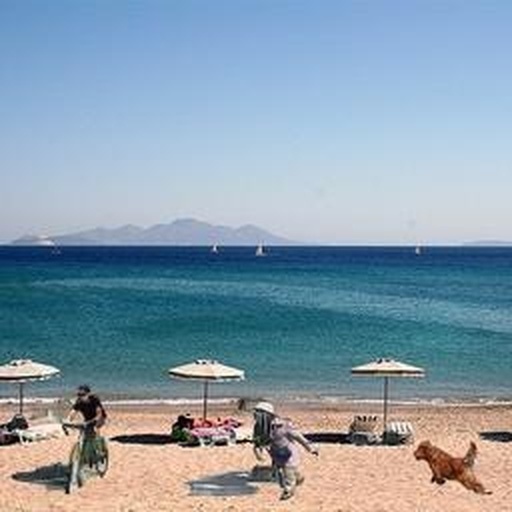} &
\im{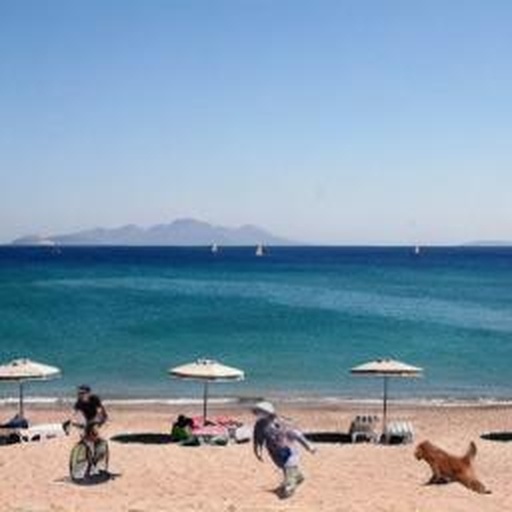} &
\im{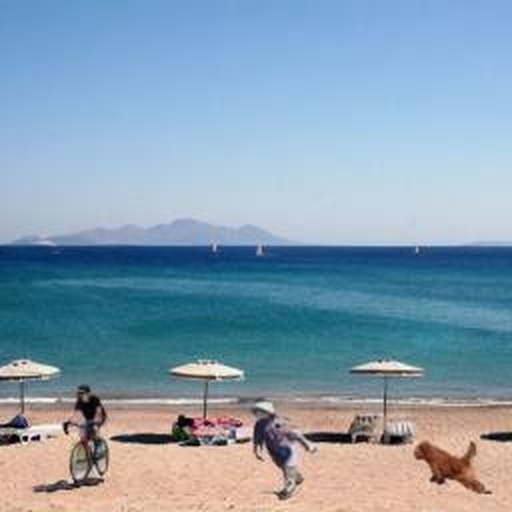} &
\im{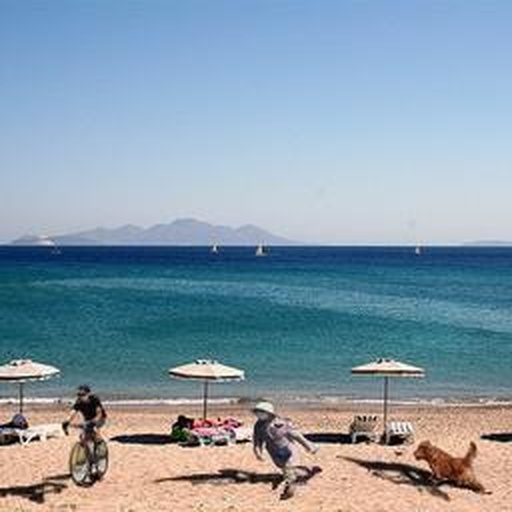} &
\im{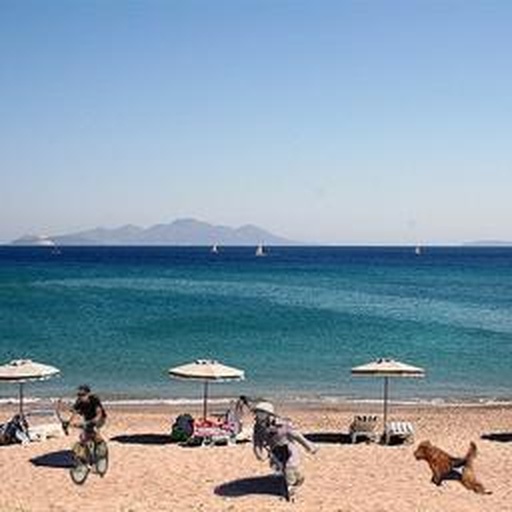} &
\im{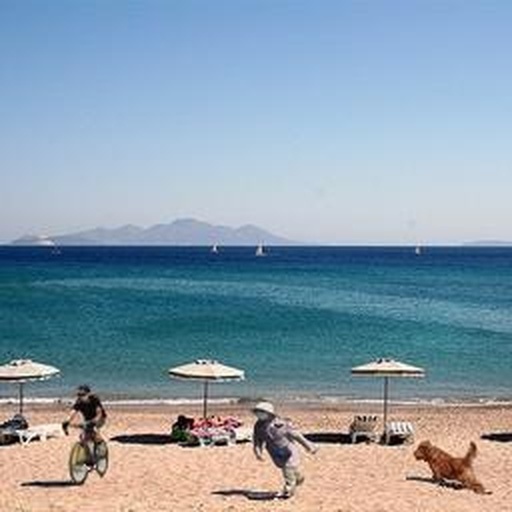} \\[\rgap]

\end{tabular}

\vspace{-1mm}
\caption{Visual comparison with state-of-the-art baseline methods on real composited images. Our method (MultiShadow), given the same image inputs plus a compact text prompt of category terms and shadow positional tokens, produces better shadows for all objects (e.g., Row~1: ``a ball casting shadow [\dots]; a bird casting shadow [\dots], an cone object casting shadow [\dots]''). Positional tokens are inserted automatically; see Sec.~\ref{sec:text_layout}.}
\label{fig:real_shadow}
\end{figure*}

\begin{figure*}[t]
\centering

\setlength{\tabcolsep}{0.5pt}        
\renewcommand{\arraystretch}{0}      

\newcommand{\colw}{0.1185\textwidth}

\newcommand{\hdr}[1]{%
  \parbox[c][3.6mm][c]{\colw}{\centering\fontsize{9}{9}\selectfont #1}%
}
\newcommand{\im}[1]{\includegraphics[width=\colw]{#1}}

\newcommand{\hgap}{0.6mm}  
\newcommand{\rgap}{0.2mm}

\begin{tabular}{@{}cccccccc@{}}
\hdr{Composite} &
\hdr{Objects Mask} &
\hdr{SIC} &
\hdr{GAAM} &
\hdr{GP} &
\hdr{SPT} &
\hdr{AAL} &
\hdr{GT} \\[\hgap]

\im{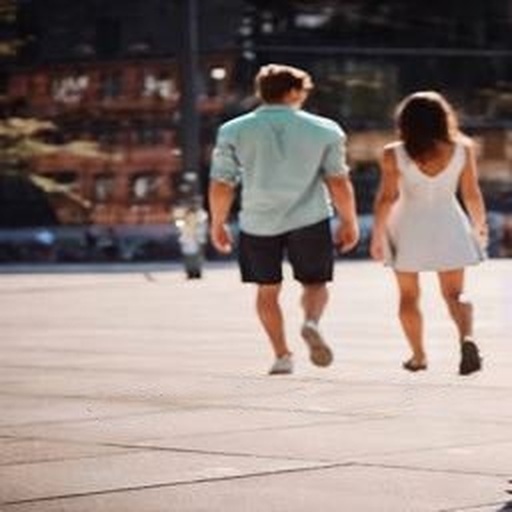} &
\im{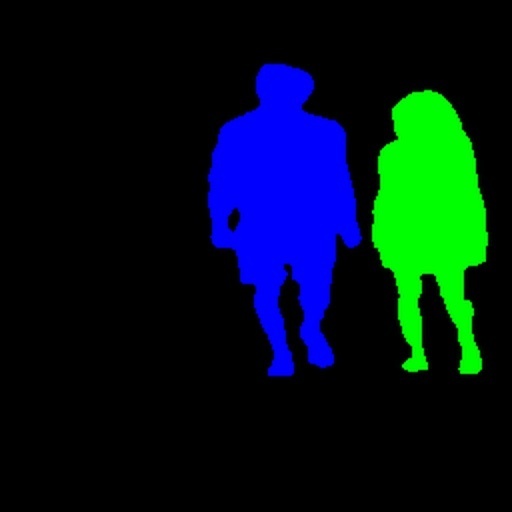} &
\im{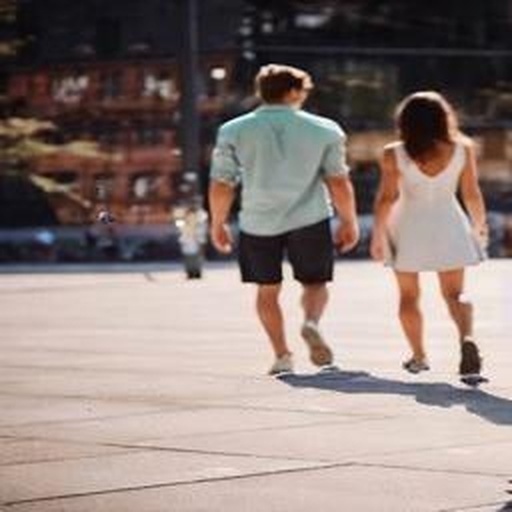} &
\im{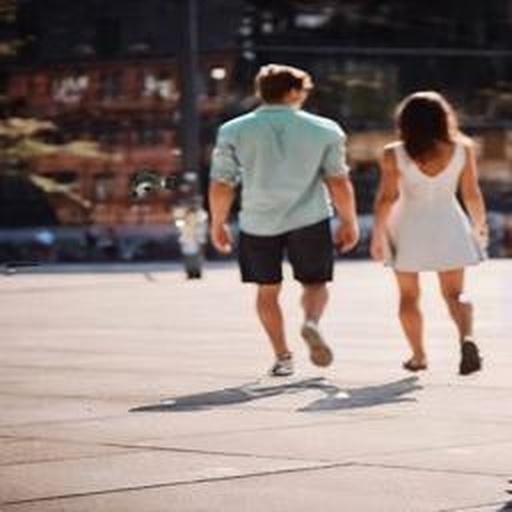} &
\im{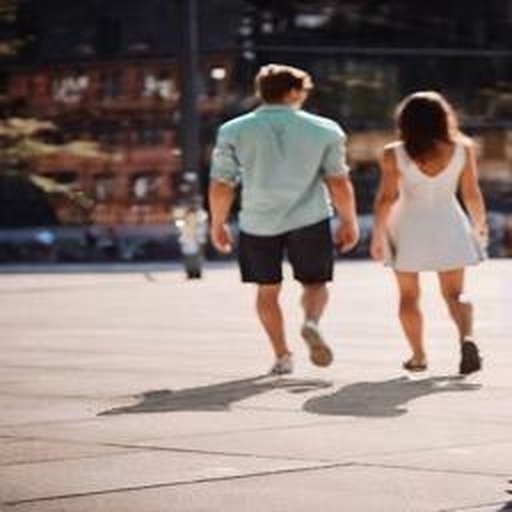} &
\im{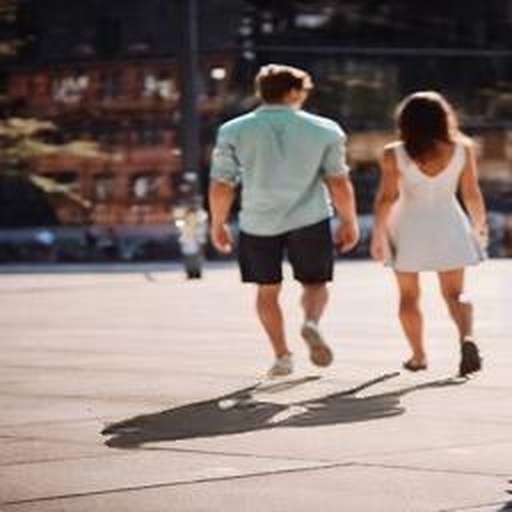} &
\im{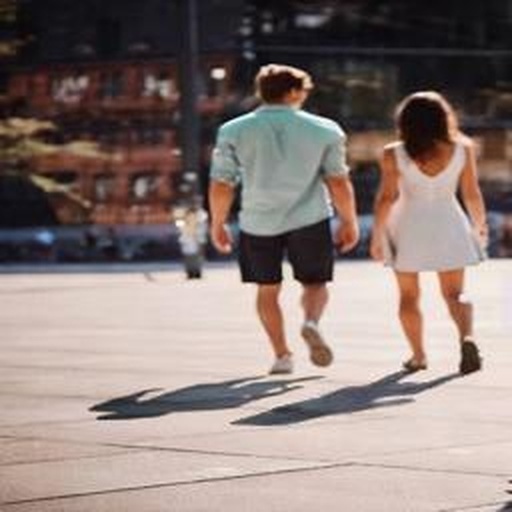} &
\im{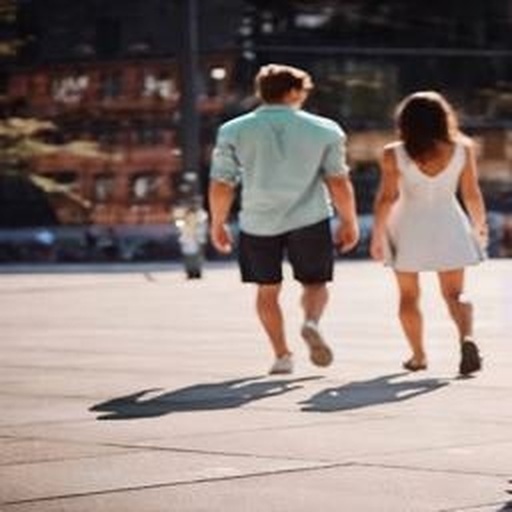} \\[\rgap]

\im{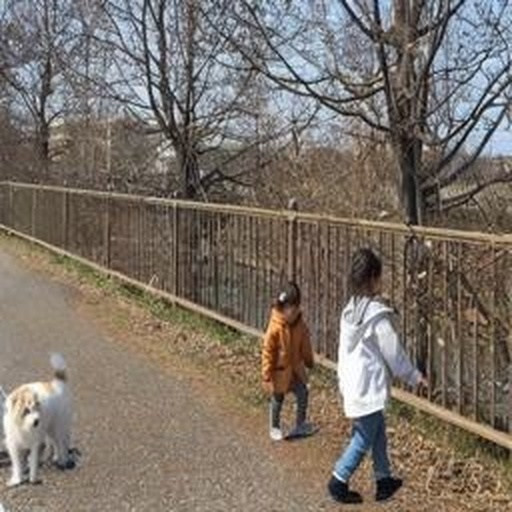} &
\im{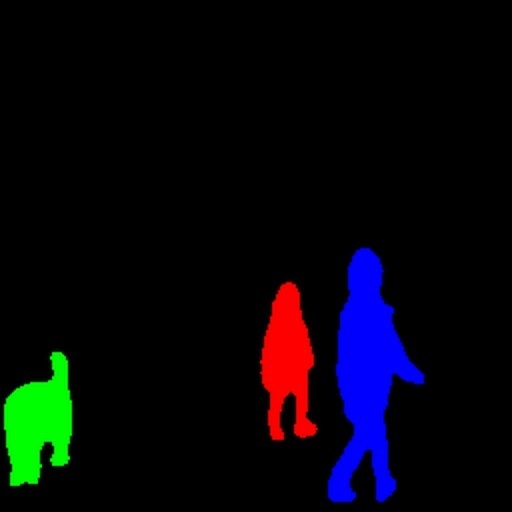} &
\im{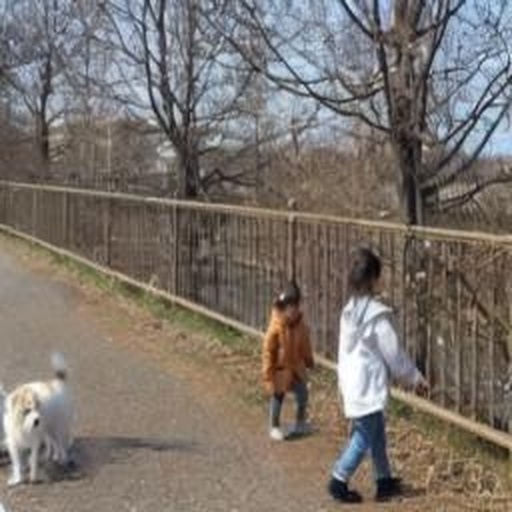} &
\im{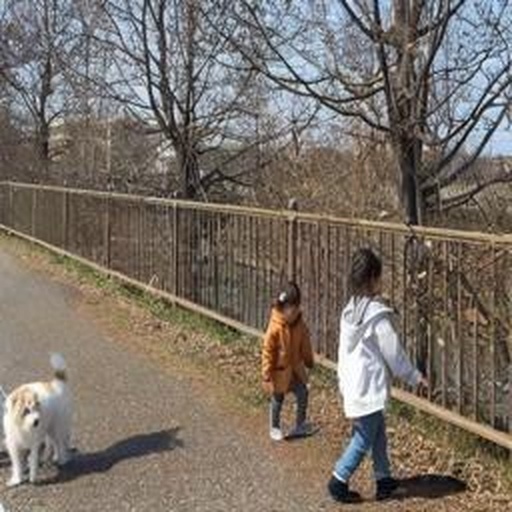} &
\im{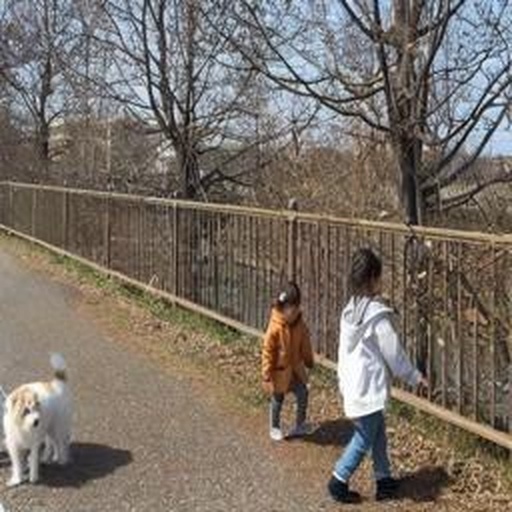} &
\im{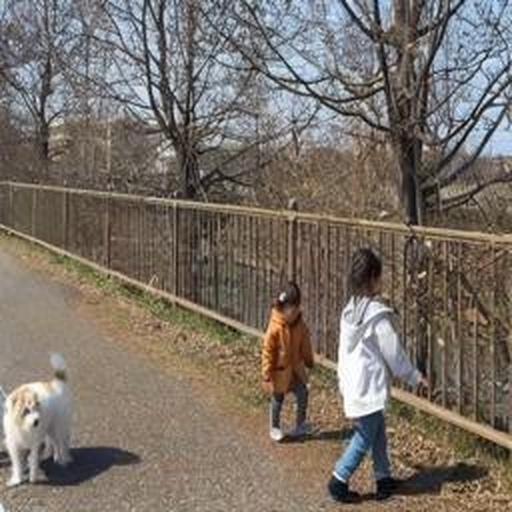} &
\im{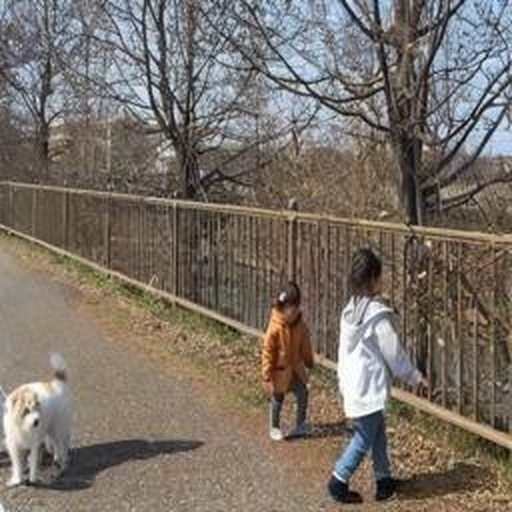} &
\im{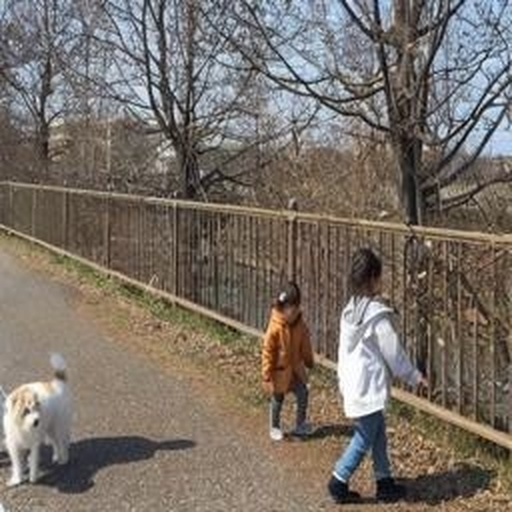} \\[\rgap]

\im{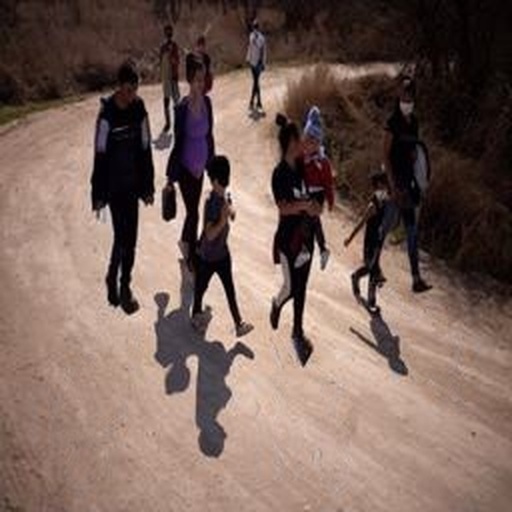} &
\im{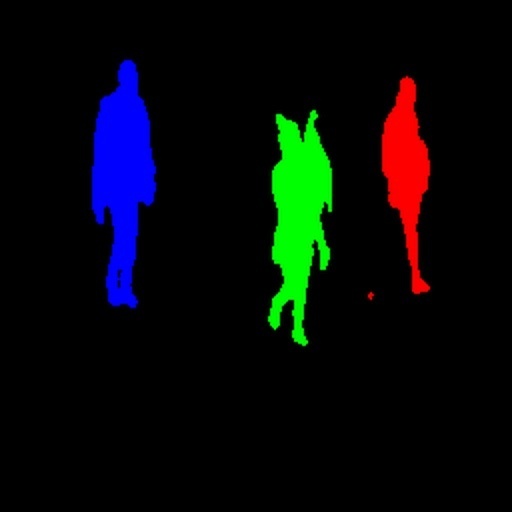} &
\im{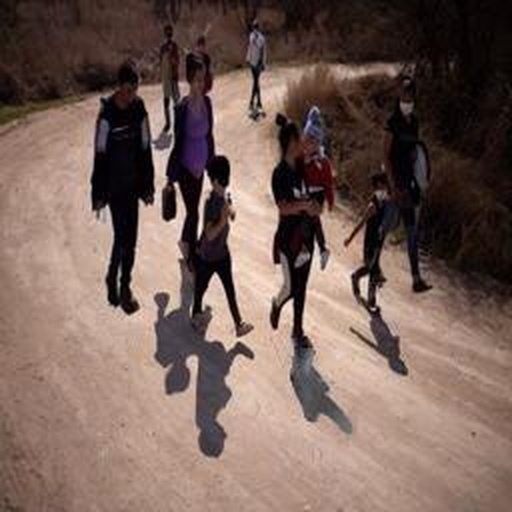} &
\im{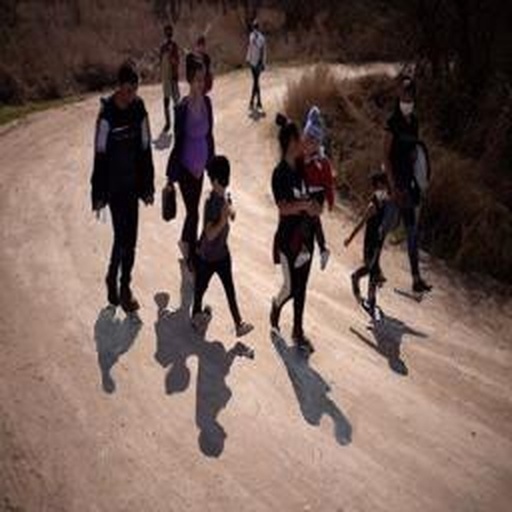} &
\im{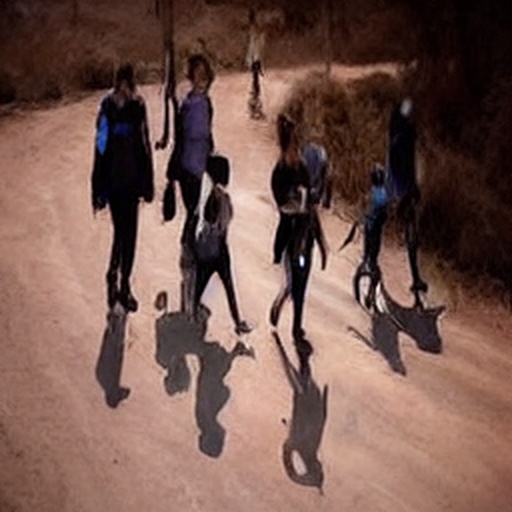} &
\im{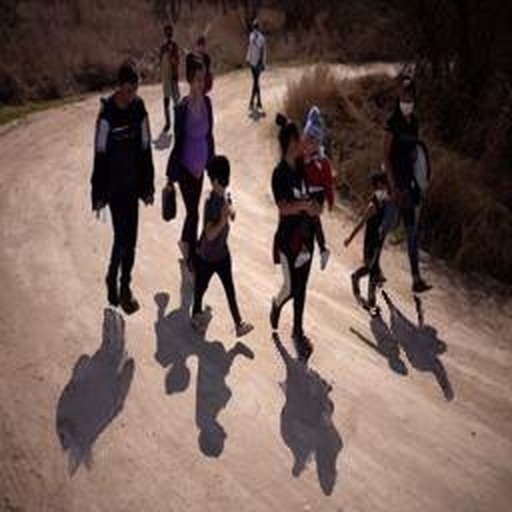} &
\im{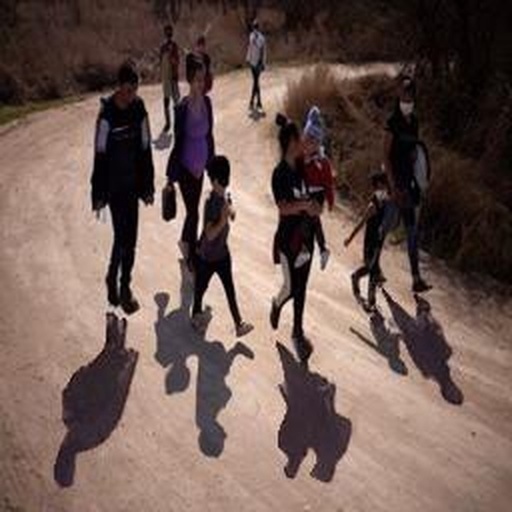} &
\im{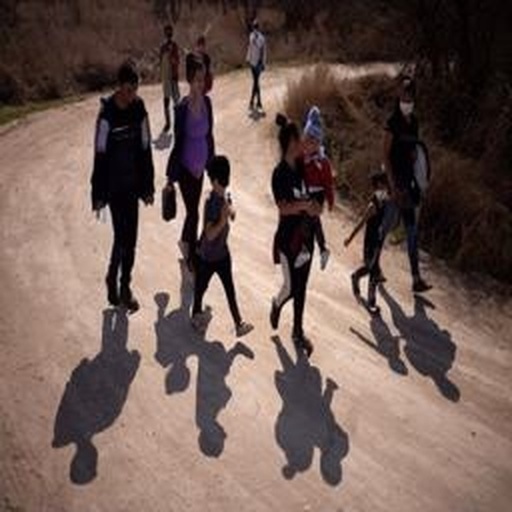} \\[\rgap]

\end{tabular}

\vspace{-1mm}
\caption{Incremental visual ablation from simple image conditioning (SIC) to the full configuration. From left to right: Shadow-free composite, Object Mask, SIC: Simple Image Conditioning, GAAM: Geometry-Aware Affine Modulation, GP: General Prompt, SPT: Shadow Positional Tokens, AAL: Attention Alignment Loss, GT: Ground Truth.}
\label{fig:ab1}
\end{figure*}

\begin{table*}[t]
\caption{Quantitative comparison of single-object shadow generation performance on the original DESOBAv2 \cite{liu2023desobav2} test split.}
\centering
\resizebox{\textwidth}{!}{%
\begin{tabular}{lcccccc|cccccc}
\hline
\multirow{2}{*}{Method} & \multicolumn{6}{c|}{BOS Test Images} & \multicolumn{6}{c}{BOS-free Test Images} \\
\cmidrule(lr){2-7} \cmidrule(lr){8-13}
 & GR $\downarrow$ & LR $\downarrow$ & GS $\uparrow$ & LS $\uparrow$ & GB $\downarrow$ & LB $\downarrow$ & GR $\downarrow$ & LR $\downarrow$ & GS $\uparrow$ & LS $\uparrow$ & GB $\downarrow$ & LB $\downarrow$ \\
\hline
SGRNet~\cite{hong2022shadow} & 7.184 & 68.255 & 0.964 & 0.206 & 0.301 & 0.596 & 15.596 & 60.350 & 0.909 & 0.100 & 0.271 & 0.534 \\
DMASNet~\cite{tao2024shadow} & 8.256 & 59.380 & 0.961 & 0.228 & 0.276 & 0.547 & 18.725 & 86.694 & 0.913 & 0.055 & 0.297 & 0.574 \\
SGDiffusion~\cite{liu2024shadow} & 6.098 & 53.611 & 0.971 & 0.370 & 0.245 & 0.487 & 15.110 & 55.874 & 0.913 & 0.117 & 0.233 & 0.452 \\
GPSDiffusion~\cite{zhao2025shadow} & 5.896 & 46.713 & 0.966 & 0.374 & 0.213 & 0.423 & 13.809 & 55.616 & 0.917 & 0.166 & 0.197 & 0.384 \\
MetaShadow~\cite{wang2025metashadow} & 5.961 & 47.328 & 0.968 & 0.373 & 0.237 & 0.458 & 14.698 & 55.781 & 0.916 & 0.152 & 0.209 & 0.426 \\
\hline
\textbf{MultiShadow} (Ours) & \textbf{5.723} & \textbf{44.892} & \textbf{0.974} & \textbf{0.398} & \textbf{0.198} & \textbf{0.395} & \textbf{13.456} & \textbf{52.134} & \textbf{0.925} & \textbf{0.185} & \textbf{0.182} & \textbf{0.361} \\
\hline
\end{tabular}%
}
\label{tab:singlecomparison}
\vspace{0.1in}
\end{table*}

\renewcommand{\arraystretch}{1.1}
\begin{table*}[t]
\caption{Performance evaluation on the extended DESOBAv2 \cite{liu2023desobav2} dataset for multi-object shadow generation.}
\label{tab:multi_object_comparison}
\centering
\resizebox{\textwidth}{!}{%
\begin{tabular}{lcccccc|cccccc}
\hline
\multirow{2}{*}{Method} & \multicolumn{6}{c|}{BOS Test Images (Multi-Object)} & \multicolumn{6}{c}{BOS-free Test Images (Multi-Object)} \\
\cmidrule(lr){2-7} \cmidrule(lr){8-13}
 & GR $\downarrow$ & LR $\downarrow$ & GS $\uparrow$ & LS $\uparrow$ & GB $\downarrow$ & LB $\downarrow$ & GR $\downarrow$ & LR $\downarrow$ & GS $\uparrow$ & LS $\uparrow$ & GB $\downarrow$ & LB $\downarrow$ \\
\hline
SGRNet~\cite{hong2022shadow} & 9.842 & 85.327 & 0.938 & 0.152 & 0.412 & 0.723 & 18.936 & 78.452 & 0.872 & 0.068 & 0.385 & 0.698 \\
DMASNet~\cite{tao2024shadow} & 10.125 & 72.893 & 0.935 & 0.178 & 0.367 & 0.654 & 21.583 & 94.217 & 0.885 & 0.042 & 0.394 & 0.712 \\
SGDiffusion~\cite{liu2024shadow} & 8.327 & 67.425 & 0.952 & 0.285 & 0.324 & 0.592 & 17.892 & 68.327 & 0.894 & 0.089 & 0.315 & 0.567 \\
GPSDiffusion~\cite{zhao2025shadow} & 7.894 & 58.642 & 0.948 & 0.301 & 0.294 & 0.528 & 16.427 & 67.893 & 0.901 & 0.124 & 0.278 & 0.489 \\
MetaShadow~\cite{wang2025metashadow} & 8.293 & 60.519 & 0.951 & 0.293 & 0.304 & 0.573 & 17.253 & 68.032 & 0.904 & 0.117 & 0.293 & 0.526 \\
\hline
\textbf{MultiShadow} (Ours) & \textbf{6.892} & \textbf{51.327} & \textbf{0.961} & \textbf{0.342} & \textbf{0.235} & \textbf{0.428} & \textbf{15.127} & \textbf{61.458} & \textbf{0.912} & \textbf{0.152} & \textbf{0.243} & \textbf{0.412} \\
\hline
\end{tabular}%
}
\vspace{0.1in}
\end{table*}

\begin{table}[t]
\centering
\caption{Object count $K$ vs.\ local BER (LB$\downarrow$; lower is better).}
\label{tab:obj_count_bins}
\small
\setlength{\tabcolsep}{6pt}
\renewcommand{\arraystretch}{1.15}

\begin{tabular}{lcccc}
\toprule
\textbf{Method} & \textbf{$K{=}1$} & \textbf{$K{=}2$} & \textbf{$K{=}3$--$4$} & \textbf{$K\ge 5$} \\
\midrule
SGRNet~\cite{hong2022shadow}        & 0.746 & 0.762 & 0.852 & 0.930 \\
DMASNet~\cite{tao2024shadow}       & 0.673 & 0.691 & 0.720 & 0.886 \\
SGDiffusion~\cite{liu2024shadow}   & 0.645 & 0.674 & 0.785 & 0.842 \\
GPSDiffusion~\cite{zhao2025shadow}  & 0.514 & 0.573 & 0.617 & 0.662 \\
MetaShadow~\cite{wang2025metashadow}  & 0.539 & 0.582 & 0.630 & 0.687 \\
\textbf{MultiShadow} (Ours) & \textbf{0.395} & \textbf{0.403} & \textbf{0.412} & \textbf{0.426} \\
\bottomrule
\end{tabular}
\end{table}

\subsection{Comparison with Baselines}
As a first step, we evaluate our method in the single-object shadow generation setting, comparing against leading baselines SGRNet \cite{hong2022shadow}, DMASNet \cite{tao2024shadow}, SGDiffusion \cite{liu2024shadow}, GPSDiffusion \cite{zhao2025shadow}, and MetaShadow \cite{wang2025metashadow}. As summarized in Table \ref{tab:singlecomparison}, our method demonstrates consistent superiority across all evaluation metrics. The improvements are especially strong on local metrics, which confirms that our method generates shadows with better geometry and cleaner attachment to object boundaries. Qualitative results, shown in Fig.~\ref{fig:single1}, further support these findings. Compared to prior methods, our results exhibit fewer artifacts, better alignment with the expected shadow region, and more coherent intensities.

For multi-object evaluation, we adapt each baseline to its architectural constraints to ensure a fair comparison. DMASNet~\cite{tao2024shadow}, designed for single-object inference, is applied sequentially, generating shadows for each object one at a time. In contrast, SGRNet~\cite{hong2022shadow}, SGDiffusion~\cite{liu2024shadow}, GPSDiffusion~\cite{zhao2025shadow}, and MetaShadow~\cite{wang2025metashadow} support one-pass multi-object synthesis by accepting multi-object mask as input, and we evaluate them under this setting. This evaluation reveals distinct failure modes across method categories,  as illustrated in Fig.~\ref{fig:mutli_comp}. Non-diffusion methods SGRNet~\cite{hong2022shadow} and DMASNet~\cite{tao2024shadow} tend to produce weak shadows with inaccurate shape and poor object attachment. Diffusion-based baselines SGDiffusion~\cite{liu2024shadow}, GPSDiffusion~\cite{zhao2025shadow}, and MetaShadow~\cite{wang2025metashadow} frequently exhibit multi-object failure modes, including missing and bleeding shadows, inconsistent geometry across instances, and non-uniform intensities. In contrast, our approach generates shadows for all objects within a shared latent context and introduces per-object shadow grounding tokens derived from predicted shadow layouts. An attention-alignment objective further encourages token-specific cross-attention to focus on the corresponding shadow regions, improving object–shadow correspondence and mitigating shadow bleeding in multi-object scenes. Table~\ref{tab:multi_object_comparison} summarizes the quantitative results, where our method achieves the best performance across all evaluation metrics. We further analyze scalability by splitting the multi-object test set according to object count ($K{=}1$, $K{=}2$, $K{=}3\text{--}4$, and $K{\ge}5$). The local BER results in Table~\ref{tab:obj_count_bins} show that our method degrades more gracefully and remains robust as the number of inserted objects increases.

\subsubsection{Results on Real Composited Scenes}
To further assess generalization, we evaluate on real composite images spanning diverse scenes and lighting conditions. As illustrated in Fig. \ref{fig:real_shadow}, our method successfully generates shadows for all composited objects that are both realistic and physically coherent. Unlike baseline methods, our model remains stable under challenging cases, thin structures, partial occlusions, and scale changes, without bleeding, double shadows, or halo artifacts, indicating strong robustness beyond the training distribution. 

Since ground-truth shadows are unavailable for real composites, we conduct a user study with over 100 participants. Using pairwise comparisons against baselines, we compute Bradley–Terry (BT) \cite{bradley1952rank} preference scores to evaluate shadow quality. The results shown in the Table \ref{tab:user_study_combined} confirm that our method is consistently preferred over existing approaches, demonstrating its effectiveness in real-world scenarios.

\begin{table}[t]
\centering
\caption{User study results on real composite images for both single-object and multi-object shadow generation.}
\label{tab:user_study_combined}
\begin{tabular}{l c c}
\toprule
\multirow{2}{*}{Method} & \multicolumn{2}{c}{Bradley–Terry Score $\uparrow$} \\
\cmidrule(lr){2-3}
 & Single-Object & Multi-Object \\
\midrule
SGRNet~\cite{hong2022shadow} & 0.191 & 0.013 \\
DMASNet~\cite{tao2024shadow} & 0.353 & 0.189 \\
SGDiffusion~\cite{liu2024shadow} & 0.552 & 0.378 \\
GPSDiffusion~\cite{zhao2025shadow} & 0.912 & 0.734 \\
MetaShadow~\cite{wang2025metashadow} & 0.889 & 0.701 \\
\midrule
\textbf{MultiShadow} (Ours) & \textbf{1.929} & \textbf{1.587} \\
\bottomrule
\end{tabular}
\end{table}

\subsubsection{Shadow-Box Predictor Evaluation}
Since our text-grounded tokens are derived from predicted shadow bounding boxes, we explicitly evaluate the shadow-box predictor's accuracy and analyze how localization errors affect the final generation quality. This analysis addresses a critical question: does imperfect box prediction undermine the benefits of text-grounded conditioning? We report box localization accuracy in Table~\ref{tab:box_diagnostics} using IoU and recall at IoU$=0.5$ (R@0.5), computed against ground-truth boxes extracted from the ground-truth shadow masks. The predictor achieves strong localization performance.

To quantify error propagation, we compare shadow generation conditioned on predicted boxes against an oracle setting in which predicted boxes are replaced with ground-truth boxes. The oracle yields only a small improvement, indicating that accurate localization is beneficial, but that our model remains effective under predicted-box conditioning. Finally, we assess robustness by perturbing ground-truth boxes with translation and scale noise of $\pm5\%$ and $\pm10\%$ of the box size. Performance degrades only slightly, demonstrating that the proposed grounding mechanism is stable under moderate layout noise.

\begin{table}[t]
\centering
\caption{Shadow-box predictor Evaluation. Localization accuracy, oracle impact, and robustness to box noise.}
\label{tab:box_diagnostics}
\begin{tabular}{l l c c c c}
\toprule
\textbf{Case} & \textbf{Setting} & \textbf{IoU} $\uparrow$ & \textbf{R@0.5} $\uparrow$ & \textbf{LR} $\downarrow$ & \textbf{LB} $\downarrow$ \\
\midrule
Predictor Quality & Pred vs. GT & 0.83 & 0.92 & -- & -- \\
\midrule
\multirow{2}{*}{Generation Impact} & Pred boxes & -- & -- & 51.32 & 0.428 \\
& Oracle (GT) & -- & -- & 49.90 & 0.415 \\
\midrule
\multirow{2}{*}{Robustness} & $\pm$5\% noise & -- & -- & 50.87 & 0.431 \\
& $\pm$10\% noise & -- & -- & 51.86 & 0.438 \\
\bottomrule
\end{tabular}
\end{table}

\subsection{Ablation Study}
\label{sec:ablation}
We conduct extensive ablation studies to validate the design choices of our framework and quantify the contribution of each component. All experiments are performed on the BOS test split of the extended multi-object DESOBAv2 dataset. To assess the impact of each module, we progressively add components and report the resulting changes in Table~\ref{tab:ablation_incremental}. Fig.~\ref{fig:ab1} complements the table with qualitative results, showing a clear step-by-step improvement as each component is introduced. 

Starting from a pre-trained text-to-image diffusion model with simple image conditioning (SIC)~\cite{zhang2023adding}, the model produces weak or missing shadows. Introducing hierarchical image conditioning through Geometry-Aware Affine Modulation (GAAM) improves shadow geometry via pixel-aligned guidance, but some objects remain shadowless. Adding generic prompts (GP), without any positional token information, provides an additional consistent gain, indicating that activating the text-conditioning pathway helps the diffusion backbone better utilize its pre-trained shadow priors and encourages more reliable shadow synthesis.

\renewcommand{\arraystretch}{1.4}
\setlength{\tabcolsep}{1.0pt}

\begin{table}[t]
\caption{Ablation study on Multi-object extension of DESOBAv2 BOS test set. SIC: Simple Image Conditioning, GAAM: Geometry-Aware Affine Modulation, GP: General Prompt, IBBox: Image-space Box, SPT: Shadow Positional Tokens, AAL: Attention Alignment Loss.}
\label{tab:ablation_incremental}
\centering
\small
\begin{tabular}{l|cccccc|ccccc}
\toprule
Case & SIC & GAAM & GP & IBBox & SPT & AAL & GR $\downarrow$ & LR $\downarrow$ & GB $\downarrow$ & LB $\downarrow$ \\
\midrule
1 & + & - & - & - & - & - & 12.831 & 73.32 & 0.378 & 0.572 \\
2 & + & + & - & - & - & - & 9.546 & 62.563 & 0.336 & 0.538 \\
3 & + & + & + & - & - & - & 8.861 & 60.492 & 0.301 & 0.516 \\
4 & + & + & + & + & - & - & 7.587 & 57.242 & 0.287 & 0.462 \\
5 & + & + & + & - & + & - & 7.388 & 53.573  & 0.245 & 0.431 \\
6 & + & + & + & - & + & + & 6.892 & 51.327  & 0.235 & 0.428 \\
\bottomrule
\end{tabular}
\end{table}

\begin{figure}[t]
\centering

\setlength{\tabcolsep}{0.5pt}          
\renewcommand{\arraystretch}{0}        
\newcommand{\colw}{0.192\columnwidth}  
\newcommand{\hgap}{0.6mm}              
\newcommand{\rgap}{0.3mm}              
\newcommand{\hdr}[1]{%
  \parbox[c][3.6mm][c]{\colw}{\centering\fontsize{8}{8}\selectfont #1}%
}
\newcommand{\im}[1]{\includegraphics[width=\colw]{#1}}
% --------------------------------------

\begin{tabular}{@{}ccccc@{}}
\hdr{Composite} &
\hdr{Objects Mask} &
\hdr{IBBox} &
\hdr{SPT{+}AAL} &
\hdr{GT} \\[\hgap]

\im{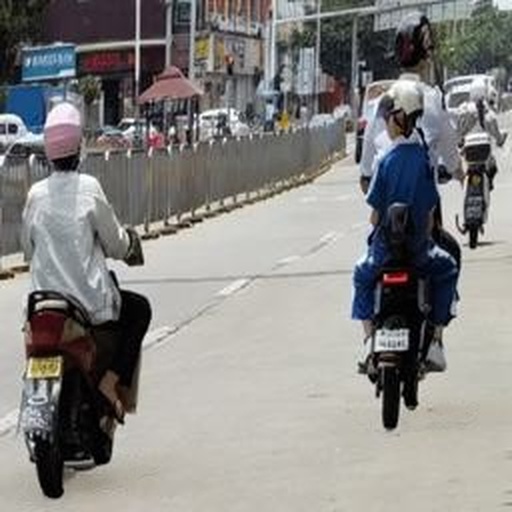} &
\im{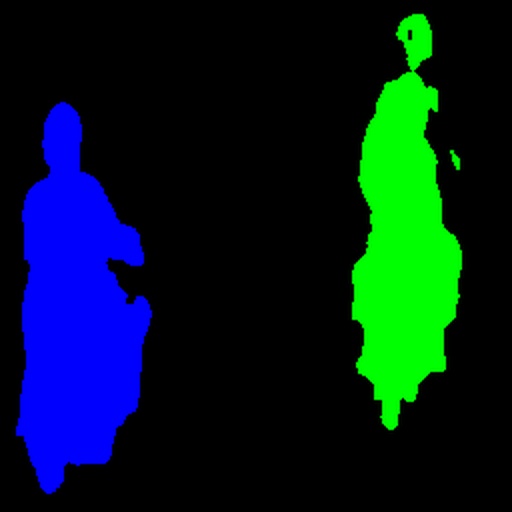} &
\im{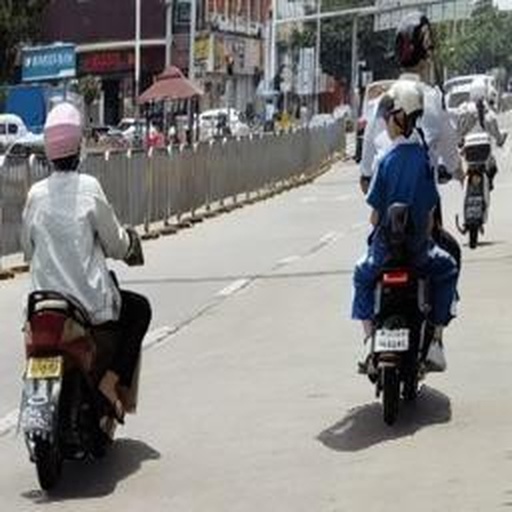} &
\im{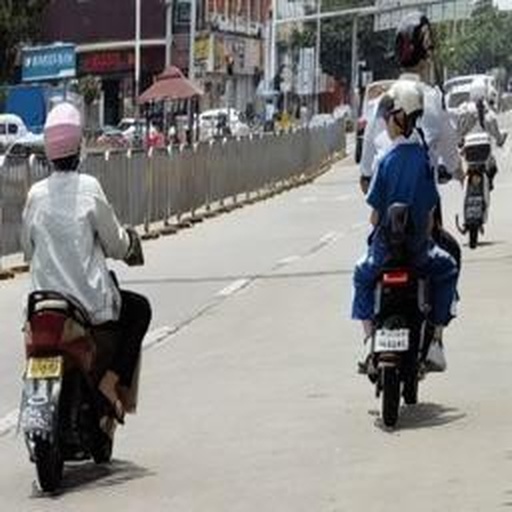} &
\im{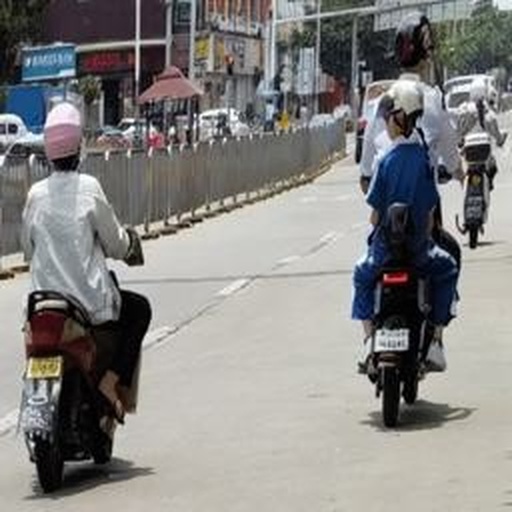} \\[\rgap]

\im{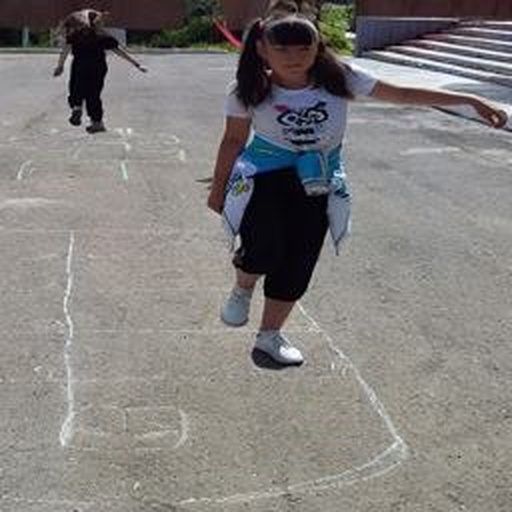} &
\im{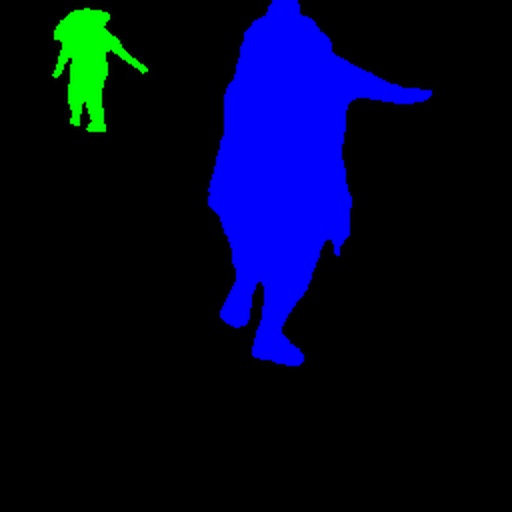} &
\im{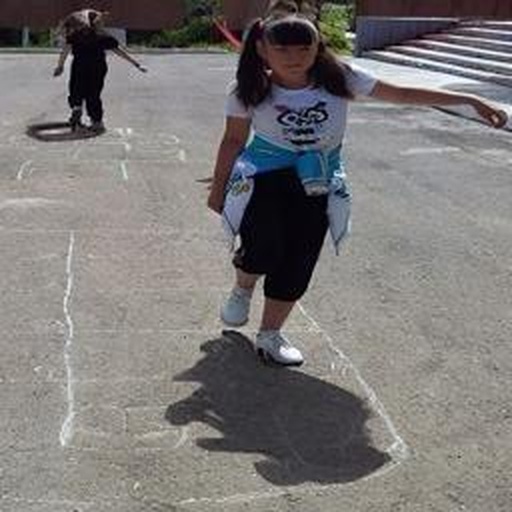} &
\im{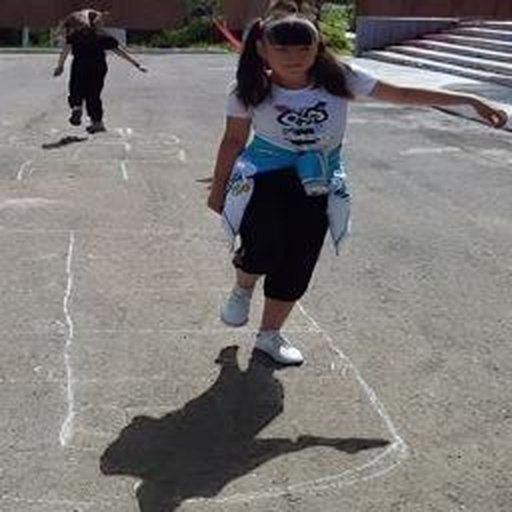} &
\im{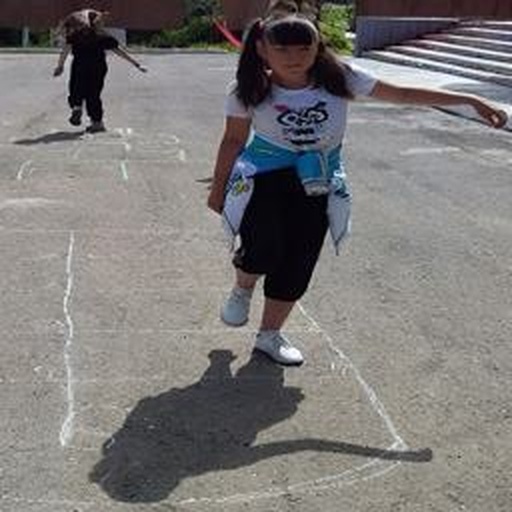} \\[\rgap]

\im{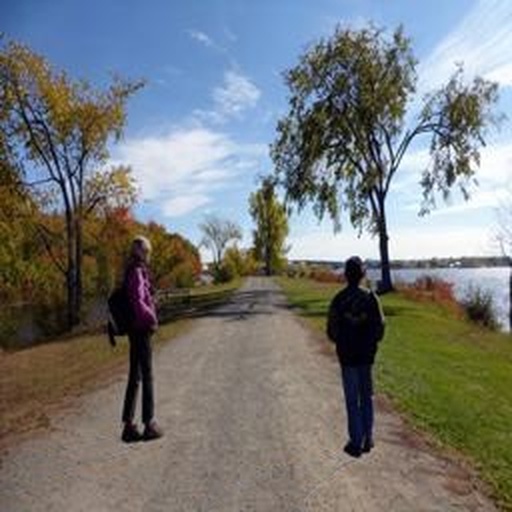} &
\im{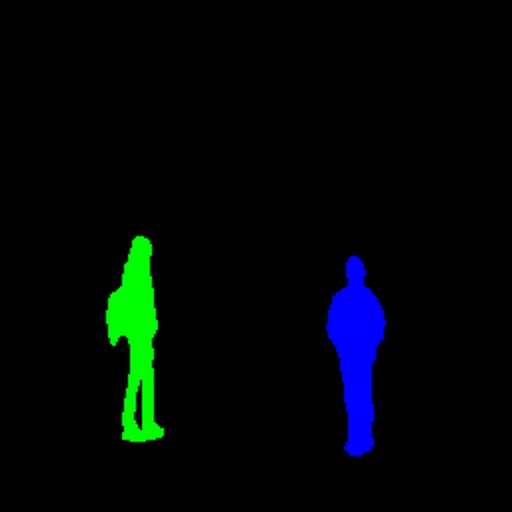} &
\im{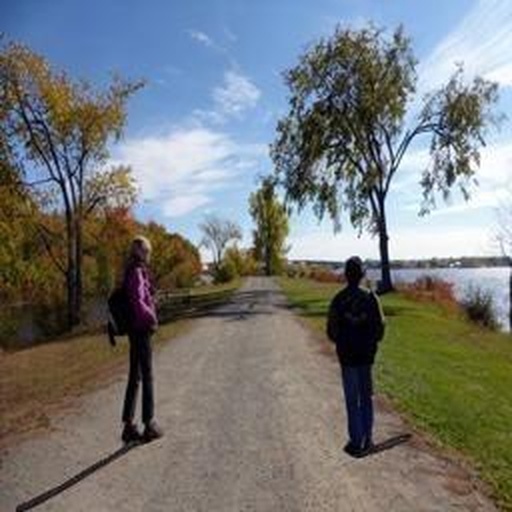} &
\im{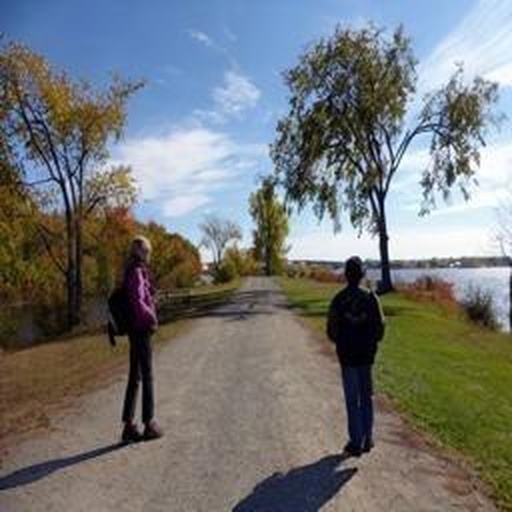} &
\im{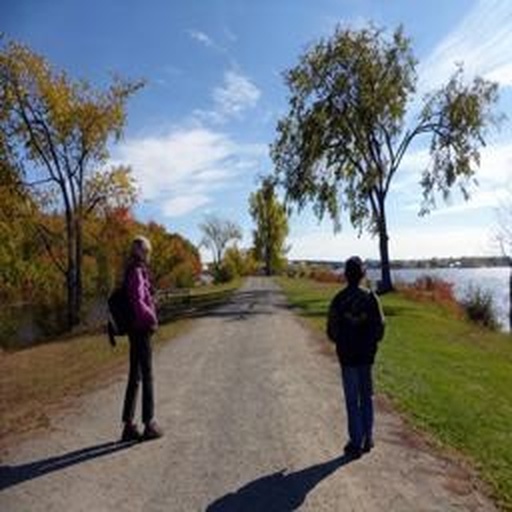} \\[\rgap]

\end{tabular}

\vspace{-1mm}
\caption{Qualitative ablation comparison between image-space bounding box (IBBox) conditioning and our text-grounded shadow prompt conditioning.}
\label{fig:ab2}
\vspace{-2mm}
\end{figure}

We next compare two design choices for providing explicit per-object shadow layout guidance. A straightforward approach concatenates per-object shadow boxes as additional image channels (IBBox), similar in spirit to GPSDiffusion~\cite{zhao2025shadow}. While IBBox provides coarse spatial guidance, it remains a purely pixel-space representation and behave inconsistently in multi-object scenes. Our key design choice is instead to encode each object’s shadow box as discrete shadow positional tokens (SPT) within the prompt, and inject them through the text–image cross-attention layers. As shown in Table~\ref{tab:ablation_incremental}, SPT consistently outperforms IBBox, particularly on local metrics, indicating sharper geometry and cleaner attachment. Figure~\ref{fig:ab2} illustrates this gap, IBBox often produces drifting or misaligned shadows and inconsistent directions across objects, whereas SPT yields more coherent per-object placemen. Adding the attention alignment loss (AAL) further improves both quantitative and qualitative results by explicitly encouraging token-specific cross-attention to focus on the corresponding shadow regions. This reduces artifacts and improves multi-object consistency.

\section{Conclusion}
We introduced a novel diffusion-based framework for shadow generation that couples image-based conditioning with text-grounded shadow positional tokens. This dual design approach explicitly encodes per-object shadow layout and steers cross-attention to the right regions, yielding state-of-the-art performance across single and multi-object scenarios. The resulting shadows exhibit stronger attachment, geometry, and photometric consistency than image-only approaches.

 \bibliographystyle{IEEEtran}
 \bibliography{ref}

\begin{thebibliography}{10}
\providecommand{\url}[1]{#1}
\csname url@samestyle\endcsname
\providecommand{\newblock}{\relax}
\providecommand{\bibinfo}[2]{#2}
\providecommand{\BIBentrySTDinterwordspacing}{\spaceskip=0pt\relax}
\providecommand{\BIBentryALTinterwordstretchfactor}{4}
\providecommand{\BIBentryALTinterwordspacing}{\spaceskip=\fontdimen2\font plus
\BIBentryALTinterwordstretchfactor\fontdimen3\font minus \fontdimen4\font\relax}
\providecommand{\BIBforeignlanguage}[2]{{%
\expandafter\ifx\csname l@#1\endcsname\relax
\typeout{** WARNING: IEEEtran.bst: No hyphenation pattern has been}%
\typeout{** loaded for the language `#1'. Using the pattern for}%
\typeout{** the default language instead.}%
\else
\language=\csname l@#1\endcsname
\fi
#2}}
\providecommand{\BIBdecl}{\relax}
\BIBdecl

\bibitem{niu2021making}
L.~Niu, W.~Cong, L.~Liu, Y.~Hong, B.~Zhang, J.~Liang, and L.~Zhang, ``Making images real again: A comprehensive survey on deep image composition,'' \emph{arXiv preprint arXiv:2106.14490}, 2021.

\bibitem{tao2024shadow}
X.~Tao, J.~Cao, Y.~Hong, and L.~Niu, ``Shadow generation with decomposed mask prediction and attentive shadow filling,'' in \emph{Proceedings of the AAAI Conference on Artificial Intelligence}, vol.~38, no.~6, 2024, pp. 5198--5206.

\bibitem{meng2023automatic}
Q.~Meng, S.~Zhang, Z.~Li, C.~Wang, W.~Zhang, and Q.~Huang, ``Automatic shadow generation via exposure fusion,'' \emph{IEEE Transactions on Multimedia}, vol.~25, pp. 9044--9056, 2023.

\bibitem{yu2024cfdiffusion}
Z.~Yu, J.~Zhou, Z.~Bao, G.~Fu, W.~He, C.~Liang, and C.~Xiao, ``Cfdiffusion: Controllable foreground relighting in image compositing via diffusion model,'' in \emph{Proceedings of the 32nd ACM International Conference on Multimedia}, 2024, pp. 3647--3656.

\bibitem{liu2024shadow}
Q.~Liu, J.~You, J.~Wang, X.~Tao, B.~Zhang, and L.~Niu, ``Shadow generation for composite image using diffusion model,'' in \emph{Proceedings of the IEEE/CVF Conference on Computer Vision and Pattern Recognition}, 2024, pp. 8121--8130.

\bibitem{zhao2025shadow}
H.~Zhao, Q.~Liu, X.~Tao, L.~Niu, and G.~Zhai, ``Shadow generation using diffusion model with geometry prior,'' in \emph{Proceedings of the Computer Vision and Pattern Recognition Conference}, 2025, pp. 7603--7612.

\bibitem{wang2025metashadow}
T.~Wang, J.~Zhang, H.~Zheng, Z.~Ding, S.~Cohen, Z.~Lin, W.~Xiong, C.-W. Fu, L.~Figueroa, and S.~Y. Kim, ``Metashadow: Object-centered shadow detection, removal, and synthesis,'' in \emph{Proceedings of the Computer Vision and Pattern Recognition Conference}, 2025, pp. 28\,252--28\,262.

\bibitem{rombach2022high}
R.~Rombach, A.~Blattmann, D.~Lorenz, P.~Esser, and B.~Ommer, ``High-resolution image synthesis with latent diffusion models,'' in \emph{Proceedings of the IEEE/CVF conference on computer vision and pattern recognition}, 2022, pp. 10\,684--10\,695.

\bibitem{yang2023reco}
Z.~Yang, J.~Wang, Z.~Gan, L.~Li, K.~Lin, C.~Wu, N.~Duan, Z.~Liu, C.~Liu, M.~Zeng \emph{et~al.}, ``Reco: Region-controlled text-to-image generation,'' in \emph{Proceedings of the IEEE/CVF Conference on Computer Vision and Pattern Recognition}, 2023, pp. 14\,246--14\,255.

\bibitem{li2023gligen}
Y.~Li, H.~Liu, Q.~Wu, F.~Mu, J.~Yang, J.~Gao, C.~Li, and Y.~J. Lee, ``Gligen: Open-set grounded text-to-image generation,'' in \emph{Proceedings of the IEEE/CVF conference on computer vision and pattern recognition}, 2023, pp. 22\,511--22\,521.

\bibitem{zheng2023layoutdiffusion}
G.~Zheng, X.~Zhou, X.~Li, Z.~Qi, Y.~Shan, and X.~Li, ``Layoutdiffusion: Controllable diffusion model for layout-to-image generation,'' in \emph{Proceedings of the IEEE/CVF Conference on Computer Vision and Pattern Recognition}, 2023, pp. 22\,490--22\,499.

\bibitem{yu2024uncovering}
H.~Yu, H.~Luo, F.~Wang, and F.~Zhao, ``Uncovering the text embedding in text-to-image diffusion models,'' \emph{arXiv preprint arXiv:2404.01154}, 2024.

\bibitem{liu2023desobav2}
Q.~Liu, J.~Wang, and L.~Niu, ``Desobav2: Towards large-scale real-world dataset for shadow generation,'' \emph{arXiv preprint arXiv:2308.09972}, 2023.

\bibitem{cai2024vip}
M.~Cai, H.~Liu, S.~K. Mustikovela, G.~P. Meyer, Y.~Chai, D.~Park, and Y.~J. Lee, ``Vip-llava: Making large multimodal models understand arbitrary visual prompts,'' in \emph{Proceedings of the IEEE/CVF Conference on Computer Vision and Pattern Recognition}, 2024, pp. 12\,914--12\,923.

\bibitem{song2022objectstitch}
Y.~Song, Z.~Zhang, Z.~Lin, S.~Cohen, B.~Price, J.~Zhang, S.~Y. Kim, and D.~Aliaga, ``Objectstitch: Generative object compositing,'' \emph{arXiv preprint arXiv:2212.00932}, 2022.

\bibitem{yang2023paint}
B.~Yang, S.~Gu, B.~Zhang, T.~Zhang, X.~Chen, X.~Sun, D.~Chen, and F.~Wen, ``Paint by example: Exemplar-based image editing with diffusion models,'' in \emph{Proceedings of the IEEE/CVF conference on computer vision and pattern recognition}, 2023, pp. 18\,381--18\,391.

\bibitem{chen2024anyscene}
R.~Chen, L.~Wang, W.~Nie, Y.~Zhang, and A.-A. Liu, ``Anyscene: Customized image synthesis with composited foreground,'' in \emph{Proceedings of the IEEE/CVF Conference on Computer Vision and Pattern Recognition}, 2024, pp. 8724--8733.

\bibitem{chen2025anydoor}
X.~Chen, L.~Huang, Y.~Liu, Y.~Shen, D.~Zhao, and H.~Zhao, ``Anydoor: zero-shot image customization with region-to-region reference,'' \emph{IEEE Transactions on Pattern Analysis and Machine Intelligence}, 2025.

\bibitem{tarres2025multitwine}
G.~C. Tarr{\'e}s, Z.~Lin, Z.~Zhang, H.~Zhang, A.~Gilbert, J.~Collomosse, and S.~Y. Kim, ``Multitwine: Multi-object compositing with text and layout control,'' in \emph{Proceedings of the Computer Vision and Pattern Recognition Conference}, 2025, pp. 8094--8104.

\bibitem{zhou2025bootplace}
H.~Zhou, X.~Zuo, R.~Ma, and L.~Cheng, ``Bootplace: Bootstrapped object placement with detection transformers,'' in \emph{Proceedings of the Computer Vision and Pattern Recognition Conference}, 2025, pp. 19\,294--19\,303.

\bibitem{sheng2021ssn}
Y.~Sheng, J.~Zhang, and B.~Benes, ``Ssn: Soft shadow network for image compositing,'' in \emph{Proceedings of the IEEE/CVF Conference on Computer Vision and Pattern Recognition}, 2021, pp. 4380--4390.

\bibitem{sheng2023pixht}
Y.~Sheng, J.~Zhang, J.~Philip, Y.~Hold-Geoffroy, X.~Sun, H.~Zhang, L.~Ling, and B.~Benes, ``Pixht-lab: Pixel height based light effect generation for image compositing,'' in \emph{Proceedings of the IEEE/CVF Conference on Computer Vision and Pattern Recognition}, 2023, pp. 16\,643--16\,653.

\bibitem{zhou2024foreground}
J.~Zhou, Z.~Yu, Z.~Bao, G.~Fu, W.~He, C.~Liang, and C.~Xiao, ``Foreground harmonization and shadow generation for composite image,'' in \emph{Proceedings of the 32nd ACM International Conference on Multimedia}, 2024, pp. 8267--8276.

\bibitem{zhang2019shadowgan}
S.~Zhang, R.~Liang, and M.~Wang, ``Shadowgan: Shadow synthesis for virtual objects with conditional adversarial networks,'' \emph{Computational Visual Media}, vol.~5, no.~1, pp. 105--115, 2019.

\bibitem{liu2020arshadowgan}
D.~Liu, C.~Long, H.~Zhang, H.~Yu, X.~Dong, and C.~Xiao, ``Arshadowgan: Shadow generative adversarial network for augmented reality in single light scenes,'' in \emph{Proceedings of the IEEE/CVF conference on computer vision and pattern recognition}, 2020, pp. 8139--8148.

\bibitem{hong2022shadow}
Y.~Hong, L.~Niu, and J.~Zhang, ``Shadow generation for composite image in real-world scenes,'' in \emph{Proceedings of the AAAI conference on artificial intelligence}, vol.~36, no.~1, 2022, pp. 914--922.

\bibitem{zhang2023adding}
L.~Zhang, A.~Rao, and M.~Agrawala, ``Adding conditional control to text-to-image diffusion models,'' in \emph{Proceedings of the IEEE/CVF international conference on computer vision}, 2023, pp. 3836--3847.

\bibitem{wang2022instance}
T.~Wang, X.~Hu, P.-A. Heng, and C.-W. Fu, ``Instance shadow detection with a single-stage detector,'' \emph{IEEE transactions on pattern analysis and machine intelligence}, vol.~45, no.~3, pp. 3259--3273, 2022.

\bibitem{SD-XL_Inpainting_0.1}
{Hugging Face} and {Diffusers}, ``Sdxl inpainting 0.1 model,'' \url{https://huggingface.co/diffusers/stable-diffusion-xl-1.0-inpainting-0.1}, 2023.

\bibitem{bradley1952rank}
R.~A. Bradley and M.~E. Terry, ``Rank analysis of incomplete block designs: I. the method of paired comparisons,'' \emph{Biometrika}, vol.~39, no. 3/4, pp. 324--345, 1952.

\end{thebibliography}

\vfill
\end{document}